\documentclass{article}
\usepackage{iclr2027_conference,times}
\usepackage{amsmath,amsfonts,bm}

\def\eqref#1{equation~\ref{#1}}
\def\1{\bm{1}}

\DeclareMathAlphabet{\mathsfit}{\encodingdefault}{\sfdefault}{m}{sl}
\SetMathAlphabet{\mathsfit}{bold}{\encodingdefault}{\sfdefault}{bx}{n}

\newcommand{\Var}{\mathrm{Var}}

\newcommand{\Cov}{\mathrm{Cov}}
\renewcommand{\eqref}[1]{\textup{(\ref{#1})}}

\usepackage{hyperref}
\hypersetup{hidelinks}
\usepackage{xurl}
\usepackage{amsmath,amssymb,amsthm}
\usepackage{graphicx}
\usepackage{array}
\usepackage{booktabs}
\usepackage{xcolor}
\usepackage{placeins}
\AtBeginEnvironment{table}{\setlength{\belowcaptionskip}{4pt}}

\newtheorem{theorem}{Theorem}
\newtheorem{proposition}{Proposition}

\newcommand{\logit}{\operatorname{logit}}
\newcommand{\passk}{\text{pass@}k}
\newcommand{\kstar}{k^{\star}}
\newcommand{\pstar}{p^{\star}}

\iclrfinalcopy

\title{When Pass@$k$ Crossovers Are Real---and Why: Paired Inference and the Markov Kernel of RLVR}

\author{Chen Yang\\
Department of Statistics, Texas A\&M University\\
\texttt{cafferychen777@tamu.edu}}

\begin{document}
\maketitle

\begin{abstract}
Reinforcement learning with verifiable rewards (RLVR) typically improves
pass@1 but can fall behind its base model at large sampling budgets $k$, a
crossover widely interpreted as evidence that RLVR only sharpens solutions the
base model already knows how to produce. Because pass@$k$ estimates at
adjacent $k$ reuse the same generations and are nearly collinear, pointwise
error bars cannot establish a sign change. We formulate the crossover as a
paired inference problem on the difference curve
$D(k)=\passk^{\mathrm{RL}}-\passk^{\mathrm{base}}$ over the same prompts: a
prompt-level multiplier bootstrap yields a one-sided dominance test and
simultaneous confidence bands that certify a crossover only when they show
both an early gain and a later loss. Across nine public RLVR checkpoints at
standard 3k--16k token limits, only one narrowly meets this criterion before
multiplicity adjustment, and per-response token limits alone can create,
mask, or reverse a crossover through asymmetric truncation. On a clean
32k-token benchmark of 1,060 deduplicated competition problems, two 1.5B
models exhibit certified crossovers around $k\approx10$--$60$, whereas two 7B
models show no supported loss; a paired variance decomposition explains why
adding prompts raises statistical power far more effectively than deeper
per-prompt sampling.

To explain why genuine crossovers occur, we examine how RLVR acts on
individual prompts. Prompts with identical base success probability $p$
diverge to widely different post-RL success probabilities $q$ that replicate
across independent generation halves and violate the Lipschitz slope bounds
of any reasonable single curve. RLVR therefore acts as a Markov kernel
$H(q\mid p)$ rather than a deterministic function $q=T(p)$, and Jensen's
inequality implies that conditional dispersion strictly depresses pass@$k$ for
$k\ge2$. Once conditional dispersion is modeled, the conditional-mean
log-odds slope falls below one across every model pair: on average, RLVR
improves hard prompts, and the late-$k$ loss arises from a $2\%$--$3\%$
minority of base-solved prompts that collapse onto a repeated modal wrong
answer and resist prefix rescue up to $4{,}096$ tokens. In controlled runs
from a single base model measured directly to $k=4096$ on 3,000 new problems,
all four GRPO arms fall behind the base near $k\approx200$ and remain behind
through $k=4096$, whereas SFT stays significantly ahead throughout.
\end{abstract}

\section{Introduction}\label{sec:intro}

Does reinforcement learning with verifiable rewards (RLVR) teach a language
model to solve problems its base model cannot, or does it only concentrate
probability on solutions the base model already knows how to produce? Since
\citet{yue2025limit}, the central empirical exhibit in this debate has been
a pass@$k$ plot: the RLVR model is ahead of its base model at $k=1$, the
base model overtakes it by $k=256$, and the two curves cross in between.
One line of work reads this crossover as evidence of sharpening within a
fixed support \citep{yue2025limit,wu2025leash}, while another attributes it
to the choice of metric or to insufficient training horizon
\citep{wen2025rlvr,liu2025prorl}. Both sides treat the visual crossing of
sample pass@$k$ curves as an established empirical fact. Where uncertainty
is reported at all, it consists of separate error bars at each $k$ across
training seeds \citep{zhou2026inversion,strozzi2026teacher}. From first
principles, separate pointwise intervals cannot establish a crossover: a
sign change is a joint claim across sampling budgets $k$, and pass@$k$
estimates at adjacent $k$ on the same prompts reuse the same generations
and have near-unit correlation.

This paper addresses both questions in turn: \emph{when} a pass@$k$
crossover is statistically real (Sections~\ref{sec:infer}--\ref{sec:experiments}),
and \emph{what mechanism} produces it when it is
(Sections~\ref{sec:model}--\ref{sec:planned}).

\begin{figure}[t]
\centering
\includegraphics[width=\textwidth]{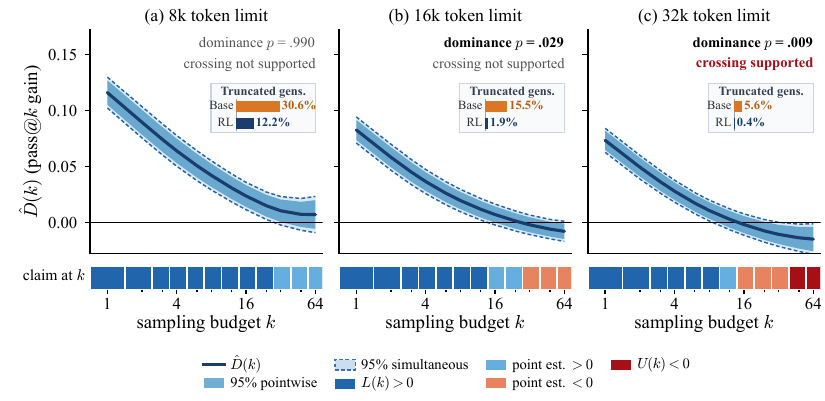}
\caption{\textbf{The per-response token limit determines which crossing
claim DeepScaleR's data support.} DeepScaleR-1.5B versus its base model
(R1-Distill-Qwen-1.5B) on the same 1,000 prompts across three per-response
token limits ($n=64$, $256$, and $64$ at 8k, 16k, and 32k tokens; $k=1,\dots,64$).
Solid curves show $\hat D(k)$ with 95\% pointwise (shaded) and simultaneous
sup-$t$ (dashed) confidence bounds $[L(k),U(k)]$. Inset horizontal bar charts
report the generation truncation rates of the base model (orange) and RL
checkpoint (blue): asymmetric base truncation falls from $30.6\%$ vs.\ $12.2\%$
at 8k to $15.5\%$ vs.\ $1.9\%$ at 16k and $5.6\%$ vs.\ $0.4\%$ at 32k. The
bottom strip colors each budget $k$ by the strongest supported claim; dark-red
cells indicating a statistically supported loss ($U(k)<0$) appear only at 32k.}
\label{fig:ladder}
\end{figure}

Properly posed, a crossover is a claim about the population paired
difference curve $D(k)=\passk^{\mathrm{RL}}-\passk^{\mathrm{base}}$
evaluated on the same prompts. Certifying that a crossover occurs within a
measured window $k\le K$ requires two simultaneous pieces of statistical
evidence: the RL model must be significantly ahead at some smaller budget
$j$ and significantly behind at some larger budget $\ell>j$. Measured by
this criterion, published crossovers at standard token limits are far more
fragile than sample curves suggest. Across nine public RLVR checkpoints
evaluated at 3k--16k token limits---including four SimpleRL-Zoo checkpoints
cited as primary evidence of sharpening---only one narrowly meets this
criterion before multiplicity adjustment, and none survives family-wise
correction. For DeepScaleR-1.5B, the per-response token limit alone
determines the statistical verdict: on the same 1,000 prompts, there is no
crossing at an 8k-token limit, a visible but statistically unsupported
crossing at 16k, and a statistically supported crossing at 32k
(Figure~\ref{fig:ladder}). Because base models generate longer or less
disciplined chains of thought than RLVR checkpoints, tight token limits
censor base completions far more heavily and can manufacture or mask a
crossover. When we eliminate training-set overlap and evaluate at a
32k-token limit on 1,060 new competition problems, DeepScaleR-1.5B and
ProRL-1.5B exhibit small, statistically certified crossovers
($\kappa\in[11,61]$ and $[12,63]$), whereas two 7B long-chain RLVR models
show no supported loss.

Once genuine crossovers are isolated from truncation artifacts and sampling
noise, we turn to our second question: what mechanism produces them? The
prevailing sharpening account models RLVR as a deterministic response
function $q=T(p)$ that maps each prompt's base success probability $p$ to a
post-RL success probability $q$, lifting easier prompts ($q>p$) and
depressing the hardest prompts ($q<p$). The paired count data refute this
premise across every model pair we study. On our clean 32k competition
benchmark, two AIME~2024 problems that the shared base model solves at the
exact same rate---$14$ out of $128$ generations ($\hat p=.109$)---diverge
completely under ProRL: one is solved in $105$ of $128$ generations
($\hat q=.820$) and the other in $0$ of $128$ ($\hat q=0$). Splitting each
prompt's generations into two independent halves of 64 confirms that this
divergence is a property of the prompts, not binomial sampling noise
($53/64$ and $52/64$ versus $0/64$ and $0/64$). Even without smoothness assumptions,
the paired counts violate the Lipschitz slope bounds of any reasonable
single response curve.

RLVR's effect on prompt difficulty is therefore a Markov kernel
$H(q\mid p)$---the conditional law of $Q$ given $P=p$---rather than a
single function $q=T(p)$. From first principles, this distinction is
decisive for pass@$k$ because the mapping $q\mapsto 1-(1-q)^k$ is strictly
concave for every $k\ge2$. By Jensen's inequality, conditional dispersion in
$Q$ given $P=p$ strictly increases the post-RL failure probability
$\mathbb E[(1-Q)^k\mid P=p]$ above $(1-\mathbb E[Q\mid P=p])^k$. For
instance, a prompt solved with probability $.01$ per attempt succeeds at
least once in $256$ attempts with probability $.92$, whereas a prompt
pushed to $.0001$ succeeds with probability only $.025$.

Separating the conditional distribution $H(q\mid p)$ from its mean
$T(p)=\mathbb E[Q\mid P=p]$ overturns the uniform-sharpening picture. Once
conditional dispersion is modeled, the conditional-mean log-odds slope falls
\emph{below one} across every model pair: on average, RLVR improves hard
prompts rather than penalizing them, and collapsing each fitted kernel onto
its conditional mean delays or eliminates the crossovers. Both nonparametric
kernel fits and model-free count decompositions show that the late-$k$ loss
is driven instead by a \emph{bimodal split} on hard problems: among prompts
that the base model solves only once in $128$--$256$ attempts ($c_i=1$), the
unlocked prompts improve by a median factor of $2\times$ to $7\times$, while
a $2\%$--$3\%$ minority of base-solved prompts collapses to zero post-RL
success and accounts for the entire late-$k$ deficit. Forensic inspection of
these locked prompts reveals neither truncation nor formatting failure: over
$99\%$ of rollouts terminate naturally with a boxed answer, converging onto
a single repeated wrong answer (often the base model's own modal error). Even
conditioning on up to $4{,}096$ tokens of the base model's correct chain of
thought in a prefix-rescue experiment fails to recover the majority of
locked prompts, showing that RLVR has reinforced an attractive but incorrect
reasoning trajectory throughout the solution.

\paragraph{Summary of contributions.}
\begin{enumerate}
\item \textbf{Paired statistical inference and evaluation design for
pass@$k$ crossovers (\S\ref{sec:infer}--\S\ref{sec:experiments}).} We
develop a prompt-level multiplier bootstrap for the paired difference curve
$D(k)$ that yields a one-sided dominance test (Theorem~\ref{thm:test}),
two-sided simultaneous sup-$t$ confidence bands certifying crossovers, and a
confidence interval $[\underline\kappa,\overline\kappa]$ for the first-loss
budget (Proposition~\ref{prop:cert}). Re-evaluating public RLVR checkpoints
exposes per-response token limits as a major truncation confound, establishes
clean 32k crossovers on 1.5B models alongside non-crossings on 7B models,
and shows via a paired variance decomposition why adding prompts ($m$)
raises statistical power far more effectively than deeper per-prompt
sampling ($n$).

\item \textbf{Refuting the single-curve sharpening model
(\S\ref{sec:model}).} Starting from the exact KL-regularized exponential-tilt
identity, we state a universal marginal sign-change bound that holds for any
Markov kernel (Proposition~\ref{prop:signs}) and characterize what a
deterministic curve $q=T(p)$ implies for tail exponents, crossover scales,
and single crossings (Theorems~\ref{thm:crossover}--\ref{thm:single}). We then refute determinism across all 20 model
pairs and protocols using a split-generation residual covariance test and a
distribution-free lower bound on the Lipschitz constant of any compatible
single curve (Theorem~\ref{thm:slope}).

\item \textbf{Resolving the slope illusion and isolating the $2\%$--$3\%$
locked-minority mechanism (\S\ref{sec:kernel}).} Estimating $H(q\mid p)$ via
a 2D nonparametric maximum likelihood estimator (NPMLE) and a Beta-Binomial
random-effects kernel reconciles statistical power, improves held-out
generation log-likelihood by hundreds to thousands of nats, and shows that
the conditional-mean slope is below one. Mean-collapse counterfactuals,
model-free count decompositions, generation-level forensic audits, and
4,096-token prefix-rescue experiments demonstrate that crossovers arise not
from uniform degradation of hard problems, but from a $2\%$--$3\%$ minority
of prompts locked onto repeated wrong answers while peer prompts at the same
base difficulty improve several-fold.

\item \textbf{Controlled training modes and deep-$k$ extrapolation to
$k=4096$ (\S\ref{sec:planned}).} Training controlled GRPO ($G\in\{4,8,16\}$
and weak KL) and SFT arms from Qwen2.5-Math-1.5B and measuring directly to
$k=4096$ on 3,000 new problems, we show that all four GRPO arms cross below
the base model around $k\approx200$ and remain behind through $k=4096$,
whereas SFT stays significantly ahead at every budget. Comparing direct
$k=4096$ measurements against kernels fitted at depth $n=256$ demonstrates
both the accuracy of kernel predictions near the fitting horizon and the
fundamental sub-$1/n$ moment-resolution limit of deep-$k$ extrapolation.
\end{enumerate}

\section{Paired Statistical Inference for \texorpdfstring{pass@$k$}{pass@k} Crossovers}\label{sec:infer}

Consider a population of prompts from which $m$ prompts $i=1,\dots,m$ are
drawn i.i.d.\ with latent base success probability $p_i\sim F$ on $[0,1]$
and post-RL success probability $q_i\in[0,1]$. On each prompt $i$, we
observe $c_i\sim\mathrm{Bin}(n_i,p_i)$ successes out of $n_i$ i.i.d.\ base
generations and $d_i\sim\mathrm{Bin}(n_i',q_i)$ successes out of $n_i'$
i.i.d.\ post-RL generations, \emph{paired on the same prompt}. At sampling
budget $k$, a model's population $\passk$ is the probability that at least
one of $k$ independent generations solves a randomly drawn prompt:
\[
\passk^{\mathrm{base}} = 1 - \mathbb E\big[(1-P)^k\big], \qquad
\passk^{\mathrm{RL}} = 1 - \mathbb E\big[(1-Q)^k\big],
\]
and the paired difference curve is
\[
D(k) = \passk^{\mathrm{RL}} - \passk^{\mathrm{base}}
= \mathbb E\big[(1-P)^k - (1-Q)^k\big].
\]

For any integer budget $k\le\min(n_i,n_i')$, the minimum-variance unbiased
estimator of prompt $i$'s pass@$k$ under binomial sampling is the classical
U-statistic
\[
\widehat{\passk}^{\mathrm{base}}_i = 1 - \frac{\binom{n_i-c_i}{k}}{\binom{n_i}{k}},
\qquad
\widehat{\passk}^{\mathrm{RL}}_i = 1 - \frac{\binom{n_i'-d_i}{k}}{\binom{n_i'}{k}}.
\]
Let $\psi_i(k)=\widehat{\passk}^{\mathrm{RL}}_i-\widehat{\passk}^{\mathrm{base}}_i$
denote the unbiased paired difference on prompt $i$, so that
$\mathbb E[\psi_i(k)]=D(k)$. Across the $m$ prompts, we form the sample
difference curve and its prompt-level standard error:
\[
\hat D(k) = \frac{1}{m}\sum_{i=1}^m \psi_i(k), \qquad
\hat s(k)^2 = \frac{1}{m-1}\sum_{i=1}^m \big\{\psi_i(k) - \hat D(k)\big\}^2, \qquad
\hat\sigma(k) = \frac{\hat s(k)}{\sqrt{m}}.
\]

The inference tools below distinguish two questions that are often conflated
in visual readings of pass@$k$ plots, and neither requires a model of how
$q_i$ relates to $p_i$. The \emph{dominance test} asks a one-sided question:
is there statistical evidence that the RL model is worse than the base
model at \emph{any} measured budget $k\le K$? The \emph{crossing criterion}
asks a two-sided structural question: is there simultaneous statistical
evidence that the RL model is better at some smaller budget and worse at
some larger budget?

\begin{theorem}[Dominance test]\label{thm:test}
Let prompts $i=1,\dots,m$ be i.i.d.\ with $s(k)^2=\Var\{\psi_i(k)\}>0$ for
every $k\le K$ ($K$ fixed), and define the studentized minimum statistic
\[
\mathcal T_m = \min_{1\le k\le K}\frac{\hat D(k)}{\hat\sigma(k)}
= \min_{1\le k\le K}\frac{\sqrt{m}\,\hat D(k)}{\hat s(k)}.
\]
For i.i.d.\ standard normal multipliers $g_1,\dots,g_m$ independent of the
data, define the bootstrap process
\[
Z_m^*(k) = \frac{1}{\sqrt{m}}\sum_{i=1}^m g_i\big\{\psi_i(k) - \hat D(k)\big\}
\]
---drawing \emph{one multiplier per prompt, shared across all budgets
$k=1,\dots,K$}---and let $\mathcal T_m^*=\min_{1\le k\le K} Z_m^*(k)/\hat s(k)$,
with $\hat q_\alpha$ denoting its conditional lower-$\alpha$ quantile given
the data. Then the test that rejects the dominance null
$H_0: D(k)\ge0\ \text{for all } k\le K$ when $\mathcal T_m < \hat q_\alpha$
satisfies
\[
\limsup_{m\to\infty}\Pr_P\big(\mathcal T_m < \hat q_\alpha\big) \le \alpha
\]
for every data-generating law $P$ satisfying $H_0$, with asymptotic
equality at the all-binding boundary $D(k)=0\ \text{for all } k\le K$.
Moreover, the test is consistent against any fixed alternative $P$ for
which $\min_{1\le k\le K} D(k) < 0$.
\end{theorem}

Why must a single multiplier $g_i$ be shared across all $k$ for each prompt?
Under $H_0$, the boundary $D\equiv0$ is least favorable with respect to the
mean drift, since $D(k)\ge0$ implies
$\hat D(k)/\hat\sigma(k)\ge\{\hat D(k)-D(k)\}/\hat\sigma(k)$ pathwise at
every $k$. However, $D\equiv0$ fixes only the marginal means and does not
fix the correlation matrix of $\{\psi_i(k)\}_{k=1}^K$ across budgets. In
pass@$k$ curves, adjacent budgets $k$ and $k+1$ are nearly collinear, so a
Bonferroni correction across $K$ pointwise tests is severely conservative,
whereas an unadjusted minimum test over-rejects. Drawing one multiplier
$g_i$ per prompt preserves the empirical across-$k$ covariance structure of
the centered prompt contributions $\psi_i(k)-\hat D(k)$ and consistently
estimates the exact limiting Gaussian-minimum distribution under the data's
true correlation (see Appendix~\ref{app:proofs} for the proof and
Appendix~\ref{app:em} for finite-sample calibration confirming nominal size
across correlation structures).

Rejecting $H_0$ establishes that the RL model falls behind the base model
at some budget $k\le K$, but it does not by itself establish a crossover: a
model that is worse everywhere ($D(k)<0$ for all $k$) also rejects $H_0$,
and because Theorem~\ref{thm:test} uses a one-sided critical value, it can
reject slightly before a two-sided confidence band excludes zero (as occurs
in the 16k panel of Figure~\ref{fig:ladder}). Certifying a crossover
requires simultaneous control of both signs, which is provided by the
two-sided sup-$t$ confidence band.

\begin{proposition}[Simultaneous confidence-band criterion for a crossover and first-loss budget]\label{prop:cert}
Let $\hat c_{1-\alpha}$ be the conditional $(1-\alpha)$-quantile of
$\max_{1\le k\le K}|Z_m^*(k)|/\hat s(k)$, and form the level-$(1-\alpha)$
simultaneous confidence band
\[
\big[L(k),\, U(k)\big] = \big[\hat D(k) - \hat c_{1-\alpha}\hat\sigma(k),\;
\hat D(k) + \hat c_{1-\alpha}\hat\sigma(k)\big], \qquad k=1,\dots,K.
\]
On the coverage event $\{L(k)\le D(k)\le U(k)\ \text{for all } k\le K\}$,
which holds with asymptotic probability at least $1-\alpha$ as
$m\to\infty$:
\begin{enumerate}
\item[(i)] \textbf{Crossing certification:} If there exist budgets $j<\ell$
such that $L(j)>0$ and $U(\ell)<0$, then the population curve $D(k)$ has at
least one positive-to-negative sign change in $[j,\ell]$.
\item[(ii)] \textbf{First-loss budget interval:} The first sampling budget
at which the RL model falls behind the base model,
$\kappa=\inf\{k\ge1: D(k)<0\}$ (with $\inf\varnothing=\infty$), is bounded
by $\kappa\in[\underline\kappa,\overline\kappa]$, where
\[
\underline\kappa = 1 + \max\big\{j\ge0 : L(r)>0\ \text{for all } 1\le r\le j\big\},
\qquad
\overline\kappa = \min\big\{j\ge1 : U(j)<0\big\},
\]
under the boundary conventions that $\underline\kappa=1$ whenever $L(1)\le0$
(the condition at $j=0$ being vacuous) and $\overline\kappa=\infty$ whenever
$U(j)\ge0$ for all $j\le K$.
\end{enumerate}
\end{proposition}

To illustrate Proposition~\ref{prop:cert}, in DeepScaleR's 32k evaluation
on 1,060 new competition problems ($K=128$), the simultaneous band satisfies
$L(k)>0$ for all $1\le k\le10$ and $U(k)<0$ for all $61\le k\le128$. Thus a
crossover is statistically supported at the 95\% simultaneous level, and
the first-loss budget is localized to $\kappa\in[11,61]$.

Theorem~\ref{thm:test} and Proposition~\ref{prop:cert} establish a strict
hierarchy of empirical evidence within the measured window $k\le K$:
(1) a \emph{visual sign change} ($\hat D(k)<0$ at some $k$, which may be
pure sampling noise), (2) a \emph{dominance rejection} ($\mathcal T_m<\hat q_\alpha$,
establishing a loss somewhere in $k\le K$), and (3) a \emph{band-supported
crossover} ($L(j)>0$ and $U(\ell)<0$ for some $j<\ell$, establishing both
an early gain and a later loss). Table~\ref{tab:notation} in
Appendix~\ref{app:proofs} provides a complete reference dictionary of
model-free and kernel estimands across Parts~I and~II.

\section{Re-evaluating Public Crossovers: Token Limits and Statistical Power}\label{sec:experiments}

\subsection{Experimental Setup and the Truncation Confound}

We apply the paired inference framework of Section~\ref{sec:infer} to two
complementary suites of public RLVR evaluations, decoded throughout at
temperature $0.6$ and top-$p$ $0.95$ with symbolic-equivalence verification
via Math-Verify \citep{mathverify2025} (full protocol and asset details are
in Appendix~\ref{app:exp}). Each suite serves a distinct scientific role:
\begin{enumerate}
\item \textbf{Protocol-audit suite at published 3k--16k token limits.} To
audit the pass@$k$ crossovers reported in prior work under their native
evaluation settings, we first evaluate nine public RLVR checkpoints paired
with their exact base models on a shared 1,000-prompt pool comprising 500
problems from MATH-500 \citep{hendrycks2021math,lightman2024verify} and 500
problems from the DeepScaleR RL training corpus \citep{luo2025deepscaler}.
Because the two prompt sources differ in difficulty and training overlap
(source-equality test $p<10^{-4}$; Appendix~\ref{app:exp}), this suite
exposes how benchmark composition and token limits affect reported
crossovers. It spans long-chain-of-thought models trained from
R1-Distill-Qwen-1.5B \citep{guo2025deepseekr1}---DeepScaleR-1.5B-Preview
\citep{luo2025deepscaler} and ProRL \citep{liu2025prorl}, evaluated across
8k, 16k, and 32k token limits and on 542 code problems from
HumanEval+/MBPP+---the four SimpleRL-Zoo checkpoints \citep{zeng2025simplerl}
on Qwen2.5-7B, 14B, 32B \citep{yang2024qwen25} and Llama-3.1-8B
\citep{grattafiori2024llama3} used as primary exhibits of sharpening by
\citet{yue2025limit}, and three additional 1.5B checkpoints: SimpleRL-Zoo
and Open-Reasoner-Zero (ORZ) \citep{hu2025orz} on Qwen2.5-1.5B, and LUFFY
\citep{yan2025luffy} on Qwen2.5-Math-1.5B (full sampling depths in
Table~\ref{tab:ladder}).

\item \textbf{Clean 32k reference benchmark on 1,060 deduplicated competition
problems.} To eliminate training-set overlap and remove token-limit
censoring as a confound, we assembled 1,060 competition problems from
OlympiadBench \citep{he2024olympiadbench}, Gaokao-2023-EN
\citep{liao2024mario}, AMC~2023, and AIME~2024, strictly deduplicated against
DeepScaleR's training set and every other prompt set in this study
(source breakdown in Appendix~\ref{app:exp}). With the analysis protocol
locked before sampling, we evaluate four long-chain RLVR models at a
32,768-token limit with $n=n'=128$: DeepScaleR-1.5B and ProRL-1.5B against
their shared base R1-Distill-Qwen-1.5B, and AceReason-Nemotron-7B
\citep{chen2025acereason} and Skywork-OR1-7B \citep{he2025skywork} against
their shared base R1-Distill-Qwen-7B. Once established below, this clean 32k
benchmark serves as our primary testbed for the mechanism analyses in
Sections~\ref{sec:model}--\ref{sec:kernel}, with complete multi-dataset
tables in the Appendix.
\end{enumerate}

\paragraph{The per-response token-limit confound.} Before examining crossing
tests, the data reveal a major measurement confound: base models and RLVR
models truncate at the per-response token limit at sharply different rates,
and truncated completions that fail to reach a boxed answer are scored as
incorrect. On the 1,000-prompt pool at an 8k-token limit, the
R1-Distill-Qwen-1.5B base model truncates $30.6\%$ of its generations,
compared with $12.2\%$ for DeepScaleR and $2.0\%$ for ProRL
(Figure~\ref{fig:ladder}). Raising the limit from 8k to 16k cuts base
truncation in half (to $15.5\%$) and reduces the logit-scale intercept shift
$\hat\gamma$ of ProRL's fitted response curve from $3.27$ to $2.56$
(Appendix~\ref{app:em}): roughly $0.7$ log-odds of ProRL's apparent pass@1
advantage at 8k was simply the base model running out of tokens. At 32k,
truncation falls to $5.6\%$ (base) and $0.35\%$ (DeepScaleR) on the
1,000-prompt mixture, and $7.8\%$ versus $0.7\%$ on the 1,060 new
competition problems. Consequently, the token limit is an integral part of
the pass@$k$ estimand.

\begin{table}[t]
\caption{\textbf{Statistical evaluation of pass@$k$ crossovers across
public RLVR pairs, token limits, and the clean 32k competition benchmark.}
All simultaneous bands and first-loss intervals
$\kappa\in[\underline\kappa,\overline\kappa]$ are computed on the dense
integer grid $k=1,\dots,K$ via the prompt-level multiplier bootstrap
($B=4000$, $\alpha=.05$). Bold entries indicate a dominance $p$-value below
the Holm family-wise threshold $.01$ together with a band-supported
crossover. $^{*}$Unadjusted per-pair $p<.05$, not family-wise significant
under Holm adjustment (Appendix~\ref{app:multiplicity}). $^{\S}$Pooled
500-prompt replication at 16k ($n=64$) stays positive throughout $k\le64$.
$^{\dagger}$Checkpoints analyzed by \citet{yue2025limit}.}
\label{tab:ladder}\label{tab:longchain}
\centering
\footnotesize
\setlength{\tabcolsep}{0pt}
\begin{tabular*}{\textwidth}{@{\extracolsep{\fill}}llcccccc@{}}
\toprule
Model / setting & Base model & \shortstack{Trunc.\\RL / base} &
\shortstack{$\hat D(1)\to$\\$\hat D(K)$} & \shortstack{Visual\\cross} &
\shortstack{Dominance\\rej.\ ($p$)} &
\shortstack{Cross.\\supp.} &
\shortstack{First-loss\\$\kappa\in$} \\
\midrule
\multicolumn{8}{@{}l@{\extracolsep{0pt}}}{\emph{Clean 32k benchmark (1,060 new problems, $n{=}128$; pass@1: $.68/.60, .72/.60, .80/.74, .78/.74$)}} \\
DeepScaleR-1.5B (32k)          & R1-Distill-1.5B & $0.7 / 7.8\%$  & $+.076\to-.020$ & $k{=}22$ & \textbf{yes (.0027)}      & \textbf{yes} & $\mathbf{[11,61]}$ \\
ProRL-1.5B (32k)               & R1-Distill-1.5B & $.03 / 7.8\%$  & $+.114\to-.021$ & $k{=}25$ & \textbf{yes (.0022)}      & \textbf{yes} & $\mathbf{[12,63]}$ \\
AceReason-7B (32k)             & R1-Distill-7B   & $0.2 / 2.8\%$  & $+.053\to+.008$ & none     & no (${>}.99$)             & no           & $[34,\infty)$ \\
Skywork-OR1-7B (32k)           & R1-Distill-7B   & $7.6 / 2.8\%$  & $+.037\to-.001$ & $k{=}15$ & no (.66)                  & no           & $[6,\infty)$  \\
\midrule
\multicolumn{8}{@{}l@{\extracolsep{0pt}}}{\emph{DeepScaleR-1.5B token-limit ladder on the 1,000-prompt mixture (and 500-prompt 16k repl.)}} \\
8k token limit ($n{=}64$)      & R1-Distill-1.5B & $12.2 / 30.6\%$& $+.191\to+.005$ & none     & no (.99)                  & no           & $[28,\infty)$ \\
16k token limit ($n{=}256$)    & R1-Distill-1.5B & $1.9 / 15.5\%$ & $+.083\to-.011$ & $k{=}27$ & unadj.\ (.029)$^{*}$      & no           & $[14,\infty)$ \\
16k repl.\ ($500\times64$)     & R1-Distill-1.5B & $1.8 / 15.2\%$ & $+.081\to+.004$ & none$^{\S}$ & no (.91)            & no           & $[9,\infty)$  \\
32k token limit ($n{=}64$)     & R1-Distill-1.5B & $0.4 / 5.6\%$  & $+.076\to-.015$ & $k{=}15$ & \textbf{yes (.0090)}      & \textbf{yes} & $\mathbf{[9,33]}$ \\
\midrule
\multicolumn{8}{@{}l@{\extracolsep{0pt}}}{\emph{Public RLVR checkpoints at 3k--16k on the 1,000-prompt mixture}} \\
ProRL-1.5B (16k, $256$)                & R1-Distill-1.5B & $0.1 / 15.5\%$ & $+.141\to-.009$ & $k{=}56$ & no (.20)                  & no        & $[24,\infty)$  \\
LUFFY-1.5B (3k, $128$)                 & Qwen2.5-Math    & $4.1 / 12.8\%$ & $+.118\to+.025$ & none     & no (${>}.99$)             & no        & $[109,\infty)$ \\
ORZ-1.5B (8k, $128$)                   & Qwen2.5-1.5B    & $3.2 / 78.4\%$ & $+.421\to+.294$ & none     & no (${>}.99$)             & no        & $[129,\infty)$ \\
SimpleRL 1.5B (8k, $128$)              & Qwen2.5-1.5B    & $2.1 / 78.4\%$ & $+.262\to+.381$ & none     & no (${>}.99$)             & no        & $[129,\infty)$ \\
SimpleRL 7B (8k, $128$)$^{\dagger}$    & Qwen2.5-7B      & $0.8 / 1.9\%$  & $+.145\to-.023$ & $k{=}25$ & unadj.\ (.017)$^{*}$      & yes$^{*}$ & $[13,114]$     \\
SimpleRL 14B (8k, $128$)$^{\dagger}$   & Qwen2.5-14B     & $0.6 / 1.4\%$  & $+.143\to-.017$ & $k{=}34$ & no (.15)                  & no        & $[14,\infty)$  \\
SimpleRL 32B (8k, $128$)$^{\dagger}$   & Qwen2.5-32B     & $0.5 / 1.1\%$  & $+.148\to-.021$ & $k{=}34$ & no (.050)                 & no        & $[16,\infty)$  \\
SimpleRL 8B (8k, $128$)$^{\dagger}$    & Llama-3.1-8B    & $16.1 / 17.2\%$& $+.080\to-.006$ & $k{=}74$ & no (.41)                  & no        & $[30,\infty)$  \\
\bottomrule
\end{tabular*}
\end{table}

\subsection{Empirical Findings on Public Checkpoints}

\paragraph{Finding 1: A visible crossing need not be statistically supported.}
Table~\ref{tab:ladder} shows a pervasive gap between visual sign changes in
$\hat D(k)$ and statistically supported crossovers. At 16k tokens with
$n=256$ generations per prompt, both DeepScaleR and ProRL cross visibly in
point estimate at $k=27$ and $k=56$, yet neither meets the simultaneous
confidence-band criterion; DeepScaleR's unadjusted dominance rejection does
not survive Holm multiplicity adjustment across the discovery family and
fails to replicate on 500 fresh prompts at 16k. The four SimpleRL-Zoo
checkpoints cited by \citet{yue2025limit} exhibit the exact same gap: all
four cross visibly between $k=25$ and $k=74$, but three of the four
(Qwen2.5-14B, Qwen2.5-32B, and Llama-3.1-8B) fail both the dominance test and
the crossing criterion (all $p$-values and intervals in Table~\ref{tab:ladder}).
Only Qwen2.5-7B narrowly clears the per-pair band threshold before
multiplicity adjustment, with its upper band $U(k)$ dipping below zero only
from $k=114$ and by less than $.001$. Meanwhile, on Qwen2.5-1.5B, the same
SimpleRL-Zoo recipe stays ahead of its base across all $k\le128$, because
its base model fails to terminate on $78\%$ of generations (Appendix~\ref{app:exp}).

The same protocol sensitivity appears in code generation on HumanEval+ and
MBPP+ \citep{liu2023evalplus} (Appendix~\ref{app:exp}). At an 8k-token limit,
both DeepScaleR and ProRL improve pass@1 over their base without any
supported crossing or dominance violation. Lowering the limit to 2k
tokens---where truncation climbs to $50\%$ for the base and $61\%$--$82\%$
for the RL checkpoints---causes both RL models to fall significantly
\emph{below} the base at every $k\le64$. Thus a single unreported protocol
parameter can flip both the sign and the statistical significance of a
pass@$k$ comparison.

\paragraph{Finding 2: At 32k tokens on 1,060 new competition problems, two
1.5B models cross while two 7B models do not.}
On the 1,000-prompt mixture at 32k, DeepScaleR's pooled band meets the
crossing criterion, though neither prompt source reaches significance in
isolation under a joint band (Appendix~\ref{app:exp}). To obtain a clean test
free of training-corpus overlap and benchmark-mixture artifacts, the top
block of Table~\ref{tab:longchain} reports the 32k evaluation on the 1,060
deduplicated competition problems. Both 1.5B models exhibit statistically
supported crossovers: DeepScaleR-1.5B gains $+.076$ at $k=1$ and loses
$-.020$ at $k=128$ with certified first-loss interval $\kappa\in[11,61]$, and
ProRL-1.5B gains $+.114$ at $k=1$ and loses $-.021$ at $k=128$ with
$\kappa\in[12,63]$ (both dominance $p<.003$; Table~\ref{tab:longchain}). In
both 1.5B pairs, residual truncation falls almost entirely on the base model
($7.8\%$ versus $\le0.7\%$), which handicaps the base; a sensitivity analysis
in Appendix~\ref{app:prereg} confirms that DeepScaleR's crossover remains
statistically supported even if up to $83\%$ of failed base truncations are
relabeled as successes.

In contrast, neither 7B model shows a supported crossover under the
identical protocol (Table~\ref{tab:longchain}). AceReason-Nemotron-7B stays
ahead of its R1-Distill-Qwen-7B base at every budget $k\le128$, with its
lower confidence bound $L(k)>0$ through $k=33$. Skywork-OR1-7B improves
pass@1 by $+.037$ and converges to its base at large $k$ ($\hat D(128)=-.001$,
dominance $p=.66$), even though Skywork truncates more frequently than its
base and thus favors the base at large $k$. A pass@$k$ crossover is therefore
a property of specific checkpoints and training regimes under a stated
evaluation protocol, not a universal law of long-chain RLVR.

\subsection{Statistical Power and Optimal Evaluation Design}

Why do visible crossings such as ProRL-16k ($k=56$, $p=.20$) and the
SimpleRL-Zoo 14B/32B pairs fail to reach significance even with
$1{,}000\times256$ generations per model? A non-significant test is
informative only alongside its statistical power. Simulating paired counts
from the nonparametric kernels fitted to each model pair in
Section~\ref{sec:kernel}, Table~\ref{tab:power-main} evaluates the power of
both the dominance test and the simultaneous crossing criterion, together
with simultaneous band coverage across $k\le n$.

\begin{table}[t]
\caption{\textbf{Kernel-simulated power, simultaneous band coverage, and
sample-size requirements.} Left block: true kernel differences $D(1),D(n)$,
power of the dominance test and crossing criterion at $\alpha=.05$, and
empirical coverage of the 95\% simultaneous band over $k=1,\dots,n$ and of
the upper bound $U(n)$ at the endpoint $k=n$ ($2{,}000$ simulated data sets
per row at the realized design). Rightmost column: minimum total
generations $m(n+n')$ on the design grid $m\in\{1000,\dots,32000\}\times
n\in\{64,128,256,512\}$ needed to reach $.80$ dominance-test power
(full grid in Appendix~\ref{app:power}).}
\label{tab:power-main}
\centering
\footnotesize
\setlength{\tabcolsep}{0pt}
\begin{tabular*}{\textwidth}{@{\extracolsep{\fill}}lrrcccccl@{}}
\toprule
Model pair & $m$ & $n$ & $D(1)$ & $D(n)$ &
\shortstack{Power ($\alpha{=}.05$)\\dom.\ / cross.} &
\shortstack{Simul.\\cov.} &
\shortstack{Endpt.\\$U(n)$} &
\shortstack{Cheapest design for $.80$\\dominance power} \\
\midrule
DeepScaleR-32k (new) & 1060 & 128 & $+.076$ & $-.019$ & .81 / .72 & .946 & .998 & $0.27$M ($m{=}1060, n{=}128$) \\
DeepScaleR-16k       & 1000 & 256 & $+.083$ & $-.011$ & .63 / .52 & .946 & .996 & $1.0$M ($m{=}4000, n{=}128$)  \\
ProRL-16k            & 1000 & 256 & $+.141$ & $-.009$ & .21 / .14 & .947 & .994 & $4.1$M ($m{=}8000, n{=}256$)  \\
GRPO $G{=}8$         & 1000 & 256 & $+.231$ & $-.004$ & .07 / .04 & .952 & .995 & $16.4$M ($m{=}32000, n{=}256$)\\
\bottomrule
\end{tabular*}
\end{table}

Three design principles follow from Table~\ref{tab:power-main} and the full
design grid in Appendix~\ref{app:power}:
\begin{enumerate}
\item \textbf{Simultaneous bands remain well-calibrated in the small-tail
regime, and multiplier sharing doubles power over Bonferroni.} Even at the
endpoint $k=n$, where $\psi_i(n)\in\{-1,0,1\}$ is discrete and the late-$k$
loss rests on a few percent of prompts \citep{bowyer2025clt}, the 95\%
simultaneous band covers the entire true curve $D(k)$ in $94.4\%$--$95.2\%$
of simulated datasets. Sharing multipliers across $k$ is essential: at the
ProRL-16k design, a Bonferroni correction over pointwise tests has size
$.015$ and power $.11$, whereas the prompt-multiplier test achieves nominal
size $.052$ and doubles power to $.22$.

\item \textbf{Adding prompts ($m$) raises power much more than adding
generations per prompt ($n$).} Let
$\Delta_i(k)=(1-p_i)^k-(1-q_i)^k$ denote prompt $i$'s true paired pass@$k$
difference. By the law of total variance, the variance of $\hat D(k)$
decomposes as
\begin{equation}\label{eq:var-decomp}
\Var\big\{\hat D(k)\big\} = \frac{1}{m}\Big\{\underbrace{\Var\big(\Delta_i(k)\big)}_{\text{between-prompt heterogeneity}}
+ \underbrace{\mathbb E\big[\Var\big(\psi_i(k)\mid p_i,q_i\big)\big]}_{\text{within-prompt binomial noise}}\Big\}.
\end{equation}
Increasing the per-prompt sampling depth $n$ reduces only the second term,
leaving the between-prompt variance $\Var(\Delta_i(k))$ untouched, whereas
increasing the number of prompts $m$ divides both terms by $m$. For
DeepScaleR-16k at the Holm threshold $\alpha=.01$, spending $1.02\times10^6$
total generations on $m=4000$ prompts at $n=128$ yields power $.88$ ($.97$
at $\alpha=.05$), whereas spending the exact same compute budget on
$m=1000$ prompts at $n=512$ yields power only $.49$. Once $n$ is large
enough to cover the largest budget $K$ of interest, additional compute
should go into broader prompt sets rather than deeper rollouts.

\item \textbf{Shallow crossovers require thousands of prompts to resolve.}
At $m=1000$ and $n=256$, ProRL-16k has only $.21$ dominance-test power and
$.14$ crossing-criterion power, making non-rejection the modal outcome;
reaching $.80$ power requires $m=8000$ prompts. For the controlled GRPO arms
of Section~\ref{sec:planned}, power at $m=1000, n=256$ is $.01$--$.07$,
which directly motivates our 3,000-problem deep-$k$ evaluation in
Section~\ref{sec:planned}.
\end{enumerate}

\section{RLVR is a Kernel, Not a Function: Refuting the Single-Curve Sharpening Hypothesis}\label{sec:model}

Part~I (Sections~\ref{sec:infer}--\ref{sec:experiments}) established
\emph{when} a pass@$k$ crossover is statistically real. Part~II
(Sections~\ref{sec:model}--\ref{sec:planned}) asks \emph{why} a crossover
occurs when it is real. We begin by deriving the exact binary-partition
identity of KL-regularized reinforcement learning, which shows from first
principles why post-RL success is generally a Markov kernel $H(q\mid p)$
rather than a deterministic function $q=T(p)$. We then state the universal
marginal laws that govern \emph{any} joint distribution $(P,Q)$, specialize
them to the prevailing single-curve sharpening hypothesis ($q=T(p)$), and
refute determinism directly on the clean 32k benchmark and across all 20
evaluated datasets.

\subsection{The Prevailing Sharpening Hypothesis and Universal Marginal Laws}

Consider KL-regularized reinforcement learning with binary reward
$r(y)\in\{0,1\}$ and reference tempering $1/\tau$, whose optimal policy
takes the Gibbs exponential-tilt form
$\pi^\star(y\mid x)\propto\pi_{\mathrm{base}}(y\mid x)^{1/\tau}e^{r(y)/\lambda}$
\citep{huang2025sharpening,polson2026rlvr,mroueh2025grpo}. Partitioning the
output space on prompt $i$ into correct solutions $C_i$ (where
$\pi_{\mathrm{base}}(C_i)=p_i$ and $\pi^\star(C_i)=q_i$) and incorrect
solutions $C_i^c$, the post-RL log-odds of success satisfy the exact
identity
\begin{equation}\label{eq:operator-exact}
\logit q_i \;=\; \frac{1}{\tau}\,\logit p_i \;+\; h_i \;+\; \frac{1}{\lambda},
\qquad
h_i = \log\frac{M_{C,i}}{M_{C^c,i}},
\end{equation}
where $M_{A,i}=\sum_{y\in A}\pi_{\mathrm{base}}(y\mid A,x_i)^{1/\tau}$ is
the within-cell R\'enyi mass of order $1/\tau$ on $A\in\{C_i,C_i^c\}$.

Equation~\eqref{eq:operator-exact} reveals the exact structural condition
separating a function from a kernel. Physically, $h_i$ measures the prompt's
\emph{solution-to-distractor concentration}: if the base model's incorrect
probability $1-p_i$ on prompt $i$ is concentrated on a single seductive trap
answer ($M_{C^c,i}$ large, making $h_i$ strongly negative), exponential
tilting amplifies that dominant distractor and drives the prompt toward a
very different post-RL outcome than a prompt of the exact same base success
rate $p_i$ whose errors are dispersed across many trajectories---directly
anticipating the modal wrong-answer collapse documented in
Section~\ref{sec:kernel}. Because $h_i$ varies across prompts of equal $p_i$,
prompts at the same base difficulty map to a \emph{conditional distribution}
$H(q\mid p)$. Only when $h_i$ is constant across prompts of equal $p$ can it
be absorbed into an intercept $\gamma=\bar h+1/\lambda$. Writing
$\beta=1/\tau$ then yields the affine-logit single-curve model
$q=T_{\beta,\gamma}(p)$ defined by
\begin{equation}\label{eq:operator}
\logit q \;=\; \beta\,\logit p + \gamma, \qquad
\pstar = \sigma\!\Big(-\frac{\gamma}{\beta-1}\Big), \qquad
\sigma(u) = \frac{1}{1+e^{-u}},
\end{equation}
where $\pstar$ ($\beta\ne1$) is the \emph{break-even difficulty}: for
$\beta>1$, prompts easier than $\pstar$ ($p>\pstar$) gain success probability
($T(p)>p$), whereas prompts harder than $\pstar$ ($p<\pstar$) lose success
probability ($T(p)<p$). To allow for prompts that drift toward zero success
when all $G$ rollouts in a group fail (which occurs with probability
$(1-p_i)^G$), the working model augments $T_{\beta,\gamma}$ with an
operational excess-zero indicator $Z_i\sim\mathrm{Bern}(1-\rho(p_i))$:
\begin{equation}\label{eq:rho}
\rho(p) = \pi_0(1-p)^G, \qquad q_i = Z_i\,T_{\beta,\gamma}(p_i),
\end{equation}
writing $G_0(k)=\mathbb E_F[(1-P)^k]$ and
$G_1(k)=\mathbb E_F[\rho(P)+(1-\rho(P))(1-T_{\beta,\gamma}(P))^k]$ so that
$D(k)=G_0(k)-G_1(k)$. In Appendix~\ref{app:model-ident}
(Theorem~\ref{thm:ident}), we prove that $(F,\beta,\gamma,\pi_0)$ is
nonparametrically identified from paired counts whenever $n_i'\ge2$ (and
show why $n_i'\ge2$ is sharp), and we detail the 200-point Grid EM algorithm
that fits $(F,\beta,\gamma,\pi_0)$ without two-stage errors-in-variables
attenuation (Appendix~\ref{app:em}).

\paragraph{Universal marginal laws versus single-curve predictions.}
Before specializing to $q=T(p)$, note that for \emph{any} joint distribution
of $(P,Q)$---whether generated by a deterministic function or a general
Markov kernel $H(q\mid p)$---integration by parts expresses $D(k)$ solely
through the marginal CDFs $F_P$ and $F_Q$:
\begin{equation}\label{eq:marginal}
D(k) = k\int_0^1 (1-u)^{k-1}\big\{F_P(u) - F_Q(u)\big\}\,du.
\end{equation}
Because the kernel $K(k,u)=(1-u)^{k-1}$ acts as a smooth, totally positive
filter on $(0,\infty)\times(0,1)$ that shifts its weight monotonically toward
$u=0$ as $k$ grows, $D(k)$ cannot oscillate over $k$ more times than the
underlying difficulty CDFs $F_P$ and $F_Q$ cross each other. Formally,
Karlin's variation-diminishing theorem \citep{karlin1968total} yields a
universal sign-change bound for any Markov kernel (proved in
Appendix~\ref{app:marginal} and noted concurrently by
\citealp{liu2026nofreelunch} for shared strategy spaces):

\begin{proposition}[Universal marginal sign-change bound]\label{prop:signs}
For any joint law of $(P,Q)$ with $F_P\ne F_Q$, if $F_P-F_Q$ changes sign
at most $r$ times on $(0,1)$, then $D(k)$ changes sign at most $r$ times
over $k>0$. Moreover, by an Abelian theorem on \eqref{eq:marginal}, the
eventual sign of $D(k)$ as $k\to\infty$ is governed by the lower tails of
$F_P(u)$ and $F_Q(u)$ as $u\downarrow0$.
\end{proposition}

Specializing Proposition~\ref{prop:signs} to a single curve $q=T(p)$ (or a
band-supported kernel) yields explicit power-law tail exponents, crossover
scales, and uniqueness guarantees (proved in
Appendices~\ref{app:model-kstar}--\ref{app:marginal}):

\begin{theorem}[Tail exponent and asymptotic crossover scale]\label{thm:crossover}\label{cor:kstar}
Suppose $F$ has hard-tail density $f(p)\sim c\,p^{a-1}$ as $p\to0^+$ ($a,c>0$),
$\pi_0=0$, and $T(p)\sim C\,p^{\beta_{\mathrm{tail}}}$ as $p\to0^+$ ($C,\beta_{\mathrm{tail}}>0$,
with $\inf_{p\in[\delta,1]}T(p)>0$ for all $\delta>0$; the affine-logit link
\eqref{eq:operator} has $\beta_{\mathrm{tail}}=\beta$ and $C=e^\gamma$). Then
as $k\to\infty$,
\begin{equation}\label{eq:tails}
G_0(k) \sim c\,\Gamma(a)\,k^{-a}, \qquad
G_1(k) \sim \frac{c\,\Gamma(a/\beta_{\mathrm{tail}})}{\beta_{\mathrm{tail}}}\,(Ck)^{-a/\beta_{\mathrm{tail}}},
\end{equation}
so RLVR maps the failure power-law exponent $a\mapsto a/\beta_{\mathrm{tail}}$,
and $\beta_{\mathrm{tail}}>1$ with $D(k_0)>0$ at some finite $k_0$ implies a
crossover (the same exponent holds for any kernel supported on
$[c_1p^{\beta_{\mathrm{tail}}},c_2p^{\beta_{\mathrm{tail}}}]$). Under
$F=\mathrm{Beta}(a,b)$ and the affine-logit link with $\beta>1$, as
$\pstar\to0$ ($\gamma\to\infty$ at fixed $a,b,\beta$), the crossover budget
$\kstar=\inf\{k\ge1:D(k)<0\}$ satisfies
\begin{equation}\label{eq:kstar}
\log\kstar = \frac{\gamma}{\beta-1}
+ \frac{\beta}{a(\beta-1)}\log\frac{\beta\,\Gamma(a)}{\Gamma(a/\beta)}
+ O(\pstar)
= \log\frac{1-\pstar}{\pstar} + O(1).
\end{equation}
If instead $\pi_0>0$, then $G_1(k)\downarrow\pi_0\,\mathbb E[(1-P)^G]>0$
while $G_0(k)\to0$, inducing an atom-driven crossover at scale
$\kstar_0=\big(c\Gamma(a)/\{\pi_0\mathbb E[(1-P)^G]\}\big)^{1/a}$.
\end{theorem}

\begin{theorem}[Single crossing under a single-fixed-point curve]\label{thm:single}
Let $\pi_0=0$ and suppose $T:[0,1]\to[0,1]$ has a single interior fixed point
$\pstar\in(0,1)$, is side-preserving ($T([0,\pstar])\subseteq[0,\pstar]$ and
$T([\pstar,1])\subseteq[\pstar,1]$), and crosses the diagonal in one
direction ($T(p)\le p$ on $[0,\pstar]$ and $T(p)\ge p$ on $[\pstar,1]$, or
both reversed), without requiring $T$ to be monotone. Then $D(k)$ changes
sign at most once over $k\ge1$.
\end{theorem}

\paragraph{Empirical cracks in the single-curve working model.}
Theorems~\ref{thm:crossover}--\ref{thm:single} make the single-curve
sharpening hypothesis quantitative: if $\beta_{\mathrm{tail}}>1$, every hard
prompt below $\pstar$ loses success probability ($T(p)<p$), forcing a unique
crossover at $\kstar\asymp 1/\pstar$. When we fit the single-curve working
model to the clean 32k competition benchmark and the 16 public and controlled
settings via Grid EM, the apparent slope $\hat\beta_{\mathrm{proj}}$ exceeds
unity on nearly every pair ($1.24$ on DeepScaleR-32k and $1.81$ on ProRL-32k;
full confidence intervals in Table~\ref{tab:pairs} of Appendix~\ref{app:em}).
Yet three diagnostics immediately signal that the single-curve model is
misspecified: (1) a likelihood-ratio test against a cubic spline link rejects
the affine link~\eqref{eq:operator} on both 32k competition pairs and on 14
of 15 other pairs ($p=.005$; Appendix~\ref{app:em}); (2) the fitted single
curves systematically predict crossovers two to three times earlier than
observed (e.g., $\kstar=11$ versus $k=25$ on ProRL-32k); and (3) simulating
from ProRL-16k's fitted single curve predicts $.98$ dominance-test power,
contradicting the non-significant empirical test in Table~\ref{tab:ladder}.
As the proof of Theorem~\ref{thm:ident} anticipates, all three anomalies are
signatures of unmodeled within-$p$ dispersion $\Var(Q\mid P=p)>0$, which
inflates the second factorial moment $\mathbb E[Q^2\mid P=p]$ and forces a
deterministic fit to steepen $\hat\beta_{\mathrm{proj}}$ and inflate
$\hat\pi_0$.

\subsection{Why Conditional Dispersion Matters for \texorpdfstring{pass@$k$}{pass@k}}

Fix a base model and a post-RL checkpoint. A Markov kernel $H(q\mid p)$
specifies the conditional probability distribution of post-RL success rate
$Q$ across prompts that share the exact same base success rate $P=p$. This
dispersion reflects deterministic heterogeneity \emph{across prompts} at a
fixed $(p,\text{checkpoint})$ pair---not randomness across training seeds
or binomial variance across individual rollouts.

\begin{figure}[t]
\centering
\includegraphics[width=\textwidth]{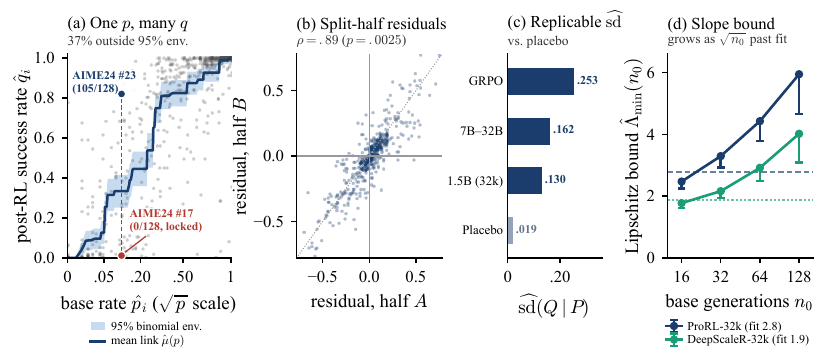}
\caption{\textbf{Prompts with equal base success respond differently to
RLVR on the clean 32k competition benchmark, refuting any single response
curve} (ProRL-1.5B at 32k tokens on 1,060 new competition problems,
$n=n'=128$; all 20 datasets appear in Tables~\ref{tab:split}
and~\ref{tab:slope} of Appendix~\ref{app:het}). \emph{(a)} Empirical success
rates $(\hat p_i,\hat q_i)$ plotted on a $\sqrt{p}$ scale against the
cross-fitted monotone mean link $\hat\mu(p)$, with $38\%$ of prompts falling
outside the 95\% binomial sampling envelope ($5\%$ expected under a single
curve). Two AIME~2024 problems with identical base rate $\hat p=14/128=.109$
diverge to $105/128$ ($.820$) versus $0/128$ ($0$, locked). \emph{(b)}
Half-sample post-RL residuals $r_{A,i}$ versus $r_{B,i}$ from disjoint
64-generation halves ($\rho=.89, p=.0025$). \emph{(c)} Replicable conditional
standard deviation $\widehat{\mathrm{sd}}(Q\mid P)$ across checkpoint families
versus the base-split placebo ($.019$). \emph{(d)} Distribution-free lower
bound $\hat\Lambda_{\min}(n_0)$ on the Lipschitz constant of any single curve
$Q=T(P)$ (Theorem~\ref{thm:slope}) across thinned base depths
$n_0\in\{16,32,64,128\}$ on ProRL-32k and DeepScaleR-32k, growing as
$\sqrt{n_0}$ past each model's fitted max slope.}
\label{fig:kernel}
\end{figure}

If we only care about pass@1, the kernel's conditional mean
$T(p)=\mathbb E[Q\mid P=p]$ suffices, because
$\mathbb E[Q]=\mathbb E[T(P)]$. For any sampling budget $k\ge2$, however,
the failure function $q\mapsto(1-q)^k$ is strictly convex on $[0,1]$. By
Jensen's inequality, whenever $\Var(Q\mid P=p)>0$,
\begin{equation}\label{eq:jensen}
1 - \mathbb E\big[(1-Q)^k \mid P=p\big]
\;<\;
1 - \big(1 - \mathbb E[Q\mid P=p]\big)^k
= 1 - \big(1 - T(p)\big)^k.
\end{equation}
Thus conditional dispersion at fixed $p$ strictly reduces $\passk^{\mathrm{RL}}$
relative to a deterministic curve with the same mean response $T(p)$, and
the penalty grows with $k$ as $(1-q)^k$ concentrates its weight on values of
$q$ near zero. On the clean 32k competition benchmark (Figure~\ref{fig:kernel}a),
the two AIME~2024 prompts noted in Section~\ref{sec:intro} both have
$14/128$ base successes ($\hat p=.109$), so at $k=128$ both have base
$\widehat{\passk}_i=1$; post-RL under ProRL-32k, one prompt rises to
$105/128$ successes ($\hat q=.820$) and contributes $0$ to $\hat D(128)$,
whereas the other drops to $0/128$ ($\hat q=0$) and contributes $-1/m$,
directly pulling the tail of $\hat D(k)$ downward.

\subsection{Refuting Determinism: Split-Generation Residual Covariance and Lipschitz Slope Bounds}

To test $H_0:\Var(Q\mid P)=0$ almost surely without relying on a parametric
link, we combine two complementary nonparametric tests:

\paragraph{1. The split-generation residual covariance test.}
Partition each prompt's post-RL generations into two disjoint halves $A$ and
$B$ (with sample success rates $\hat q_{A,i},\hat q_{B,i}$) and its base
generations into two disjoint halves $1$ and $2$ (with sample success rates
$\hat p_i^{(1)},\hat p_i^{(2)}$). Using five-fold cross-fitting across
prompts, we fit an isotonic regression $\hat\mu^{(-i)}$ of $\hat q$ on
$\hat p$ using the out-of-fold prompts and form the half-sample residual
product:
\begin{equation}\label{eq:split-stat}
r_{A,i} = \hat q_{A,i} - \hat\mu^{(-i)}\!\big(\hat p_i^{(1)}\big), \qquad
r_{B,i} = \hat q_{B,i} - \hat\mu^{(-i)}\!\big(\hat p_i^{(2)}\big), \qquad
S = \frac{1}{m}\sum_{i=1}^m r_{A,i}\,r_{B,i}.
\end{equation}
Intuitively, if a single curve $q=T(p)$ governed all prompts, the residuals
$r_{A,i}$ and $r_{B,i}$ on disjoint generation halves would reflect only
independent binomial sampling noise around that curve and their product would
average to zero; a positive cross-product $S>0$ directly measures
reproducible prompt-specific departure from the curve. Formally, because
$(\hat p_i^{(1)},\hat q_{A,i})$ and $(\hat p_i^{(2)},\hat q_{B,i})$ use
disjoint generations, their binomial sampling errors are conditionally
independent given $(P_i,Q_i)$ and $\hat\mu^{(-i)}$. Writing
$\bar\mu(p)=\mathbb E[\hat\mu^{(-i)}(\hat p_i^{(1)})\mid P_i=p,\hat\mu^{(-i)}]$,
the conditional expectation of the product decomposes exactly as
\begin{equation}\label{eq:split-decomp}
\mathbb E\big[r_{A,i}\,r_{B,i} \mid \hat\mu^{(-i)}\big]
\;=\; \underbrace{\mathbb E\big[\Var(Q\mid P)\big]}_{\text{conditional dispersion}}
\;+\; \underbrace{\mathbb E\Big[\big(T(P)-\bar\mu(P)\big)^2 \,\Big|\, \hat\mu^{(-i)}\Big]}_{\text{mean-link attenuation bias }\bar S_{\mathrm{null}}\ge0}.
\end{equation}
Calibrating $S$ against parametric bootstrap datasets under both a flexible
deterministic-curve null ($\pi_0=0$) and a zero-inflated spline null
($\pi_0\ge0$) eliminates the smoothing bias $\bar S_{\mathrm{null}}$ and
yields the bias-corrected conditional dispersion estimate
$\widehat{\mathrm{sd}}=\sqrt{\max(0,S-\bar S_{\mathrm{null}})}$
(Appendix~\ref{app:het}). Across all 20 evaluated datasets---spanning all
four 32k competition pairs, the 3k--16k public checkpoints, the controlled
GRPO and SFT arms, code generation on HumanEval+/MBPP+, and two non-Qwen base
models (Llama-3.1-8B and DeepSeek-Math-7B)---every analysis rejects both
single-curve nulls at the bootstrap floor ($p\le.0025$), whereas the
within-model base placebo stays near zero (Figure~\ref{fig:kernel}b,c; full
test statistics in Tables~\ref{tab:split} and~\ref{tab:nonqwen} of
Appendix~\ref{app:het}).

\paragraph{2. Distribution-free Lipschitz slope bound.}
Could the vertical spread in Figure~\ref{fig:kernel}(a--c) be explained by a
single deterministic function $q=T(p)$ with sharp local kinks across each
binomial resolution window $P_i=c/n\pm O(1/\sqrt{n})$? Theorem~\ref{thm:slope}
(proved in Appendix~\ref{app:slope}) bounds how steep any single curve
$Q=T(P)$ would have to be to generate the observed within-count variance
$A^\star_n=\mathbb E[\Var(Q\mid c)]$:

\begin{theorem}[Distribution-free Lipschitz slope bound]\label{thm:slope}
Let $n,n'\ge2$. Both $A^\star_n=\mathbb E[\Var(Q\mid c)]$ and
$\mathbb E[P(1-P)]$ are nonparametrically identified from the joint law of
$(c,d)$. If the paired count law is generated by a deterministic curve
$Q=T(P)$ that is $\Lambda$-Lipschitz on $[0,1]$, then
\begin{equation}\label{eq:slope-bound}
\Lambda \;\ge\; \Lambda_{\min}(n)
\;:=\; \left(\frac{n\,A^\star_n}{\mathbb E[P(1-P)]}\right)^{1/2}.
\end{equation}
\end{theorem}

Under a single $\Lambda$-Lipschitz curve, $A^\star_{n_0}=O(1/n_0)$ as the
base count is thinned to depth $n_0\le n$, so $\Lambda_{\min}(n_0)$ remains
bounded; under genuine conditional dispersion $\mathbb E[\Var(Q\mid P)]>0$,
$A^\star_{n_0}\to\mathbb E[\Var(Q\mid P)]>0$ and $\Lambda_{\min}(n_0)$ grows
like $\sqrt{n_0}$. Figure~\ref{fig:kernel}(d) shows that the data follow the
$\sqrt{n_0}$ scaling on both 32k competition pairs, reaching
$\hat\Lambda_{\min}(128)=5.92$ on ProRL-32k and $4.02$ on DeepScaleR-32k---more
than double each pair's steepest fitted slope ($2.78$ and $1.88$)---and across
19 of the 20 datasets (full bounds and bootstrap intervals in
Table~\ref{tab:slope}, Appendix~\ref{app:het-nonqwen}).

\section{What Drives the Crossover: The 2\%--3\% Locked Minority, Not Uniform Hard-Tail Degradation}\label{sec:kernel}

\subsection{Resolving the Slope Illusion: Conditional-Mean Slopes Below One and the Mean-Collapse Counterfactual}

To estimate the Markov kernel $H(q\mid p)$ directly from the paired counts,
we use two complementary models (full equations in Appendix~\ref{app:het}):
(1) a \textbf{2D nonparametric maximum likelihood estimator (2D NPMLE)} that
places joint mixing weights $w_{\ell j}=\Pr(P=p_\ell,Q=q_j)$ on an
$80\times80$ logit grid (including an explicit $q=0$ atom) and maximizes the
exact paired binomial log-likelihood via multiplicative EM; and (2) a
\textbf{Beta-Binomial random-effects kernel} that models the non-zero branch
of Section~\ref{sec:model} as
$Q\mid(P=p,Z=1)\sim\mathrm{Beta}\big(T_{\beta,\gamma}(p)\phi,\,(1-T_{\beta,\gamma}(p))\phi\big)$,
profiling the conditional precision $\phi>0$ ($\phi=\infty$ recovers the
single curve) inside the 200-point Grid EM.

\begin{figure}[t]
\centering
\includegraphics[width=\textwidth]{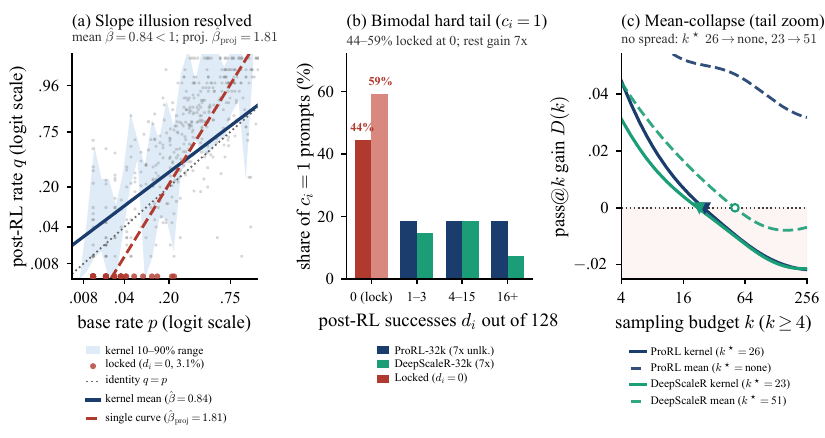}
\caption{\textbf{Once conditional dispersion is modeled, the mean log-odds
slope falls below one, and the crossover is driven by a bimodal split on
hard prompts.} \emph{(a)} Logit-scale resolution of the slope illusion on
ProRL-32k (1,060 new competition problems): the misspecified single-curve
projection ($\hat\beta_{\mathrm{proj}}=1.81>1$, dashed red) is pulled below
the identity line $q=p$ by the locked prompts along the bottom strip, whereas
the Beta-Binomial conditional mean ($\hat\beta=0.84<1$, solid blue) stays
above $q=p$. \emph{(b)} Among hard prompts that the base model solves only
once ($c_i=1$), post-RL outcomes split bimodally: $44\%$--$59\%$ on the 32k
pairs ($30\%$ on GRPO) are locked at $d_i=0$ (red), while the unlocked
majority improves its median success rate by $7.0\times$ on 32k and $3.0\times$
on GRPO. \emph{(c)} Mean-collapse counterfactual on ProRL-32k and
DeepScaleR-32k (tail zoom $k\in[4,256]$, rose region $D(k)<0$): replacing each
kernel slice $\hat H(\cdot\mid p)$ by a point mass at its conditional mean
$\hat T(p)$ eliminates ProRL-32k's crossover ($\kstar:26\to\text{none}$) and
more than doubles DeepScaleR-32k's crossing budget ($\kstar:23\to51$).}
\label{fig:collapse}
\end{figure}

\begin{table}[t]
\caption{\textbf{Kernel refit, mean-collapse counterfactual, and model-free
locked-minority decomposition on the clean 32k benchmark and controlled
GRPO} (complete 10- to 20-pair tables in Tables~\ref{tab:phiprofile},
\ref{tab:mean-main}, and~\ref{tab:lockin-full} of Appendix~\ref{app:het}).
\emph{Panel~A}: single-curve projection slope $\hat\beta_{\mathrm{proj}}$
($\phi=\infty$) versus Beta-Binomial conditional-mean slope $\hat\beta$
($\phi$ profiled) and log-likelihood gain $\Delta\ell$, alongside the 2D NPMLE
kernel crossing $\kstar$ versus the mean-collapsed crossing $\kstar_{\mathrm{mean}}$
and endpoint difference $D(K)$. \emph{Panel~B}: model-free locked prompts
$L=\{i:c_i>0,d_i=0\}$, sample crossing after removing $L$ versus removing the
$|L|$ hardest base-solved prompts blind to $d_i$, and the bimodal split on
$c_i=1$ prompts ($\%$ locked vs.\ median $\hat q/\hat p$ among unlocked).
$^{\dagger}$Band-supported crossover.}
\label{tab:kernel-main}\label{tab:lockin}
\centering
\footnotesize
\setlength{\tabcolsep}{0pt}
\begin{tabular*}{\textwidth}{@{\extracolsep{\fill}}lcccccc@{}}
\toprule
\multicolumn{7}{@{}l}{\textbf{Panel A: Kernel slope refit (Beta-Binomial) and 2D NPMLE mean-collapse counterfactual}} \\
\midrule
& \multicolumn{3}{c}{Beta-Binomial slope refit} &
\multicolumn{3}{c}{2D NPMLE kernel vs.\ mean-collapsed} \\
\cmidrule(lr){2-4}\cmidrule(lr){5-7}
Model pair & $\hat\beta_{\mathrm{proj}}$ ($\phi{=}\infty$) & $\hat\beta$ ($\hat\phi$) & $\Delta\ell$ &
Full $\kstar$ & Mean $\kstar_{\mathrm{mean}}$ & $D(K)$: Full $\to$ Mean \\
\midrule
DeepScaleR-1.5B (32k)     & 1.24 & \textbf{0.96} (8.0) & $+702$  & $23^{\dagger}$ & 51   & $-.020\to-.005$ \\
ProRL-1.5B (32k)          & 1.81 & \textbf{0.84} (2.2) & $+1467$ & $26^{\dagger}$ & none & $-.021\to+.006$ \\
AceReason-7B (32k)        & 1.16 & \textbf{0.91} (7.6) & $+592$  & none           & none & $+.008\to+.018$ \\
Skywork-OR1-7B (32k)      & 1.21 & \textbf{0.95} (8.2) & $+483$  & 16             & none & $-.001\to+.015$ \\
GRPO $G{=}4$ ($n{=}256$)  & 2.08 & \textbf{0.85} (1.4) & $+4761$ & 140            & none & $-.002\to+.026$ \\
GRPO $G{=}16$ ($n{=}256$) & 2.18 & \textbf{0.93} (1.4) & $+4720$ & 168            & none & $-.001\to+.027$ \\
\bottomrule
\end{tabular*}

\vspace{4pt}

\begin{tabular*}{\textwidth}{@{\extracolsep{\fill}}lcccccc@{}}
\toprule
\multicolumn{7}{@{}l}{\textbf{Panel B: Model-free locked minority $L$ ($c_i>0, d_i=0$) and $c_i=1$ bimodal hard-tail split}} \\
\midrule
& \multicolumn{4}{c}{Model-free locked set $L$ removal} &
\multicolumn{2}{c}{$c_i=1$ bimodal split} \\
\cmidrule(lr){2-5}\cmidrule(lr){6-7}
Model pair & $|L|/m$ & $\kstar$ (All) & $\kstar$ ($-L$/Hard) & $\hat D(K)$: All $\to{-}L$ &
Locked & Unlk.\ $\hat q/\hat p$ \\
\midrule
DeepScaleR-1.5B (32k)     & $33/1060$ ($3.1\%$) & $22^{\dagger}$ & none / $18$          & $-.020\to+.012$ & $59\%$ & $7.0\times$ \\
ProRL-1.5B (32k)          & $33/1060$ ($3.1\%$) & $25^{\dagger}$ & none / $17^{\dagger}$& $-.021\to+.011$ & $44\%$ & $7.0\times$ \\
AceReason-7B (32k)        & $14/1060$ ($1.3\%$) & none           & none / none          & $+.008\to+.021$ & $33\%$ & $5.5\times$ \\
Skywork-OR1-7B (32k)      & $22/1060$ ($2.1\%$) & 15             & none / none          & $-.001\to+.020$ & $48\%$ & $4.0\times$ \\
GRPO $G{=}4$ ($n{=}256$)  & $20/1000$ ($2.0\%$) & 117            & none / $91$          & $-.002\to+.018$ & $30\%$ & $3.0\times$ \\
GRPO $G{=}16$ ($n{=}256$) & $18/1000$ ($1.8\%$) & 150            & none / none          & $-.001\to+.017$ & $41\%$ & $2.0\times$ \\
\bottomrule
\end{tabular*}
\end{table}

Figure~\ref{fig:collapse} and Table~\ref{tab:kernel-main} establish three
central results that resolve the single-curve anomalies and overturn the
uniform hard-tail degradation picture:
\begin{enumerate}
\item \textbf{Resolving the slope illusion ($\hat\beta<1$).} Once within-$p$
dispersion is modeled, the conditional-mean logit slope $\hat\beta$ falls
\emph{below one} across every model pair---dropping from $1.24$ to $0.96$ on
DeepScaleR-32k, from $1.81$ to $0.84$ on ProRL-32k, and from $2.08$ to $0.85$
on GRPO $G=4$ (Table~\ref{tab:kernel-main}; all 10 refitted pairs with
confidence intervals appear in Table~\ref{tab:phiprofile},
Appendix~\ref{app:het}). As Figure~\ref{fig:collapse}(a) illustrates, the
apparent slope $\hat\beta_{\mathrm{proj}}>1$ was a statistical illusion
caused by forcing a single curve to pass through both the improving majority
and the locked prompts near $q=0$. Profiling $\phi$ improves in-sample and
held-out generation log-likelihood by hundreds to thousands of nats
(Table~\ref{tab:heldout}) and reconciles ProRL-16k's simulated dominance
power with the empirical test ($.98\to.21$).
\item \textbf{The mean-collapse counterfactual.} Replacing each kernel slice
$\hat H(\cdot\mid p_\ell)$ by a point mass at its conditional mean
$\hat T(p_\ell)=\mathbb E[Q\mid P=p_\ell]$ while holding $\hat F_P$ fixed
completely removes the crossover on ProRL-32k and on every 16k and controlled
pair, and more than doubles the crossing budget on DeepScaleR-32k from
$\kstar=23$ to $51$ (Figure~\ref{fig:collapse}c and Table~\ref{tab:mean-main}).
Even on the hardest base nodes where the $3.1\%$ locked prompts pull the
unconditional grid mean $\hat T(p_\ell)$ slightly below $p_\ell$, collapsing
within-node dispersion eliminates or delays the crossover.
\item \textbf{Bimodal divergence on $c_i=1$ prompts.} Among the $27$
competition problems at 32k that the base model solves exactly once in $128$
draws ($c_i=1$), a deterministic sharpening curve $T(p)<p$ would depress all
$27$ prompts uniformly. Instead, the unlocked prompts improve their median
success rate by $7.0\times$ the base rate, while the remaining $59\%$
(DeepScaleR) and $44\%$ (ProRL) drop to $d_i=0$ (Figure~\ref{fig:collapse}b).
\end{enumerate}

\subsection{Forensic Anatomy of the 2\%--3\% Locked Minority and Prefix Rescue}

We do not need any parametric model to verify that this small locked
minority drives the entire late-$k$ loss. At the maximum budget $k=n=n'$,
the unbiased per-prompt estimator $\widehat{\passk}_i$ reduces to the
indicator $\mathbb 1\{c_i>0\}$, yielding the exact endpoint accounting
identity
\begin{equation}\label{eq:lockin-id}
\hat D(n) = \frac{|R| - |L|}{m}, \qquad
L = \big\{i : c_i > 0,\ d_i = 0\big\}, \qquad
R = \big\{i : c_i = 0,\ d_i > 0\big\},
\end{equation}
where $L$ is the set of \emph{locked prompts} (solved by the base model in
$n$ draws, but never by the RL model) and $R$ is the set of newly solved
prompts. Notice how \eqref{eq:lockin-id} immediately explains the sample-size
power law in Section~\ref{sec:experiments}: because $\hat D(n)=(|R|-|L|)/m$
is a difference of small prompt proportions, its standard error scales as
$1/\sqrt{m}$ with the number of prompts rather than shrinking indefinitely
with $n$.

\begin{figure}[t]
\centering
\includegraphics[width=\textwidth]{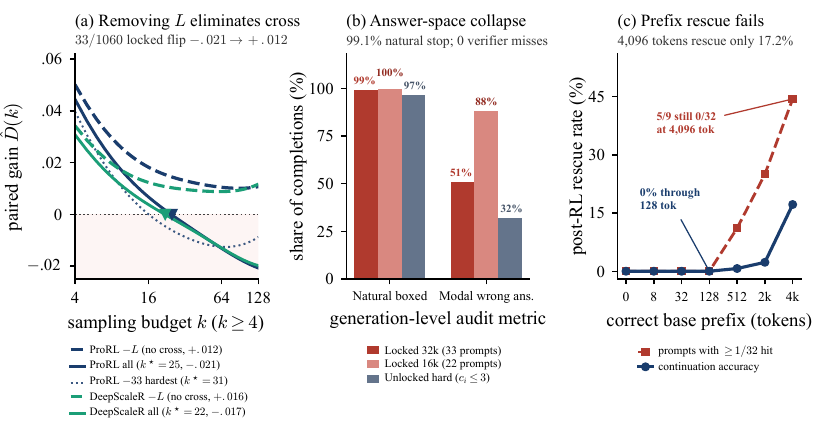}
\caption{\textbf{A $2\%$--$3\%$ locked minority drives the 32k crossovers
via repeated wrong-answer collapse rather than truncation, and resists
prefix rescue up to 4,096 tokens.} \emph{(a)} Tail zoom ($k\in[4,128]$, rose
region $\hat D(k)<0$) on the 1,060 new competition problems at 32k: removing
the $33$ locked prompts ($3.1\%$ of prompts, dashed curves) eliminates the
crossover on both ProRL-32k and DeepScaleR-32k and flips $\hat D(128)$ from
$-.021$ and $-.020$ to $+.011$ and $+.012$, whereas removing the $33$ hardest
base-solved prompts chosen blind to $d_i$ (dotted) overlaps with $L$ on fewer
than half the prompts and leaves the crossings intact. \emph{(b)}
Generation-level forensic audit of the locked prompts on DeepScaleR-32k,
ProRL-32k, and ProRL-16k versus difficulty-matched unlocked hard controls:
$99.1\%$--$99.9\%$ of locked generations terminate naturally with a
well-formed \texttt{\textbackslash boxed\{\}} answer ($0$ verifier misses
across $4{,}224$ completions), while $51\%$--$88\%$ collapse onto a single
modal wrong answer (versus $32\%$ on unlocked hard prompts). \emph{(c)}
Conditioning ProRL on the first $\{8,32,128,512,2048,4096\}$ tokens of the
base model's verified correct chain of thought on persistent locked prompts
recovers $0\%$ through $128$ tokens ($0/288$), $0.7\%$ at $512$ tokens
($2/288$), $2.3\%$ at $2{,}048$ tokens ($6/256$), and only $17.2\%$ at
$4{,}096$ tokens ($44/256$, with $5/9$ prompts remaining at $0/32$).}
\label{fig:forensic}
\end{figure}

Figure~\ref{fig:forensic} and Table~\ref{tab:kernel-main} (with extended
difficulty-matched removal controls in Table~\ref{tab:lockin-full},
Appendix~\ref{app:lockin}) establish three forensic facts about the locked
set $L$:
\begin{enumerate}
\item \textbf{Removing the $2\%$--$3\%$ locked minority eliminates every
crossover.} Across all crossing pairs, $L$ comprises only $1.8\%$--$3.1\%$ of
prompts ($33/1060$ on both DeepScaleR-32k and ProRL-32k). Removing $L$
eliminates the crossover on every pair and flips $\hat D(128)$ on the 32k
competition benchmark from $-.020$ and $-.021$ to $+.012$ and $+.011$
(Figure~\ref{fig:forensic}a). Removing an equal number of the hardest
base-solved prompts chosen blind to $d_i$---or removing difficulty-matched
unlocked prompts with the exact same base counts $c_i$
(Table~\ref{tab:lockin-full})---leaves the crossings intact.
\item \textbf{Natural termination and modal wrong-answer collapse.}
Auditing all $4{,}224$ post-RL completions on the $33$ locked prompts of
DeepScaleR-32k and ProRL-32k alongside ProRL-16k's $22$ locked prompts rules
out token-limit truncation, formatting failures, or verifier false negatives:
over $99\%$ of locked generations across all three evaluations terminate
naturally with a parsed \texttt{\textbackslash boxed\{...\}} expression, and
manual plus symbolic cross-verification finds zero uncredited correct answers
(Figure~\ref{fig:forensic}b). Instead, the RL model collapses onto a
\emph{repeated incorrect answer}: a single modal wrong answer accounts for
$51\%$--$58\%$ of completions on the 32k pairs and $88\%$ on ProRL-16k---versus
$32\%$ on difficulty-matched unlocked prompts---and in half the cases
coincides with the base model's own most frequent distractor. Even under
$4{,}352$ draws per prompt (including $2{,}048$ draws at $\text{top-}p=1.0$),
$13$ of ProRL-16k's $22$ locked prompts remain at $0/4{,}352$ successes.
\item \textbf{4,096-token prefix rescue shows trajectory-wide lock-in.}
To test whether wrong-answer lock-in is decided at the opening tokens, we
conditioned ProRL on the first $\{8,32,128,512,2048,4096\}$ tokens of the
base model's verified correct chain of thought on the nine persistent locked
prompts that yielded a complete correct base trace in a fresh harvest batch
($32$ rollouts per prompt and prefix length; Figure~\ref{fig:forensic}c).
Prefixes up to $128$ correct tokens yield $0\%$ success ($0/288$), $512$ and
$2{,}048$ tokens yield only $0.7\%$ and $2.3\%$, and even $4{,}096$ correct
tokens---nearly the entire solution---rescue only $17.2\%$ of rollouts
($44/256$, leaving $5$ of the $9$ prompts at $0/32$; per-prompt breakdown in
Table~\ref{tab:case-study}, Appendix~\ref{app:lockin}). Verbatim reasoning
traces on locked competition problems (Table~\ref{tab:case-study}) reveal the
mechanism behind this trajectory-wide basin of attraction: even after the RLVR
model explicitly identifies the subtle mathematical step required for the true
answer (such as overlapping corner regions at $(3/5,0,0)$ because
$3/5+3/5=6/5>1$ on \texttt{olympiadbench-521}), it overrides its own deduction
via late-stage self-doubt (\emph{``Wait, perhaps another way\ldots\ perhaps the
initial calculation is correct''}) and collapses back onto the naive distractor
(\texttt{\textbackslash boxed\{89/125\}} in $125/128$ rollouts, versus the base
model's verified inclusion--exclusion answer
\texttt{\textbackslash boxed\{179/250\}}).
\end{enumerate}

\section{Controlled Training Modes and \texorpdfstring{Deep-$k$}{Deep-k} Extrapolation to \texorpdfstring{$k=4096$}{k=4096}}\label{sec:planned}

Public checkpoints differ simultaneously in base model, training corpus,
update count, and objective. To isolate how the post-training paradigm
shapes the response kernel $H(q\mid p)$ and large-$k$ behavior, we trained
controlled models from a single base checkpoint and evaluated them out to
$k=4096$ on 3,000 new problems.

\subsection{Controlled Comparison of GRPO and SFT from a Single Base}

Starting from Qwen2.5-Math-1.5B \citep{yang2024qwen25math} and a fixed
pool of 8,000 problems from the DeepScaleR training corpus, we trained five
arms (full hyperparameters in Appendix~\ref{app:exp}): four on-policy GRPO
arms \citep{shao2024deepseekmath} for 100 updates---varying the rollout
group size $G\in\{4,8,16\}$ at KL coefficient $\lambda=10^{-3}$, plus a
weak-KL arm ($G=4, \lambda=10^{-4}$)---and a supervised fine-tuning (SFT)
arm trained for 3 epochs (132 updates) on the 1,405 problems from the same
pool that carry reference solutions. Evaluating the base model and all five
trained arms at a 3,072-token limit ($n=n'=256$) establishes two controlled
contrasts:
\begin{enumerate}
\item \textbf{Universal dispersion, opposite answer-space concentration.}
Every controlled arm---both GRPO and SFT---strongly rejects a single
response curve in both the split-generation test and the Lipschitz slope
bound (Tables~\ref{tab:split} and~\ref{tab:slope}). In answer space, however,
the two modes move in opposite directions (Appendix~\ref{app:exp}): GRPO
collapses the mean number of distinct extracted answers per prompt from
$18.2$ (base) to $4.0$--$4.4$ and raises the modal answer share from $51\%$
to $83\%$--$85\%$, amplifying the base model's dominant plausible distractor
whenever a rollout group contains no positive reward, whereas SFT preserves
twice the answer diversity ($8.6$ distinct answers, $69\%$ modal share) while
lifting verified solutions.
\item \textbf{Steeper central response under GRPO than SFT, with
insensitivity to $G$ and KL strength.} Fitting a flexible monotone Bernstein
link inside the Grid EM shows that GRPO has a much steeper central logit
slope than SFT across the 20th--80th difficulty percentiles
($\bar\beta_{20:80}=1.88$--$1.99$ for GRPO versus $1.39$ for SFT, $p<.002$;
Figure~\ref{fig:shape} and Table~\ref{tab:pairs} in Appendix~\ref{app:em}),
whereas varying the rollout group size $G\in\{4,8,16\}$ or weakening the KL
penalty tenfold alters the GRPO slope by at most $.11$.
\end{enumerate}

\subsection{Direct Measurement to \texorpdfstring{$k=4096$}{k=4096} on 3,000 New Problems}

On the 1,000 evaluation prompts at $n=n'=256$, all four GRPO arms turn
negative in point estimate beyond $k\approx128$ while SFT stays positive,
yet none of the GRPO crossings reaches statistical significance ($p\ge.20$),
matching the low power ($\le.07$) predicted by Table~\ref{tab:power-main}. To
resolve the large-$k$ comparison with high power and test whether kernels
fitted at $n=256$ can predict behavior an order of magnitude deeper, we
conducted a two-stage evaluation out to $k=4096$ on $3{,}000$ new problems
(1,500 from the DeepScaleR corpus and 1,500 from the MATH test set, strictly
deduplicated against all training, validation, and SFT sets and all other
prompts in this paper; Appendix~\ref{app:deepk}): in \textbf{Stage~1}, we
sampled $n=256$ generations per model on all $3{,}000$ problems and recorded
the fitted 2D NPMLE kernels before drawing any deep rollouts; in
\textbf{Stage~2}, on the $700$ hard-tail problems with at most $8$ successes
in $256$ under at least one model, we drew $4{,}096$ fresh generations per
model (setting the contribution of the $2{,}300$ easy problems with
$\ge9/256$ successes on all models to zero for $k>256$, which incurs a
Jeffreys-bounded error below $1.9\times10^{-5}$; Appendix~\ref{app:deepk}).

\begin{figure}[t]
\centering
\includegraphics[width=\textwidth]{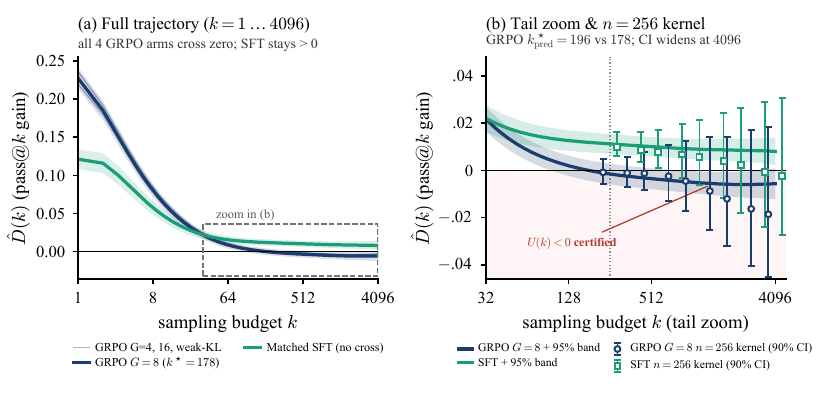}
\caption{\textbf{Measured directly to $k=4096$ on 3,000 new problems, all
four GRPO arms fall behind the base model and stay behind, whereas SFT stays
ahead.} Paired difference curves
$\hat D(k)=\passk^{\mathrm{trained}}-\passk^{\mathrm{base}}$ against
Qwen2.5-Math-1.5B on 3,000 deduplicated problems. \emph{(a)} Full trajectory
across $k=1,\dots,4096$ on a log scale, showing the large early gains at
$k=1$ ($+.23$ for GRPO $G=8$ and $+.12$ for SFT) and convergence toward zero
for $k\ge32$; all four GRPO arms ($G=8$ bold blue; $G=4,16,\text{weak KL}$
thin blue) cross below zero near $k\approx200$. \emph{(b)} Tail zoom on
$k\in[32,4096]$ with 95\% simultaneous confidence bands ($B=4000$) and
out-of-window predictions (open circles for GRPO $G=8$ and open squares for
SFT, horizontally dodged with 90\% bootstrap intervals) from the 2D NPMLE
kernel fitted strictly to the $n=256$ Stage~1 counts (dotted vertical line)
before Stage~2 sampling.}
\label{fig:deepk}
\end{figure}

\paragraph{Measured to $k=4096$, all four GRPO arms fall behind the base and
stay behind, whereas SFT remains significantly ahead.}
Figure~\ref{fig:deepk} and Table~\ref{tab:deepk} establish a clean
separation between on-policy GRPO and SFT at large sampling budgets:
\begin{itemize}
\item \textbf{GRPO crosses near $k\approx200$ and never recovers:} All four
GRPO arms cross below the base model between $k=178$ and $k=234$ and stay
negative all the way to $k=4096$, trailing the base model by $10$ to $17$
problems out of $3{,}000$ at $k=4096$. On the $G=8$ arm, the 95\%
simultaneous confidence band certifies the crossover with first-loss interval
$\kappa\in[95,791]$ and dominance $p=.008$, and the $G=4$ and $G=16$ arms
also reject dominance at $\alpha=.05$ (Table~\ref{tab:deepk}).
\item \textbf{SFT stays ahead through $k=4096$:} SFT never crosses the base
model at any budget $k\le4096$. At $k=4096$, SFT leads the base by $24$ of
$3{,}000$ problems ($\hat D(4096)=+.0080$), and its 95\% simultaneous lower
bound remains strictly positive across the entire range ($L(4096)=+.0023>0$).
\end{itemize}

\begin{table}[t]
\caption{\textbf{Controlled training arms versus Qwen2.5-Math-1.5B on 3,000
new problems, measured to $k=4096$.} Columns 2--6 report direct paired
inference over $k=1,\dots,4096$: the first integer $k$ where $\hat D(k)<0$,
the simultaneous first-loss interval $\kappa\in[\underline\kappa,\overline\kappa]$,
the dominance-test $p$-value, the endpoint difference $\hat D(4096)$, and
the 95\% simultaneous band $[L(4096),U(4096)]$. Columns 7--8 report the
crossover $\kstar$ and endpoint prediction $D(4096)$ (with 90\% bootstrap
interval) of the 2D NPMLE kernel fitted only to the $n=256$ Stage~1 counts.}
\label{tab:deepk}
\centering
\footnotesize
\setlength{\tabcolsep}{2.4pt}
\begin{tabular*}{\textwidth}{@{\extracolsep{\fill}}lccccccc@{}}
\toprule
Arm & \shortstack{Obs.\\cross} & \shortstack{First-loss\\$\kappa\in$} &
\shortstack{Dom.\\$p$} & $\hat D(4096)$ & \shortstack{95\% band at\\$k{=}4096$} &
\shortstack{Kernel $\kstar$\\($n{=}256$)} & \shortstack{Kernel $D(4096)$\\(90\% interval)} \\
\midrule
GRPO $G{=}4$   & $214$ & $[103,\infty)$  & $.043$          & $-.0033$ & $[-.0097,\,+.0030]$ & $202$  & $+.012$ $[-.022,\,+.039]$ \\
GRPO $G{=}8$   & $178$ & $\mathbf{[95,791]}$ & $\mathbf{.008}$ & $-.0057$ & $[-.0122,\,+.0009]$ & $196$  & $-.019$ $[-.045,\,+.019]$ \\
GRPO $G{=}16$  & $234$ & $[114,\infty)$  & $.045$          & $-.0053$ & $[-.0121,\,+.0014]$ & $196$  & $-.004$ $[-.040,\,+.031]$ \\
GRPO weak KL   & $216$ & $[103,\infty)$  & $.075$          & $-.0047$ & $[-.0110,\,+.0017]$ & $321$  & $-.020$ $[-.045,\,+.014]$ \\
SFT            & none  & $[4097,\infty)$ & $1.00$          & $+.0080$ & $\mathbf{[+.0023,\,+.0137]}$ & $2875$ & $-.003$ $[-.027,\,+.031]$ \\
\bottomrule
\end{tabular*}
\end{table}

\paragraph{First-principles limits of extrapolating pass@$k$ beyond the
fitting depth.} Comparing the last two columns of Table~\ref{tab:deepk}
against the direct measurements reveals both the utility and the
fundamental resolution limit of fitted response models. Kernels fitted to
the $n=256$ counts accurately locate the initial GRPO crossovers just
around the fitting horizon ($\kstar=196$--$321$ predicted versus
$178$--$234$ observed), and across all five arms and all nine pre-recorded
prediction budgets up to $k=4096$, all 45 observed $\hat D(k)$ values fall
inside the kernels' 90\% bootstrap intervals.

However, at $k=4096$---sixteen times the fitting depth $n=256$---those
bootstrap intervals are $\approx.06$ wide (ten times the observed effect
size), and the point predictions fail to separate the arms: the $n=256$
kernel predicts $+.012$ for GRPO $G=4$ and $-.003$ (with a crossing at
$\kstar=2875$) for SFT, opposite the observed signs ($-.0033$ and $+.0080$).
Why does any model fitted at depth $n$ lose sign resolution at $k\gg n$?
From first principles, $n$ binomial trials expose only the first $n$
moments of the mixing law and cannot resolve success probabilities below
$q\sim 1/n$: at $n=256$, a prompt with $q_i=10^{-3}$ (which succeeds with
probability $.98$ at $k=4096$) and a prompt with $q_i=10^{-5}$ (which
succeeds with probability $.04$ at $k=4096$) both yield $d_i=0$ with
probability $>77\%$. Because $\passk$ at $k\gg n$ is governed precisely by
how mass is distributed inside the sub-resolution bin $[0,1/n]$, a kernel
fitted at depth $n$ can reliably diagnose behavior up to $k\approx n$, but
claims at $k\gg n$ require sampling at those budgets.

\section{Related Work}\label{sec:related}

\paragraph{The RLVR capability debate and mechanism models.}
\citet{yue2025limit} documented pass@$k$ crossovers across six RLVR
algorithms and interpreted them as evidence that RLVR sharpens the base
model's distribution without expanding its support. Subsequent work has
contested or refined this reading: \citet{wen2025rlvr} argue that RLVR
improves valid intermediate reasoning chains (CoT-Pass@K),
\citet{wu2025leash} formalize RLVR as support-constrained conservative
reweighting, and \citet{liu2025prorl} contend that prolonged training across
diverse domains expands the reasoning boundary. \citet{kim2025rlvd} and
\citet{zhao2026sharpening} examine post-RL accuracy stratified by base
success rate, providing descriptive counterparts to our mean response curve
$T(p)$. Because bin averages integrate out within-bin variation, however,
they cannot test whether prompts of equal base difficulty respond alike,
which our split-generation test and Lipschitz slope bound resolve.

Several recent studies propose training mechanisms and interventions for
pass@$k$ crossovers. \citet{zhou2026inversion} attributes crossovers to
boundary problems whose rare correct rollouts miss finite rollout groups,
models post-RL success as a deterministic threshold function of base success,
and anchors updates to base difficulty; \citet{yuan2026overtraining}
attribute crossovers to continued updates after a problem's pass@$k$
contribution saturates; and \citet{nguyen2025boundary} point to negative
gradient interference across problems that favors high-$p$ prompts. While
negative interference and rollout-group censoring are natural sources of
the conditional dispersion we measure, any deterministic threshold map is a
single response curve of the type rejected by our split test and slope
bound. Empirically, these works evaluate point estimates of pass@$k$ or
attach per-$k$ intervals across training seeds
\citep{zhou2026inversion,strozzi2026teacher}; our paired simultaneous bands
provide the statistical test needed to verify whether a training
intervention genuinely removes a crossover. In controlled synthetic settings,
\citet{clay2026demystifying} show that dense rewards can rescue
low-probability base behaviors where sparse rewards fail, reinforcing our
conclusion that a pass@$k$ crossover reflects how training redistributes
mass across prompts rather than an absolute ceiling on base capability.

\paragraph{Statistical inference and scaling laws for pass@$k$.}
\citet{brown2024monkeys} introduced repeated-sampling scaling laws for
language models, and \citet{schaeffer2025monkeys} derived power-law
pass@$k$ tails from Beta-distributed prompt difficulty $F_P$, the starting
point for our Theorem~\ref{thm:crossover}. \citet{kazdan2025passk}
extrapolate pass@$k$ within a single model via Beta-Binomial fits, and
\citet{levi2026hardtail} analyzes a solvable supervised model in which the
pass@$k$ tail exponent rises with data size---mirroring the contrast we
observe in Section~\ref{sec:planned} between SFT and GRPO.
\citet{hariri2025bayes} propose Bayesian single-model evaluation, and
\citet{beyond2025passk} note that random guessing creates an artificial
floor at $k\sim 1/p_{\mathrm{guess}}$ (which our free-form symbolic and
execution verifiers avoid). For uncertainty quantification,
\citet{miller2024errorbars} and \citet{bowyer2025clt} study confidence
intervals for scalar benchmark accuracy, while \citet{bay2026ceiling} show
that within-prompt correlation caps the information gained from deeper
per-prompt sampling---the exact phenomenon captured by our paired variance
decomposition \eqref{eq:var-decomp}. Crossing tests exist for independent
survival hazard curves \citep{hazardcross2009}, whereas our multiplier
bootstrap exploits the prompt-paired design and across-$k$ correlation.

\paragraph{Exponential tilting in KL-regularized RL.}
\citet{huang2025sharpening} characterize KL-regularized RL as a Gibbs
exponential tilt, and \citet{chen2025tilting}, \citet{mroueh2025grpo},
\citet{polson2026rlvr}, and \citet{tomihari2026power} develop variational,
Bayesian, and power-distillation perspectives on this optimal policy. We use
the exact binary-partition tilt identity \eqref{eq:operator-exact} to show
why within-cell R\'enyi mass ratios $h_i$ break any functional mapping
$p\mapsto q$ and induce a Markov kernel $H(q\mid p)$. In concurrent work,
\citet{liu2026nofreelunch} models RLVR as an exponential tilt over latent
solution strategies shared across prompts, establishes conditions under which
pass@1 gains force losses on some prompts, and independently notes the
marginal variation-diminishing bound of Proposition~\ref{prop:signs};
shared latent strategies provide a concrete structural mechanism for why
prompts of identical base success $p$ carry different $h_i$ and diverge into
a conditional distribution $H(q\mid p)$.

\section{Conclusion}\label{sec:discussion}

A pass@$k$ crossover is a joint statistical claim about an entire paired
difference curve $D(k)$ and must be evaluated as one. Treating crossovers
through paired simultaneous confidence bands and nonparametric kernel
estimation reshapes both the empirical picture and the mechanistic
interpretation of RLVR:
\begin{enumerate}
\item Most visually reported crossovers at standard token limits (3k--16k)
are not statistically established at typical sample sizes, and per-response
token limits act as a powerful, frequently unreported confound via
asymmetric truncation. When evaluated at a 32k-token limit on 1,060 new
competition problems, crossovers are statistically confirmed on two 1.5B
long-chain RLVR models ($\kappa\in[11,61]$ and $[12,63]$) but absent on two
7B models.
\item RLVR does not act on prompt difficulty as a deterministic function
$q=T(p)$; it acts as a Markov kernel $H(q\mid p)$. Across every model pair,
prompts of identical base success diverge reproducibly across independent
generation halves and violate the Lipschitz slope bounds of any reasonable
single curve.
\item When a crossover occurs, it does not imply that RLVR degrades hard
prompts across the board. Once conditional dispersion is accounted for, the
average prompt improves across nearly all difficulty levels; the late-$k$
loss is driven by a bimodal split in which a small minority ($2\%$--$3\%$)
of base-solved prompts are locked into repeating an incorrect modal answer.
In controlled runs measured to $k=4096$ on 3,000 new problems, GRPO
exhibits this late-$k$ loss from $k\approx200$ onward, whereas SFT stays
ahead of the base model throughout.
\end{enumerate}
The productive scientific question is therefore not whether RLVR globally
``expands'' or ``only sharpens'' reasoning capability, but what features of
a prompt and training algorithm cause a minority of solvable problems to
collapse into wrong-answer lock-in while other problems of the exact same
base difficulty improve several-fold.

\paragraph{Five recommendations for evaluating and reporting pass@$k$
comparisons.}
\begin{enumerate}
\item \emph{Report truncation rates and check token-limit sensitivity.}
Always report per-response token limits and truncation rates for both models,
and verify whether raising the token limit alters the comparison.
\item \emph{Prioritize more prompts ($m$) over excessive depth ($n$).}
Set the per-prompt rollout depth $n$ to cover the largest budget $K$ of
substantive interest, then allocate remaining compute to increasing the
number of prompts $m$, which reduces both between-prompt and within-prompt
variance \eqref{eq:var-decomp}.
\item \emph{Use paired simultaneous bands to certify crossovers.}
Require both $L(j)>0$ at some smaller $k=j$ and $U(\ell)<0$ at some larger
$k=\ell>j$ (Proposition~\ref{prop:cert}), rather than relying on visual
crossings, pointwise error bars, or one-sided dominance tests alone.
\item \emph{Run the split-generation test before interpreting scalar
sharpening exponents.} The test \eqref{eq:split-stat} requires no extra
sampling; when it rejects determinism, any fitted slope $\hat\beta$ is a
marginal summary of a kernel $H(q\mid p)$ rather than a prompt-level law.
\item \emph{Sample directly at the budgets $k$ about which you make claims.}
Because $n$ binomial draws cannot resolve success probabilities below
$1/n$, response models fitted at depth $n$ cannot reliably extrapolate the
sign of $D(k)$ to budgets $k\gg n$.
\end{enumerate}

\paragraph{Limitations.}
Our statistically supported public crossovers are established at a 32k-token
limit on mathematical competition problems; while the split-generation test
and token-limit sensitivity replicate on code (HumanEval+/MBPP+) and on
non-Qwen bases (Llama-3.1-8B and DeepSeek-Math-7B), evaluating large-$k$
crossovers in multi-turn agentic or tool-use settings remains open. Finally,
our controlled GRPO and SFT runs use 100--132 updates on a 1.5B base model
(17 to 30 times fewer steps than DeepScaleR or ProRL); extending controlled
$k=4096$ evaluations across longer training horizons and larger base scales
is a natural next step.

\subsection*{Code and data}
The \texttt{passk-inference} package implementing the dominance test,
simultaneous confidence bands, first-loss intervals, split-generation test,
Lipschitz slope bound, 2D NPMLE kernel fit, and power calculations, along
with all per-prompt count datasets, is available at
\url{https://github.com/cafferychen777/passk-inference}.

\subsection*{AI use statement}
Large language models were used as coding and writing assistants throughout
this project: drafting and refactoring analysis code, managing compute jobs,
literature search, and prose editing, all under author direction and review.
The theoretical results, experimental design, and interpretation were
verified by the authors; all numerical claims trace to released code and
per-prompt counts.

\bibliography{references}

\begin{thebibliography}{55}
\providecommand{\natexlab}[1]{#1}
\providecommand{\url}[1]{\texttt{#1}}
\expandafter\ifx\csname urlstyle\endcsname\relax
  \providecommand{\doi}[1]{doi: #1}\else
  \providecommand{\doi}{doi: \begingroup \urlstyle{rm}\Url}\fi

\bibitem[Austin et~al.(2021)Austin, Odena, Nye, Bosma, Michalewski, Dohan,
  Jiang, Cai, Terry, Le, and Sutton]{austin2021mbpp}
Jacob Austin, Augustus Odena, Maxwell Nye, Maarten Bosma, Henryk Michalewski,
  David Dohan, Ellen Jiang, Carrie Cai, Michael Terry, Quoc Le, and Charles
  Sutton.
\newblock Program synthesis with large language models.
\newblock \emph{arXiv preprint arXiv:2108.07732}, 2021.
\newblock \doi{10.48550/arXiv.2108.07732}.
\newblock URL \url{https://arxiv.org/abs/2108.07732}.

\bibitem[Bay \& Yearick(2026)Bay and Yearick]{bay2026ceiling}
Yong~Yi Bay and Kathleen~A. Yearick.
\newblock When more sampling hurts: The modal ceiling and correlation ceiling
  of test-time scaling.
\newblock \emph{arXiv preprint arXiv:2606.28661}, 2026.
\newblock \doi{10.48550/arXiv.2606.28661}.
\newblock URL \url{https://arxiv.org/abs/2606.28661}.

\bibitem[Bowyer et~al.(2025)Bowyer, Aitchison, and Ivanova]{bowyer2025clt}
Sam Bowyer, Laurence Aitchison, and Desi~R. Ivanova.
\newblock Position: Don't use the {CLT} in {LLM} evals with fewer than a few
  hundred datapoints.
\newblock In \emph{Proceedings of the 42nd International Conference on Machine
  Learning}, volume 267 of \emph{Proceedings of Machine Learning Research},
  pp.\  81143--81184. PMLR, 2025.
\newblock URL \url{https://proceedings.mlr.press/v267/bowyer25a.html}.

\bibitem[Brown et~al.(2024)Brown, Juravsky, Ehrlich, Clark, Le, R{\'e}, and
  Mirhoseini]{brown2024monkeys}
Bradley Brown, Jordan Juravsky, Ryan Ehrlich, Ronald Clark, Quoc~V. Le,
  Christopher R{\'e}, and Azalia Mirhoseini.
\newblock Large language monkeys: Scaling inference compute with repeated
  sampling.
\newblock \emph{arXiv preprint arXiv:2407.21787}, 2024.
\newblock \doi{10.48550/arXiv.2407.21787}.
\newblock URL \url{https://arxiv.org/abs/2407.21787}.

\bibitem[Chen et~al.(2021)Chen, Tworek, Jun, Yuan, Pinto, Kaplan, Edwards,
  Burda, Joseph, Brockman, Ray, Puri, Krueger, Petrov, Khlaaf, Sastry, Mishkin,
  Chan, Gray, Ryder, Pavlov, Power, Kaiser, Bavarian, Winter, Tillet, Such,
  Cummings, Plappert, Chantzis, Barnes, Herbert-Voss, Guss, Nichol, Paino,
  Tezak, Tang, Babuschkin, Balaji, Jain, Saunders, Hesse, Carr, Leike, Achiam,
  Misra, Morikawa, Radford, Knight, Brundage, Murati, Mayer, Welinder, McGrew,
  Amodei, McCandlish, Sutskever, and Zaremba]{chen2021humaneval}
Mark Chen, Jerry Tworek, Heewoo Jun, Qiming Yuan, Henrique Ponde de~Oliveira
  Pinto, Jared Kaplan, Harri Edwards, Yuri Burda, Nicholas Joseph, Greg
  Brockman, Alex Ray, Raul Puri, Gretchen Krueger, Michael Petrov, Heidy
  Khlaaf, Girish Sastry, Pamela Mishkin, Brooke Chan, Scott Gray, Nick Ryder,
  Mikhail Pavlov, Alethea Power, Lukasz Kaiser, Mohammad Bavarian, Clemens
  Winter, Philippe Tillet, Felipe~Petroski Such, Dave Cummings, Matthias
  Plappert, Fotios Chantzis, Elizabeth Barnes, Ariel Herbert-Voss,
  William~Hebgen Guss, Alex Nichol, Alex Paino, Nikolas Tezak, Jie Tang, Igor
  Babuschkin, Suchir Balaji, Shantanu Jain, William Saunders, Christopher
  Hesse, Andrew~N. Carr, Jan Leike, Josh Achiam, Vedant Misra, Evan Morikawa,
  Alec Radford, Matthew Knight, Miles Brundage, Mira Murati, Katie Mayer, Peter
  Welinder, Bob McGrew, Dario Amodei, Sam McCandlish, Ilya Sutskever, and
  Wojciech Zaremba.
\newblock Evaluating large language models trained on code.
\newblock \emph{arXiv preprint arXiv:2107.03374}, 2021.
\newblock \doi{10.48550/arXiv.2107.03374}.
\newblock URL \url{https://arxiv.org/abs/2107.03374}.

\bibitem[Chen et~al.(2025)Chen, Yang, Liu, Lee, Xu, Shoeybi, Catanzaro, and
  Ping]{chen2025acereason}
Yang Chen, Zhuolin Yang, Zihan Liu, Chankyu Lee, Peng Xu, Mohammad Shoeybi,
  Bryan Catanzaro, and Wei Ping.
\newblock {AceReason-Nemotron}: Advancing math and code reasoning through
  reinforcement learning.
\newblock \emph{arXiv preprint arXiv:2505.16400}, 2025.
\newblock \doi{10.48550/arXiv.2505.16400}.
\newblock URL \url{https://arxiv.org/abs/2505.16400}.

\bibitem[Cheng et~al.(2009)Cheng, Qiu, Tan, and Tu]{hazardcross2009}
Ming-Yen Cheng, Peihua Qiu, Xianming Tan, and Dongsheng Tu.
\newblock Confidence intervals for the first crossing point of two hazard
  functions.
\newblock \emph{Lifetime Data Analysis}, 15\penalty0 (4):\penalty0 441--454,
  2009.
\newblock \doi{10.1007/s10985-009-9132-6}.

\bibitem[Clay et~al.(2026)Clay, Gollapudi, Harilal, Jang, Morrison, Oh, and
  Jaques]{clay2026demystifying}
Donovan Clay, Saket Gollapudi, Sankar Harilal, Min Jang, Jacob Morrison,
  Sewoong Oh, and Natasha Jaques.
\newblock Demystifying reinforcement learning post-training of language models,
  2026.
\newblock URL \url{https://arxiv.org/abs/2608.24949}.

\bibitem[Dragoi et~al.(2025)Dragoi, Pintilie, Gogianu, and
  Brad]{beyond2025passk}
Marius Dragoi, Ioana Pintilie, Florin Gogianu, and Florin Brad.
\newblock Beyond pass@k: Breadth-depth metrics for reasoning boundaries.
\newblock In \emph{First Workshop on Foundations of Reasoning in Language
  Models}, 2025.
\newblock URL \url{https://openreview.net/forum?id=49pjTbyu7d}.

\bibitem[Dvoretzky et~al.(1951)Dvoretzky, Wald, and
  Wolfowitz]{dvoretzky1951elimination}
Aryeh Dvoretzky, Abraham Wald, and Jacob Wolfowitz.
\newblock Elimination of randomization in certain statistical decision
  procedures and zero-sum two-person games.
\newblock \emph{The Annals of Mathematical Statistics}, 22\penalty0
  (1):\penalty0 1--21, 1951.

\bibitem[Fan et~al.(2026)Fan, Han, Wang, Chen, Zhang, and
  Zhou]{zhao2026sharpening}
Mingyuan Fan, Weiguang Han, Daixin Wang, Cen Chen, Zhiqiang Zhang, and Jun
  Zhou.
\newblock When sharpening becomes collapse: Sampling bias and semantic coupling
  in {RL} with verifiable rewards.
\newblock \emph{arXiv preprint arXiv:2601.15609}, 2026.
\newblock \doi{10.48550/arXiv.2601.15609}.
\newblock URL \url{https://arxiv.org/abs/2601.15609}.

\bibitem[Fieller(1954)]{fieller1954}
E.~C. Fieller.
\newblock Some problems in interval estimation.
\newblock \emph{Journal of the Royal Statistical Society: Series B
  (Methodological)}, 16\penalty0 (2):\penalty0 175--185, 1954.
\newblock \doi{10.1111/j.2517-6161.1954.tb00159.x}.

\bibitem[Grattafiori et~al.(2024)Grattafiori, Dubey, Jauhri, Pandey, Kadian,
  Al-Dahle, Letman, Mathur, Schelten, Vaughan, Yang, Fan, Goyal, Hartshorn,
  Yang, Mitra, Sravankumar, Korenev, Hinsvark, Rao, Zhang, Rodriguez,
  Gregerson, Spataru, Roziere, Biron, Tang, Chern, Caucheteux, Nayak, Bi,
  Marra, McConnell, Keller, Touret, Wu, Wong, Ferrer, Nikolaidis, Allonsius,
  Song, Pintz, Livshits, Wyatt, Esiobu, Choudhary, Mahajan, Garcia-Olano,
  Perino, Hupkes, Lakomkin, AlBadawy, Lobanova, Dinan, Smith, Radenovic,
  Guzm{\'{a}}n, Zhang, Synnaeve, Lee, Anderson, Thattai, Nail, Mialon, Pang,
  Cucurell, Nguyen, Korevaar, Xu, Touvron, Zarov, Ibarra, Kloumann, Misra,
  Evtimov, Zhang, Copet, Lee, Geffert, Vranes, Park, Mahadeokar, Shah, van~der
  Linde, Billock, Hong, Lee, Fu, Chi, Huang, Liu, Wang, Yu, Bitton, Spisak,
  Park, Rocca, Johnstun, Saxe, Jia, Alwala, Prasad, Upasani, Plawiak, Li,
  Heafield, Stone, El-Arini, Iyer, Malik, Chiu, Bhalla, Lakhotia,
  Rantala-Yeary, van~der Maaten, Chen, Tan, Jenkins, Martin, Madaan, Malo,
  Blecher, Landzaat, de~Oliveira, Muzzi, Pasupuleti, Singh, Paluri, Kardas,
  Tsimpoukelli, Oldham, Rita, Pavlova, Kambadur, Lewis, Si, Singh, Hassan,
  Goyal, Torabi, Bashlykov, Bogoychev, Chatterji, Zhang, Duchenne,
  {\c{C}}elebi, Alrassy, Zhang, Li, Vasic, Weng, Bhargava, Dubal, Krishnan,
  Koura, Xu, He, Dong, Srinivasan, Ganapathy, Calderer, Cabral, Stojnic,
  Raileanu, Maheswari, Girdhar, Patel, Sauvestre, Polidoro, Sumbaly, Taylor,
  Silva, Hou, Wang, Hosseini, Chennabasappa, Singh, Bell, Kim, Edunov, Nie,
  Narang, Raparthy, Shen, Wan, Bhosale, Zhang, Vandenhende, Batra, Whitman,
  Sootla, Collot, Gururangan, Borodinsky, Herman, Fowler, Sheasha, Georgiou,
  Scialom, Speckbacher, Mihaylov, Xiao, Karn, Goswami, Gupta, Ramanathan,
  Kerkez, Gonguet, Do, Vogeti, Albiero, Petrovic, Chu, Xiong, Fu, Meers,
  Martinet, Wang, Wang, Tan, Xia, Xie, Jia, Wang, Goldschlag, Gaur, Babaei,
  Wen, Song, Zhang, Li, Mao, Coudert, Yan, Chen, Papakipos, Singh, Srivastava,
  Jain, Kelsey, Shajnfeld, Gangidi, Victoria, Goldstand, Menon, Sharma,
  Boesenberg, Baevski, Feinstein, Kallet, Sangani, Teo, Yunus, Lupu, Alvarado,
  Caples, Gu, Ho, Poulton, Ryan, Ramchandani, Dong, Franco, Goyal, Saraf,
  Chowdhury, Gabriel, Bharambe, Eisenman, Yazdan, James, Maurer, Leonhardi,
  Huang, Loyd, De~Paola, Paranjape, Liu, Wu, Ni, Hancock, Wasti, Spence,
  Stojkovic, Gamido, Montalvo, Parker, Burton, Mejia, Liu, Wang, Kim, Zhou, Hu,
  Chu, Cai, Tindal, Feichtenhofer, Gao, Civin, Beaty, Kreymer, Li, Adkins, Xu,
  Testuggine, David, Parikh, Liskovich, Foss, Wang, Le, Holland, Dowling,
  Jamil, Montgomery, Presani, Hahn, Wood, Le, Brinkman, Arcaute, Dunbar,
  Smothers, Sun, Kreuk, Tian, Kokkinos, Ozgenel, Caggioni, Kanayet, Seide,
  Florez, Schwarz, Badeer, Swee, Halpern, Herman, Sizov, Zhang,
  Lakshminarayanan, Inan, Shojanazeri, Zou, Wang, Zha, Habeeb, Rudolph, Suk,
  Aspegren, Goldman, Zhan, Damlaj, Molybog, Tufanov, Leontiadis, Veliche, Gat,
  Weissman, Geboski, Kohli, Lam, Asher, Gaya, Marcus, Tang, Chan, Zhen,
  Reizenstein, Teboul, Zhong, Jin, Yang, Cummings, Carvill, Shepard, McPhie,
  Torres, Ginsburg, Wang, Wu, U, Saxena, Khandelwal, Zand, Matosich,
  Veeraraghavan, Michelena, Li, Jagadeesh, Huang, Chawla, Huang, Chen, Garg, A,
  Silva, Bell, Zhang, Guo, Yu, Moshkovich, Wehrstedt, Khabsa, Avalani, Bhatt,
  Mankus, Hasson, Lennie, Reso, Groshev, Naumov, Lathi, Keneally, Liu, Seltzer,
  Valko, Restrepo, Patel, Vyatskov, Samvelyan, Clark, Macey, Wang, Hermoso,
  Metanat, Rastegari, Bansal, Santhanam, Parks, White, Bawa, Singhal, Egebo,
  Usunier, Mehta, Laptev, Dong, Cheng, Chernoguz, Hart, Salpekar, Kalinli,
  Kent, Parekh, Saab, Balaji, Rittner, Bontrager, Roux, Dollar, Zvyagina,
  Ratanchandani, Yuvraj, Liang, Alao, Rodriguez, Ayub, Murthy, Nayani, Mitra,
  Parthasarathy, Li, Hogan, Battey, Wang, Howes, Rinott, Mehta, Siby, Bondu,
  Datta, Chugh, Hunt, Dhillon, Sidorov, Pan, Mahajan, Verma, Yamamoto,
  Ramaswamy, Lindsay, Lindsay, Feng, Lin, Zha, Patil, Shankar, Zhang, Zhang,
  Wang, Agarwal, Sajuyigbe, Chintala, Max, Chen, Kehoe, Satterfield,
  Govindaprasad, Gupta, Deng, Cho, Virk, Subramanian, Choudhury, Goldman,
  Remez, Glaser, Best, Koehler, Robinson, Li, Zhang, Matthews, Chou, Shaked,
  Vontimitta, Ajayi, Montanez, Mohan, Kumar, Mangla, Ionescu, Poenaru,
  Mihailescu, Ivanov, Li, Wang, Jiang, Bouaziz, Constable, Tang, Wu, Wang, Wu,
  Gao, Kleinman, Chen, Hu, Jia, Qi, Li, Zhang, Zhang, Adi, Nam, Wang, Zhao,
  Hao, Qian, Li, He, Rait, DeVito, Rosnbrick, Wen, Yang, Zhao, and
  Ma]{grattafiori2024llama3}
Aaron Grattafiori, Abhimanyu Dubey, Abhinav Jauhri, Abhinav Pandey, Abhishek
  Kadian, Ahmad Al-Dahle, Aiesha Letman, Akhil Mathur, Alan Schelten, Alex
  Vaughan, Amy Yang, Angela Fan, Anirudh Goyal, Anthony Hartshorn, Aobo Yang,
  Archi Mitra, Archie Sravankumar, Artem Korenev, Arthur Hinsvark, Arun Rao,
  Aston Zhang, Aurelien Rodriguez, Austen Gregerson, Ava Spataru, Baptiste
  Roziere, Bethany Biron, Binh Tang, Bobbie Chern, Charlotte Caucheteux, Chaya
  Nayak, Chloe Bi, Chris Marra, Chris McConnell, Christian Keller, Christophe
  Touret, Chunyang Wu, Corinne Wong, Cristian~Canton Ferrer, Cyrus Nikolaidis,
  Damien Allonsius, Daniel Song, Danielle Pintz, Danny Livshits, Danny Wyatt,
  David Esiobu, Dhruv Choudhary, Dhruv Mahajan, Diego Garcia-Olano, Diego
  Perino, Dieuwke Hupkes, Egor Lakomkin, Ehab AlBadawy, Elina Lobanova, Emily
  Dinan, Eric~Michael Smith, Filip Radenovic, Francisco Guzm{\'{a}}n, Frank
  Zhang, Gabriel Synnaeve, Gabrielle Lee, Georgia~Lewis Anderson, Govind
  Thattai, Graeme Nail, Gregoire Mialon, Guan Pang, Guillem Cucurell, Hailey
  Nguyen, Hannah Korevaar, Hu~Xu, Hugo Touvron, Iliyan Zarov, Imanol~Arrieta
  Ibarra, Isabel Kloumann, Ishan Misra, Ivan Evtimov, Jack Zhang, Jade Copet,
  Jaewon Lee, Jan Geffert, Jana Vranes, Jason Park, Jay Mahadeokar, Jeet Shah,
  Jelmer van~der Linde, Jennifer Billock, Jenny Hong, Jenya Lee, Jeremy Fu,
  Jianfeng Chi, Jianyu Huang, Jiawen Liu, Jie Wang, Jiecao Yu, Joanna Bitton,
  Joe Spisak, Jongsoo Park, Joseph Rocca, Joshua Johnstun, Joshua Saxe, Junteng
  Jia, Kalyan~Vasuden Alwala, Karthik Prasad, Kartikeya Upasani, Kate Plawiak,
  Ke~Li, Kenneth Heafield, Kevin Stone, Khalid El-Arini, Krithika Iyer, Kshitiz
  Malik, Kuenley Chiu, Kunal Bhalla, Kushal Lakhotia, Lauren Rantala-Yeary,
  Laurens van~der Maaten, Lawrence Chen, Liang Tan, Liz Jenkins, Louis Martin,
  Lovish Madaan, Lubo Malo, Lukas Blecher, Lukas Landzaat, Luke de~Oliveira,
  Madeline Muzzi, Mahesh Pasupuleti, Mannat Singh, Manohar Paluri, Marcin
  Kardas, Maria Tsimpoukelli, Mathew Oldham, Mathieu Rita, Maya Pavlova,
  Melanie Kambadur, Mike Lewis, Min Si, Mitesh~Kumar Singh, Mona Hassan, Naman
  Goyal, Narjes Torabi, Nikolay Bashlykov, Nikolay Bogoychev, Niladri
  Chatterji, Ning Zhang, Olivier Duchenne, Onur {\c{C}}elebi, Patrick Alrassy,
  Pengchuan Zhang, Pengwei Li, Petar Vasic, Peter Weng, Prajjwal Bhargava,
  Pratik Dubal, Praveen Krishnan, Punit~Singh Koura, Puxin Xu, Qing He,
  Qingxiao Dong, Ragavan Srinivasan, Raj Ganapathy, Ramon Calderer,
  Ricardo~Silveira Cabral, Robert Stojnic, Roberta Raileanu, Rohan Maheswari,
  Rohit Girdhar, Rohit Patel, Romain Sauvestre, Ronnie Polidoro, Roshan
  Sumbaly, Ross Taylor, Ruan Silva, Rui Hou, Rui Wang, Saghar Hosseini, Sahana
  Chennabasappa, Sanjay Singh, Sean Bell, Seohyun~Sonia Kim, Sergey Edunov,
  Shaoliang Nie, Sharan Narang, Sharath Raparthy, Sheng Shen, Shengye Wan,
  Shruti Bhosale, Shun Zhang, Simon Vandenhende, Soumya Batra, Spencer Whitman,
  Sten Sootla, Stephane Collot, Suchin Gururangan, Sydney Borodinsky, Tamar
  Herman, Tara Fowler, Tarek Sheasha, Thomas Georgiou, Thomas Scialom, Tobias
  Speckbacher, Todor Mihaylov, Tong Xiao, Ujjwal Karn, Vedanuj Goswami, Vibhor
  Gupta, Vignesh Ramanathan, Viktor Kerkez, Vincent Gonguet, Virginie Do, Vish
  Vogeti, V{\'{i}}tor Albiero, Vladan Petrovic, Weiwei Chu, Wenhan Xiong,
  Wenyin Fu, Whitney Meers, Xavier Martinet, Xiaodong Wang, Xiaofang Wang,
  Xiaoqing~Ellen Tan, Xide Xia, Xinfeng Xie, Xuchao Jia, Xuewei Wang, Yaelle
  Goldschlag, Yashesh Gaur, Yasmine Babaei, Yi~Wen, Yiwen Song, Yuchen Zhang,
  Yue Li, Yuning Mao, Zacharie~Delpierre Coudert, Zheng Yan, Zhengxing Chen,
  Zoe Papakipos, Aaditya Singh, Aayushi Srivastava, Abha Jain, Adam Kelsey,
  Adam Shajnfeld, Adithya Gangidi, Adolfo Victoria, Ahuva Goldstand, Ajay
  Menon, Ajay Sharma, Alex Boesenberg, Alexei Baevski, Allie Feinstein, Amanda
  Kallet, Amit Sangani, Amos Teo, Anam Yunus, Andrei Lupu, Andres Alvarado,
  Andrew Caples, Andrew Gu, Andrew Ho, Andrew Poulton, Andrew Ryan, Ankit
  Ramchandani, Annie Dong, Annie Franco, Anuj Goyal, Aparajita Saraf,
  Arkabandhu Chowdhury, Ashley Gabriel, Ashwin Bharambe, Assaf Eisenman, Azadeh
  Yazdan, Beau James, Ben Maurer, Benjamin Leonhardi, Bernie Huang, Beth Loyd,
  Beto De~Paola, Bhargavi Paranjape, Bing Liu, Bo~Wu, Boyu Ni, Braden Hancock,
  Bram Wasti, Brandon Spence, Brani Stojkovic, Brian Gamido, Britt Montalvo,
  Carl Parker, Carly Burton, Catalina Mejia, Ce~Liu, Changhan Wang, Changkyu
  Kim, Chao Zhou, Chester Hu, Ching-Hsiang Chu, Chris Cai, Chris Tindal,
  Christoph Feichtenhofer, Cynthia Gao, Damon Civin, Dana Beaty, Daniel
  Kreymer, Daniel Li, David Adkins, David Xu, Davide Testuggine, Delia David,
  Devi Parikh, Diana Liskovich, Didem Foss, Dingkang Wang, Duc Le, Dustin
  Holland, Edward Dowling, Eissa Jamil, Elaine Montgomery, Eleonora Presani,
  Emily Hahn, Emily Wood, Eric-Tuan Le, Erik Brinkman, Esteban Arcaute, Evan
  Dunbar, Evan Smothers, Fei Sun, Felix Kreuk, Feng Tian, Filippos Kokkinos,
  Firat Ozgenel, Francesco Caggioni, Frank Kanayet, Frank Seide,
  Gabriela~Medina Florez, Gabriella Schwarz, Gada Badeer, Georgia Swee, Gil
  Halpern, Grant Herman, Grigory Sizov, Guangyi~(Jack) Zhang, Guna
  Lakshminarayanan, Hakan Inan, Hamid Shojanazeri, Han Zou, Hannah Wang, Hanwen
  Zha, Haroun Habeeb, Harrison Rudolph, Helen Suk, Henry Aspegren, Hunter
  Goldman, Hongyuan Zhan, Ibrahim Damlaj, Igor Molybog, Igor Tufanov, Ilias
  Leontiadis, Irina-Elena Veliche, Itai Gat, Jake Weissman, James Geboski,
  James Kohli, Janice Lam, Japhet Asher, Jean-Baptiste Gaya, Jeff Marcus, Jeff
  Tang, Jennifer Chan, Jenny Zhen, Jeremy Reizenstein, Jeremy Teboul, Jessica
  Zhong, Jian Jin, Jingyi Yang, Joe Cummings, Jon Carvill, Jon Shepard,
  Jonathan McPhie, Jonathan Torres, Josh Ginsburg, Junjie Wang, Kai Wu, Kam~Hou
  U, Karan Saxena, Kartikay Khandelwal, Katayoun Zand, Kathy Matosich, Kaushik
  Veeraraghavan, Kelly Michelena, Keqian Li, Kiran Jagadeesh, Kun Huang, Kunal
  Chawla, Kyle Huang, Lailin Chen, Lakshya Garg, Lavender A, Leandro Silva, Lee
  Bell, Lei Zhang, Liangpeng Guo, Licheng Yu, Liron Moshkovich, Luca Wehrstedt,
  Madian Khabsa, Manav Avalani, Manish Bhatt, Martynas Mankus, Matan Hasson,
  Matthew Lennie, Matthias Reso, Maxim Groshev, Maxim Naumov, Maya Lathi,
  Meghan Keneally, Miao Liu, Michael~L. Seltzer, Michal Valko, Michelle
  Restrepo, Mihir Patel, Mik Vyatskov, Mikayel Samvelyan, Mike Clark, Mike
  Macey, Mike Wang, Miquel~Jubert Hermoso, Mo~Metanat, Mohammad Rastegari,
  Munish Bansal, Nandhini Santhanam, Natascha Parks, Natasha White, Navyata
  Bawa, Nayan Singhal, Nick Egebo, Nicolas Usunier, Nikhil Mehta,
  Nikolay~Pavlovich Laptev, Ning Dong, Norman Cheng, Oleg Chernoguz, Olivia
  Hart, Omkar Salpekar, Ozlem Kalinli, Parkin Kent, Parth Parekh, Paul Saab,
  Pavan Balaji, Pedro Rittner, Philip Bontrager, Pierre Roux, Piotr Dollar,
  Polina Zvyagina, Prashant Ratanchandani, Pritish Yuvraj, Qian Liang, Rachad
  Alao, Rachel Rodriguez, Rafi Ayub, Raghotham Murthy, Raghu Nayani, Rahul
  Mitra, Rangaprabhu Parthasarathy, Raymond Li, Rebekkah Hogan, Robin Battey,
  Rocky Wang, Russ Howes, Ruty Rinott, Sachin Mehta, Sachin Siby, Sai~Jayesh
  Bondu, Samyak Datta, Sara Chugh, Sara Hunt, Sargun Dhillon, Sasha Sidorov,
  Satadru Pan, Saurabh Mahajan, Saurabh Verma, Seiji Yamamoto, Sharadh
  Ramaswamy, Shaun Lindsay, Shaun Lindsay, Sheng Feng, Shenghao Lin,
  Shengxin~Cindy Zha, Shishir Patil, Shiva Shankar, Shuqiang Zhang, Shuqiang
  Zhang, Sinong Wang, Sneha Agarwal, Soji Sajuyigbe, Soumith Chintala,
  Stephanie Max, Stephen Chen, Steve Kehoe, Steve Satterfield, Sudarshan
  Govindaprasad, Sumit Gupta, Summer Deng, Sungmin Cho, Sunny Virk, Suraj
  Subramanian, Sy~Choudhury, Sydney Goldman, Tal Remez, Tamar Glaser, Tamara
  Best, Thilo Koehler, Thomas Robinson, Tianhe Li, Tianjun Zhang, Tim Matthews,
  Timothy Chou, Tzook Shaked, Varun Vontimitta, Victoria Ajayi, Victoria
  Montanez, Vijai Mohan, Vinay~Satish Kumar, Vishal Mangla, Vlad Ionescu, Vlad
  Poenaru, Vlad~Tiberiu Mihailescu, Vladimir Ivanov, Wei Li, Wenchen Wang,
  Wenwen Jiang, Wes Bouaziz, Will Constable, Xiaocheng Tang, Xiaojian Wu,
  Xiaolan Wang, Xilun Wu, Xinbo Gao, Yaniv Kleinman, Yanjun Chen, Ye~Hu,
  Ye~Jia, Ye~Qi, Yenda Li, Yilin Zhang, Ying Zhang, Yossi Adi, Youngjin Nam,
  Yu~(Sid) Wang, Yu~Zhao, Yuchen Hao, Yundi Qian, Yunlu Li, Yuzi He, Zach Rait,
  Zachary DeVito, Zef Rosnbrick, Zhaoduo Wen, Zhenyu Yang, Zhiwei Zhao, and
  Zhiyu Ma.
\newblock The {Llama} 3 herd of models.
\newblock \emph{arXiv preprint arXiv:2407.21783}, 2024.
\newblock \doi{10.48550/arXiv.2407.21783}.
\newblock URL \url{https://arxiv.org/abs/2407.21783}.

\bibitem[Guo et~al.(2025)Guo, Yang, Zhang, Song, Wang, Zhu, Xu, Zhang, Ma, Bi,
  Zhang, Yu, Wu, Wu, Gou, Shao, Li, Gao, Liu, Xue, Wang, Wu, Feng, Lu, Zhao,
  Deng, Ruan, Dai, Chen, Ji, Li, Lin, Dai, Luo, Hao, Chen, Li, Zhang, Xu, Ding,
  Gao, Qu, Li, Guo, Li, Chen, Yuan, Tu, Qiu, Li, Cai, Ni, Liang, Chen, Dong,
  Hu, You, Gao, Guan, Huang, Yu, Wang, Zhang, Zhao, Wang, Zhang, Xu, Xia,
  Zhang, Zhang, Tang, Zhou, Li, Wang, Li, Tian, Huang, Zhang, Wang, Chen, Du,
  Ge, Zhang, Pan, Wang, Chen, Jin, Chen, Lu, Zhou, Chen, Ye, Wang, Yu, Zhou,
  Pan, Li, Zhou, Wu, Yun, Pei, Sun, Wang, Zeng, Liu, Liang, Gao, Yu, Zhang,
  Xiao, An, Liu, Wang, Chen, Nie, Cheng, Liu, Xie, Liu, Yang, Li, Su, Lin, Li,
  Jin, Shen, Chen, Sun, Wang, Song, Zhou, Wang, Shan, Li, Wang, Wei, Zhang, Xu,
  Li, Zhao, Sun, Wang, Yu, Zhang, Shi, Xiong, He, Piao, Wang, Tan, Ma, Liu,
  Guo, Ou, Wang, Gong, Zou, He, Xiong, Luo, You, Liu, Zhou, Zhu, Huang, Li,
  Zheng, Zhu, Ma, Tang, Zha, Yan, Ren, Ren, Sha, Fu, Xu, Xie, Zhang, Hao, Ma,
  Yan, Wu, Gu, Zhu, Liu, Li, Xie, Song, Pan, Huang, Xu, Zhang, and
  Zhang]{guo2025deepseekr1}
Daya Guo, Dejian Yang, Haowei Zhang, Junxiao Song, Peiyi Wang, Qihao Zhu,
  Runxin Xu, Ruoyu Zhang, Shirong Ma, Xiao Bi, Xiaokang Zhang, Xingkai Yu,
  Yu~Wu, Z.~F. Wu, Zhibin Gou, Zhihong Shao, Zhuoshu Li, Ziyi Gao, Aixin Liu,
  Bing Xue, Bingxuan Wang, Bochao Wu, Bei Feng, Chengda Lu, Chenggang Zhao,
  Chengqi Deng, Chong Ruan, Damai Dai, Deli Chen, Dongjie Ji, Erhang Li,
  Fangyun Lin, Fucong Dai, Fuli Luo, Guangbo Hao, Guanting Chen, Guowei Li,
  H.~Zhang, Hanwei Xu, Honghui Ding, Huazuo Gao, Hui Qu, Hui Li, Jianzhong Guo,
  Jiashi Li, Jingchang Chen, Jingyang Yuan, Jinhao Tu, Junjie Qiu, Junlong Li,
  J.~L. Cai, Jiaqi Ni, Jian Liang, Jin Chen, Kai Dong, Kai Hu, Kaichao You,
  Kaige Gao, Kang Guan, Kexin Huang, Kuai Yu, Lean Wang, Lecong Zhang, Liang
  Zhao, Litong Wang, Liyue Zhang, Lei Xu, Leyi Xia, Mingchuan Zhang, Minghua
  Zhang, Minghui Tang, Mingxu Zhou, Meng Li, Miaojun Wang, Mingming Li, Ning
  Tian, Panpan Huang, Peng Zhang, Qiancheng Wang, Qinyu Chen, Qiushi Du, Ruiqi
  Ge, Ruisong Zhang, Ruizhe Pan, Runji Wang, R.~J. Chen, R.~L. Jin, Ruyi Chen,
  Shanghao Lu, Shangyan Zhou, Shanhuang Chen, Shengfeng Ye, Shiyu Wang,
  Shuiping Yu, Shunfeng Zhou, Shuting Pan, S.~S. Li, Shuang Zhou, Shaoqing Wu,
  Tao Yun, Tian Pei, Tianyu Sun, T.~Wang, Wangding Zeng, Wen Liu, Wenfeng
  Liang, Wenjun Gao, Wenqin Yu, Wentao Zhang, W.~L. Xiao, Wei An, Xiaodong Liu,
  Xiaohan Wang, Xiaokang Chen, Xiaotao Nie, Xin Cheng, Xin Liu, Xin Xie,
  Xingchao Liu, Xinyu Yang, Xinyuan Li, Xuecheng Su, Xuheng Lin, X.~Q. Li,
  Xiangyue Jin, Xiaojin Shen, Xiaosha Chen, Xiaowen Sun, Xiaoxiang Wang, Xinnan
  Song, Xinyi Zhou, Xianzu Wang, Xinxia Shan, Y.~K. Li, Y.~Q. Wang, Y.~X. Wei,
  Yang Zhang, Yanhong Xu, Yao Li, Yao Zhao, Yaofeng Sun, Yaohui Wang, Yi~Yu,
  Yichao Zhang, Yifan Shi, Yiliang Xiong, Ying He, Yishi Piao, Yisong Wang,
  Yixuan Tan, Yiyang Ma, Yiyuan Liu, Yongqiang Guo, Yuan Ou, Yuduan Wang, Yue
  Gong, Yuheng Zou, Yujia He, Yunfan Xiong, Yuxiang Luo, Yuxiang You, Yuxuan
  Liu, Yuyang Zhou, Y.~X. Zhu, Yanping Huang, Yaohui Li, Yi~Zheng, Yuchen Zhu,
  Yunxian Ma, Ying Tang, Yukun Zha, Yuting Yan, Z.~Z. Ren, Zehui Ren, Zhangli
  Sha, Zhe Fu, Zhean Xu, Zhenda Xie, Zhengyan Zhang, Zhewen Hao, Zhicheng Ma,
  Zhigang Yan, Zhiyu Wu, Zihui Gu, Zijia Zhu, Zijun Liu, Zilin Li, Ziwei Xie,
  Ziyang Song, Zizheng Pan, Zhen Huang, Zhipeng Xu, Zhongyu Zhang, and Zhen
  Zhang.
\newblock {DeepSeek-R1} incentivizes reasoning in {LLMs} through reinforcement
  learning.
\newblock \emph{Nature}, 645:\penalty0 633--638, 2025.
\newblock \doi{10.1038/s41586-025-09422-z}.
\newblock URL \url{https://doi.org/10.1038/s41586-025-09422-z}.

\bibitem[GX-Chen et~al.(2026)GX-Chen, Prakash, Guo, Fergus, and
  Ranganath]{chen2025tilting}
Anthony GX-Chen, Jatin Prakash, Jeff Guo, Rob Fergus, and Rajesh Ranganath.
\newblock {KL}-regularized reinforcement learning for generative modelling is
  designed to mode collapse.
\newblock In \emph{International Conference on Learning Representations},
  volume 2026, pp.\  50665--50693, 2026.
\newblock URL
  \url{https://proceedings.iclr.cc/paper_files/paper/2026/hash/5306f6ce1082d29e4968acdacab84fd7-Abstract-Conference.html}.

\bibitem[Hariri et~al.(2026)Hariri, Samandar, Hinczewski, and
  Chaudhary]{hariri2025bayes}
Mohsen Hariri, Amirhossein Samandar, Michael Hinczewski, and Vipin Chaudhary.
\newblock Don't pass@k: A {Bayesian} framework for large language model
  evaluation.
\newblock In \emph{The Fourteenth International Conference on Learning
  Representations}, 2026.
\newblock URL \url{https://openreview.net/forum?id=PTXi3Ef4sT}.

\bibitem[He et~al.(2024)He, Luo, Bai, Hu, Thai, Shen, Hu, Han, Huang, Zhang,
  Liu, Qi, Liu, and Sun]{he2024olympiadbench}
Chaoqun He, Renjie Luo, Yuzhuo Bai, Shengding Hu, Zhen Thai, Junhao Shen, Jinyi
  Hu, Xu~Han, Yujie Huang, Yuxiang Zhang, Jie Liu, Lei Qi, Zhiyuan Liu, and
  Maosong Sun.
\newblock {O}lympiad{B}ench: A challenging benchmark for promoting {AGI} with
  olympiad-level bilingual multimodal scientific problems.
\newblock In Lun-Wei Ku, Andre Martins, and Vivek Srikumar (eds.),
  \emph{Proceedings of the 62nd Annual Meeting of the Association for
  Computational Linguistics (Volume 1: Long Papers)}, pp.\  3828--3850,
  Bangkok, Thailand, August 2024. Association for Computational Linguistics.
\newblock \doi{10.18653/v1/2024.acl-long.211}.
\newblock URL \url{https://aclanthology.org/2024.acl-long.211/}.

\bibitem[He et~al.(2025)He, Liu, Liu, Yan, Wang, Cheng, Zhang, Zhang, Xu, Shen,
  Li, Zeng, Wei, Cheng, An, Liu, and Zhou]{he2025skywork}
Jujie He, Jiacai Liu, Chris~Yuhao Liu, Rui Yan, Chaojie Wang, Peng Cheng,
  Xiaoyu Zhang, Fuxiang Zhang, Jiacheng Xu, Wei Shen, Siyuan Li, Liang Zeng,
  Tianwen Wei, Cheng Cheng, Bo~An, Yang Liu, and Yahui Zhou.
\newblock {Skywork} open reasoner 1 technical report.
\newblock \emph{arXiv preprint arXiv:2505.22312}, 2025.
\newblock \doi{10.48550/arXiv.2505.22312}.
\newblock URL \url{https://arxiv.org/abs/2505.22312}.

\bibitem[Hendrycks et~al.(2021)Hendrycks, Burns, Kadavath, Arora, Basart, Tang,
  Song, and Steinhardt]{hendrycks2021math}
Dan Hendrycks, Collin Burns, Saurav Kadavath, Akul Arora, Steven Basart, Eric
  Tang, Dawn Song, and Jacob Steinhardt.
\newblock Measuring mathematical problem solving with the {MATH} dataset.
\newblock In \emph{Proceedings of the Neural Information Processing Systems
  Track on Datasets and Benchmarks}, volume~1, 2021.
\newblock URL
  \url{https://datasets-benchmarks-proceedings.neurips.cc/paper_files/paper/2021/hash/be83ab3ecd0db773eb2dc1b0a17836a1-Abstract-round2.html}.

\bibitem[Hu et~al.(2025)Hu, Zhang, Han, Jiang, Zhang, and Shum]{hu2025orz}
Jingcheng Hu, Yinmin Zhang, Qi~Han, Daxin Jiang, Xiangyu Zhang, and Heung-Yeung
  Shum.
\newblock {Open-Reasoner-Zero}: An open source approach to scaling up
  reinforcement learning on the base model.
\newblock In \emph{Advances in Neural Information Processing Systems},
  volume~38, pp.\  162239--162262. Curran Associates, Inc., 2025.
\newblock \doi{10.52202/085713-5418}.
\newblock URL
  \url{https://proceedings.neurips.cc/paper_files/paper/2025/file/ed873d79e7c268c020c4b4db13a2812a-Paper-Conference.pdf}.

\bibitem[Huang et~al.(2025)Huang, Block, Foster, Rohatgi, Zhang, Simchowitz,
  Ash, and Krishnamurthy]{huang2025sharpening}
Audrey Huang, Adam Block, Dylan~J. Foster, Dhruv Rohatgi, Cyril Zhang, Max
  Simchowitz, Jordan~T. Ash, and Akshay Krishnamurthy.
\newblock Self-improvement in language models: The sharpening mechanism.
\newblock In \emph{International Conference on Learning Representations},
  volume 2025, pp.\  76687--76739, 2025.
\newblock URL
  \url{https://proceedings.iclr.cc/paper_files/paper/2025/hash/bee8c2bc757f6bbc3efd7cf1b979f0c9-Abstract-Conference.html}.

\bibitem[Jiang et~al.(2023)Jiang, Sablayrolles, Mensch, Bamford, Chaplot,
  Casas, Bressand, Lengyel, Lample, Saulnier, Lavaud, Lachaux, Stock, Scao,
  Lavril, Wang, Lacroix, and Sayed]{jiang2023mistral}
Albert~Q. Jiang, Alexandre Sablayrolles, Arthur Mensch, Chris Bamford,
  Devendra~Singh Chaplot, Diego de~las Casas, Florian Bressand, Gianna Lengyel,
  Guillaume Lample, Lucile Saulnier, L{\'{e}}lio~Renard Lavaud, Marie-Anne
  Lachaux, Pierre Stock, Teven~Le Scao, Thibaut Lavril, Thomas Wang,
  Timoth{\'{e}}e Lacroix, and William~El Sayed.
\newblock {Mistral} 7{B}.
\newblock \emph{arXiv preprint arXiv:2310.06825}, 2023.
\newblock \doi{10.48550/arXiv.2310.06825}.
\newblock URL \url{https://arxiv.org/abs/2310.06825}.

\bibitem[Karlin(1968)]{karlin1968total}
Samuel Karlin.
\newblock \emph{Total Positivity}, volume~1.
\newblock Stanford University Press, 1968.

\bibitem[Kazdan et~al.(2025)Kazdan, Schaeffer, Allouah, Sullivan, Yu, Levi, and
  Koyejo]{kazdan2025passk}
Joshua Kazdan, Rylan Schaeffer, Youssef Allouah, Colin Sullivan, Kyssen Yu,
  Noam Levi, and Sanmi Koyejo.
\newblock Efficient prediction of pass@k scaling in large language models.
\newblock \emph{arXiv preprint arXiv:2510.05197}, 2025.
\newblock \doi{10.48550/arXiv.2510.05197}.
\newblock URL \url{https://arxiv.org/abs/2510.05197}.

\bibitem[Kiefer \& Wolfowitz(1956)Kiefer and Wolfowitz]{kiefer1956}
J.~Kiefer and J.~Wolfowitz.
\newblock Consistency of the maximum likelihood estimator in the presence of
  infinitely many incidental parameters.
\newblock \emph{The Annals of Mathematical Statistics}, 27\penalty0
  (4):\penalty0 887--906, 1956.
\newblock \doi{10.1214/aoms/1177728066}.

\bibitem[Kim et~al.(2025)Kim, Shrestha, Shrestha, Nepal, and Ross]{kim2025rlvd}
Minwu Kim, Anubhav Shrestha, Safal Shrestha, Aadim Nepal, and Keith Ross.
\newblock Reinforcement learning vs.\ distillation: Understanding accuracy and
  capability in {LLM} reasoning.
\newblock \emph{arXiv preprint arXiv:2505.14216}, 2025.
\newblock \doi{10.48550/arXiv.2505.14216}.
\newblock URL \url{https://arxiv.org/abs/2505.14216}.

\bibitem[Kwon et~al.(2023)Kwon, Li, Zhuang, Sheng, Zheng, Yu, Gonzalez, Zhang,
  and Stoica]{kwon2023vllm}
Woosuk Kwon, Zhuohan Li, Siyuan Zhuang, Ying Sheng, Lianmin Zheng, Cody~Hao Yu,
  Joseph Gonzalez, Hao Zhang, and Ion Stoica.
\newblock Efficient memory management for large language model serving with
  {PagedAttention}.
\newblock In \emph{Proceedings of the 29th Symposium on Operating Systems
  Principles}, pp.\  611--626, 2023.
\newblock \doi{10.1145/3600006.3613165}.
\newblock URL \url{https://doi.org/10.1145/3600006.3613165}.

\bibitem[Kydl{\'{i}}{\v{c}}ek(2025)]{mathverify2025}
Hynek Kydl{\'{i}}{\v{c}}ek.
\newblock {Math-Verify}: Math verification library.
\newblock \url{https://github.com/huggingface/Math-Verify}, 2025.
\newblock URL \url{https://github.com/huggingface/Math-Verify}.

\bibitem[Levi(2026)]{levi2026hardtail}
Noam Levi.
\newblock Learning shrinks the hard tail: Training-dependent inference scaling
  in a solvable linear model.
\newblock In \emph{The Fourteenth International Conference on Learning
  Representations}, 2026.
\newblock URL \url{https://iclr.cc/virtual/2026/poster/10010127}.

\bibitem[Liao et~al.(2024)Liao, Li, Luo, Wu, and Fan]{liao2024mario}
Minpeng Liao, Chengxi Li, Wei Luo, Jing Wu, and Kai Fan.
\newblock {MARIO}: {MA}th reasoning with code interpreter output - a
  reproducible pipeline.
\newblock In Lun-Wei Ku, Andre Martins, and Vivek Srikumar (eds.),
  \emph{Findings of the Association for Computational Linguistics: ACL 2024},
  pp.\  905--924, Bangkok, Thailand, August 2024. Association for Computational
  Linguistics.
\newblock \doi{10.18653/v1/2024.findings-acl.53}.
\newblock URL \url{https://aclanthology.org/2024.findings-acl.53/}.

\bibitem[Lightman et~al.(2024)Lightman, Kosaraju, Burda, Edwards, Baker, Lee,
  Leike, Schulman, Sutskever, and Cobbe]{lightman2024verify}
Hunter Lightman, Vineet Kosaraju, Yura Burda, Harri Edwards, Bowen Baker, Teddy
  Lee, Jan Leike, John Schulman, Ilya Sutskever, and Karl Cobbe.
\newblock Let's verify step by step.
\newblock In \emph{The Twelfth International Conference on Learning
  Representations}, 2024.
\newblock URL \url{https://openreview.net/forum?id=v8L0pN6EOi}.

\bibitem[Liu et~al.(2023)Liu, Xia, Wang, and Zhang]{liu2023evalplus}
Jiawei Liu, Chunqiu~Steven Xia, Yuyao Wang, and Lingming Zhang.
\newblock Is your code generated by {ChatGPT} really correct? rigorous
  evaluation of large language models for code generation.
\newblock In \emph{Advances in Neural Information Processing Systems},
  volume~36, pp.\  21558--21572. Curran Associates, Inc., 2023.
\newblock \doi{10.52202/075280-0943}.
\newblock URL
  \url{https://proceedings.neurips.cc/paper_files/paper/2023/file/43e9d647ccd3e4b7b5baab53f0368686-Paper-Conference.pdf}.

\bibitem[Liu et~al.(2025)Liu, Diao, Lu, Hu, Dong, Choi, Kautz, and
  Dong]{liu2025prorl}
Mingjie Liu, Shizhe Diao, Ximing Lu, Jian Hu, Xin Dong, Yejin Choi, Jan Kautz,
  and Yi~Dong.
\newblock {ProRL}: Prolonged reinforcement learning expands reasoning
  boundaries in large language models.
\newblock In \emph{Advances in Neural Information Processing Systems},
  volume~38, pp.\  17998--18031. Curran Associates, Inc., 2025.
\newblock \doi{10.52202/085713-0608}.
\newblock URL
  \url{https://proceedings.neurips.cc/paper_files/paper/2025/hash/1a22b912945fb7c0bdd079e792b31b6f-Abstract-Conference.html}.

\bibitem[Liu(2026)]{liu2026nofreelunch}
Yinuo Liu.
\newblock There is no free lunch in {RLVR}: Every pass@1 gain has a pass@k
  price.
\newblock \emph{arXiv preprint arXiv:2610.39295}, 2026.
\newblock \doi{10.48550/arXiv.2610.39295}.
\newblock URL \url{https://arxiv.org/abs/2610.39295}.

\bibitem[Luo et~al.(2025)Luo, Tan, Wong, Shi, Tang, Roongta, Cai, Luo, Li,
  Popa, and Stoica]{luo2025deepscaler}
Michael Luo, Sijun Tan, Justin Wong, Xiaoxiang Shi, William~Y. Tang, Manan
  Roongta, Colin Cai, Jeffrey Luo, Li~Erran Li, Raluca~Ada Popa, and Ion
  Stoica.
\newblock {DeepScaleR}: Surpassing {O1-Preview} with a 1.5{B} model by scaling
  {RL}.
\newblock Notion blog,
  \url{https://pretty-radio-b75.notion.site/DeepScaleR-Surpassing-O1-Preview-with-a-1-5B-Model-by-Scaling-RL-19681902c1468005bed8ca303013a4e2},
  2025.

\bibitem[Miller(2024)]{miller2024errorbars}
Evan Miller.
\newblock Adding error bars to evals: A statistical approach to language model
  evaluations.
\newblock \emph{arXiv preprint arXiv:2411.00640}, 2024.
\newblock \doi{10.48550/arXiv.2411.00640}.
\newblock URL \url{https://arxiv.org/abs/2411.00640}.

\bibitem[Mroueh(2026)]{mroueh2025grpo}
Youssef Mroueh.
\newblock Reinforcement learning with verifiable rewards: {GRPO}'s loss,
  dynamics, and success amplification.
\newblock In \emph{Proceedings of the 43rd International Conference on Machine
  Learning}, volume 306 of \emph{Proceedings of Machine Learning Research},
  pp.\  90821--90847. PMLR, 2026.
\newblock URL \url{https://proceedings.mlr.press/v306/mroueh26a.html}.

\bibitem[Nguyen et~al.(2025)Nguyen, La, Nguyen, Chawla, Nguyen, and
  Doan]{nguyen2025boundary}
Phuc~Minh Nguyen, Chinh~D. La, Duy M.~H. Nguyen, Nitesh~V. Chawla, Binh~T.
  Nguyen, and Khoa~D. Doan.
\newblock The reasoning boundary paradox: How reinforcement learning constrains
  language models.
\newblock \emph{arXiv preprint arXiv:2510.02230}, 2025.
\newblock \doi{10.48550/arXiv.2510.02230}.
\newblock URL \url{https://arxiv.org/abs/2510.02230}.

\bibitem[Polson et~al.(2026)Polson, Sokolov, and Zantedeschi]{polson2026rlvr}
Nicholas~G. Polson, Vadim Sokolov, and Daniel Zantedeschi.
\newblock Quantile {GBC} for {RL} with verifiable rewards.
\newblock \emph{SSRN Working Paper 6970879}, 2026.
\newblock \doi{10.2139/ssrn.6970879}.
\newblock URL
  \url{https://papers.ssrn.com/sol3/papers.cfm?abstract_id=6970879}.

\bibitem[{Qwen} et~al.(2024){Qwen}, Yang, Yang, Zhang, Hui, Zheng, Yu, Li, Liu,
  Huang, Wei, Lin, Yang, Tu, Zhang, Yang, Yang, Zhou, Lin, Dang, Lu, Bao, Yang,
  Yu, Li, Xue, Zhang, Zhu, Men, Lin, Li, Tang, Xia, Ren, Ren, Fan, Su, Zhang,
  Wan, Liu, Cui, Zhang, and Qiu]{yang2024qwen25}
{Qwen}, An~Yang, Baosong Yang, Beichen Zhang, Binyuan Hui, Bo~Zheng, Bowen Yu,
  Chengyuan Li, Dayiheng Liu, Fei Huang, Haoran Wei, Huan Lin, Jian Yang,
  Jianhong Tu, Jianwei Zhang, Jianxin Yang, Jiaxi Yang, Jingren Zhou, Junyang
  Lin, Kai Dang, Keming Lu, Keqin Bao, Kexin Yang, Le~Yu, Mei Li, Mingfeng Xue,
  Pei Zhang, Qin Zhu, Rui Men, Runji Lin, Tianhao Li, Tianyi Tang, Tingyu Xia,
  Xingzhang Ren, Xuancheng Ren, Yang Fan, Yang Su, Yichang Zhang, Yu~Wan,
  Yuqiong Liu, Zeyu Cui, Zhenru Zhang, and Zihan Qiu.
\newblock {Qwen2.5} technical report.
\newblock \emph{arXiv preprint arXiv:2412.15115}, 2024.
\newblock \doi{10.48550/arXiv.2412.15115}.
\newblock URL \url{https://arxiv.org/abs/2412.15115}.

\bibitem[Schaeffer et~al.(2025)Schaeffer, Kazdan, Hughes, Juravsky, Price,
  Lynch, Jones, Kirk, Mirhoseini, and Koyejo]{schaeffer2025monkeys}
Rylan Schaeffer, Joshua Kazdan, John Hughes, Jordan Juravsky, Sara Price,
  Aengus Lynch, Erik Jones, Robert Kirk, Azalia Mirhoseini, and Sanmi Koyejo.
\newblock How do large language monkeys get their power ({L}aws)?
\newblock In \emph{Proceedings of the 42nd International Conference on Machine
  Learning}, volume 267 of \emph{Proceedings of Machine Learning Research},
  pp.\  53132--53176. PMLR, 2025.
\newblock URL \url{https://proceedings.mlr.press/v267/schaeffer25a.html}.

\bibitem[Shao et~al.(2024)Shao, Wang, Zhu, Xu, Song, Bi, Zhang, Zhang, Li, Wu,
  and Guo]{shao2024deepseekmath}
Zhihong Shao, Peiyi Wang, Qihao Zhu, Runxin Xu, Junxiao Song, Xiao Bi, Haowei
  Zhang, Mingchuan Zhang, Y.~K. Li, Y.~Wu, and Daya Guo.
\newblock {DeepSeekMath}: Pushing the limits of mathematical reasoning in open
  language models.
\newblock \emph{arXiv preprint arXiv:2402.03300}, 2024.
\newblock \doi{10.48550/arXiv.2402.03300}.
\newblock URL \url{https://arxiv.org/abs/2402.03300}.

\bibitem[Sheng et~al.(2025)Sheng, Zhang, Ye, Wu, Zhang, Zhang, Peng, Lin, and
  Wu]{sheng2025hybridflow}
Guangming Sheng, Chi Zhang, Zilingfeng Ye, Xibin Wu, Wang Zhang, Ru~Zhang,
  Yanghua Peng, Haibin Lin, and Chuan Wu.
\newblock {HybridFlow}: A flexible and efficient {RLHF} framework.
\newblock In \emph{Proceedings of the Twentieth European Conference on Computer
  Systems}, pp.\  1279--1297, 2025.
\newblock \doi{10.1145/3689031.3696075}.
\newblock URL \url{https://doi.org/10.1145/3689031.3696075}.

\bibitem[Strozzi(2026)]{strozzi2026teacher}
Igor~Lima Strozzi.
\newblock Teacher-free self-training amplifies but does not compound: A
  pass@{$K$} crossover on a free-verifier domain.
\newblock \emph{arXiv preprint arXiv:2606.07856}, 2026.
\newblock \doi{10.48550/arXiv.2606.07856}.
\newblock URL \url{https://arxiv.org/abs/2606.07856}.

\bibitem[Teicher(1963)]{teicher1963}
Henry Teicher.
\newblock Identifiability of finite mixtures.
\newblock \emph{The Annals of Mathematical Statistics}, 34\penalty0
  (4):\penalty0 1265--1269, 1963.
\newblock \doi{10.1214/aoms/1177703862}.

\bibitem[Tomihari \& Sato(2026)Tomihari and Sato]{tomihari2026power}
Akiyoshi Tomihari and Issei Sato.
\newblock Power distribution bridges sampling, self-reward {RL}, and
  self-distillation.
\newblock \emph{arXiv preprint arXiv:2605.04542}, 2026.
\newblock \doi{10.48550/arXiv.2605.04542}.
\newblock URL \url{https://arxiv.org/abs/2605.04542}.

\bibitem[van~der Vaart \& Wellner(1996)van~der Vaart and
  Wellner]{vaart1996weak}
Aad~W. van~der Vaart and Jon~A. Wellner.
\newblock \emph{Weak Convergence and Empirical Processes: With Applications to
  Statistics}.
\newblock Springer Series in Statistics. Springer New York, 1996.
\newblock ISBN 978-1-4757-2547-6.
\newblock \doi{10.1007/978-1-4757-2545-2}.

\bibitem[Wen et~al.(2026)Wen, Liu, Zheng, Ye, Wu, Wang, Xu, Liang, Li, Miao,
  Bian, and Yang]{wen2025rlvr}
Xumeng Wen, Zihan Liu, Shun Zheng, Shengyu Ye, Zhirong Wu, Yang Wang, Zhijian
  Xu, Xiao Liang, Junjie Li, Ziming Miao, Jiang Bian, and Mao Yang.
\newblock Reinforcement learning with verifiable rewards implicitly
  incentivizes correct reasoning in base {LLMs}.
\newblock In \emph{The Fourteenth International Conference on Learning
  Representations}, 2026.
\newblock URL \url{https://iclr.cc/virtual/2026/poster/10007896}.

\bibitem[Wu et~al.(2025)Wu, Xuan, Lu, Liu, Dong, Harchaoui, and
  Choi]{wu2025leash}
Fang Wu, Weihao Xuan, Ximing Lu, Mingjie Liu, Yi~Dong, Zaid Harchaoui, and
  Yejin Choi.
\newblock The invisible leash: Why {RLVR} may or may not escape its origin.
\newblock \emph{arXiv preprint arXiv:2507.14843}, 2025.
\newblock \doi{10.48550/arXiv.2507.14843}.
\newblock URL \url{https://arxiv.org/abs/2507.14843}.

\bibitem[Yan et~al.(2025)Yan, Li, Hu, Wang, Cui, Qu, Cheng, and
  Zhang]{yan2025luffy}
Jianhao Yan, Yafu Li, Zican Hu, Zhi Wang, Ganqu Cui, Xiaoye Qu, Yu~Cheng, and
  Yue Zhang.
\newblock Learning to reason under off-policy guidance.
\newblock In \emph{Advances in Neural Information Processing Systems},
  volume~38, pp.\  117157--117186. Curran Associates, Inc., 2025.
\newblock \doi{10.52202/085713-3904}.
\newblock URL
  \url{https://proceedings.neurips.cc/paper_files/paper/2025/file/a9d5c33e38442450f9781bd7bad3c81e-Paper-Conference.pdf}.

\bibitem[Yang et~al.(2024)Yang, Zhang, Hui, Gao, Yu, Li, Liu, Tu, Zhou, Lin,
  Lu, Xue, Lin, Liu, Ren, and Zhang]{yang2024qwen25math}
An~Yang, Beichen Zhang, Binyuan Hui, Bofei Gao, Bowen Yu, Chengpeng Li,
  Dayiheng Liu, Jianhong Tu, Jingren Zhou, Junyang Lin, Keming Lu, Mingfeng
  Xue, Runji Lin, Tianyu Liu, Xingzhang Ren, and Zhenru Zhang.
\newblock {Qwen2.5-Math} technical report: Toward mathematical expert model via
  self-improvement.
\newblock \emph{arXiv preprint arXiv:2409.12122}, 2024.
\newblock \doi{10.48550/arXiv.2409.12122}.
\newblock URL \url{https://arxiv.org/abs/2409.12122}.

\bibitem[Yuan et~al.(2026)Yuan, Chen, Zheng, Li, Feng, Wang, Xiang, Liu, and
  An]{yuan2026overtraining}
Suqin Yuan, Jinkun Chen, Jiyang Zheng, Muyang Li, Lei Feng, Dadong Wang, Tao
  Xiang, Tongliang Liu, and Bo~An.
\newblock Understanding diversity collapse in {RLVR} via the lens of
  overtraining.
\newblock \emph{arXiv preprint arXiv:2606.15455}, 2026.
\newblock \doi{10.48550/arXiv.2606.15455}.
\newblock URL \url{https://arxiv.org/abs/2606.15455}.

\bibitem[Yue et~al.(2025)Yue, Chen, Lu, Zhao, Wang, Yue, Song, and
  Huang]{yue2025limit}
Yang Yue, Zhiqi Chen, Rui Lu, Andrew Zhao, Zhaokai Wang, Yang Yue, Shiji Song,
  and Gao Huang.
\newblock Does reinforcement learning really incentivize reasoning capacity in
  {LLMs} beyond the base model?
\newblock In \emph{Advances in Neural Information Processing Systems},
  volume~38, pp.\  57654--57689. Curran Associates, Inc., 2025.
\newblock \doi{10.52202/085713-1933}.
\newblock URL
  \url{https://proceedings.neurips.cc/paper_files/paper/2025/hash/537d5aa768c2d534016a4d06f87bc8fb-Abstract-Conference.html}.

\bibitem[Zeng et~al.(2025)Zeng, Huang, Liu, Liu, He, Ma, and
  He]{zeng2025simplerl}
Weihao Zeng, Yuzhen Huang, Qian Liu, Wei Liu, Keqing He, Zejun Ma, and Junxian
  He.
\newblock {SimpleRL-Zoo}: Investigating and taming zero reinforcement learning
  for open base models in the wild.
\newblock In \emph{Second Conference on Language Modeling}, 2025.
\newblock URL \url{https://openreview.net/forum?id=vSMCBUgrQj}.

\bibitem[Zhou(2026)]{zhou2026inversion}
Todd Zhou.
\newblock When {RLVR} shrinks the reasoning boundary: Diagnosing pass@k
  inversion.
\newblock \emph{arXiv preprint arXiv:2607.20543}, 2026.
\newblock \doi{10.48550/arXiv.2607.20543}.
\newblock URL \url{https://arxiv.org/abs/2607.20543}.

\end{thebibliography}
\bibliographystyle{iclr2027_conference}

\appendix
\section{Mathematical Proofs and Notation Reference}\label{app:proofs}

Table~\ref{tab:notation} summarizes the notation and estimators used across
Part~I (Sections~\ref{sec:infer}--\ref{sec:experiments}) and Part~II
(Sections~\ref{sec:model}--\ref{sec:planned}) of the main text.
Throughout this appendix, we write $\theta_p=\logit p=\log\{p/(1-p)\}$,
$\sigma(u)=(1+e^{-u})^{-1}$ for the logistic function,
$T_{\beta,\gamma}(p)=\sigma(\beta\theta_p+\gamma)$, $u(p)=(1-p)^G$, and
$\rho(p)=\pi_0 u(p)$.

\begin{table}[!ht]
\centering
\footnotesize
\setlength{\tabcolsep}{4pt}
\caption{\textbf{Summary of notation and estimators across Part~I (Inference)
and Part~II (Mechanism).}}
\label{tab:notation}
\begin{tabular}{@{}lp{11.2cm}@{}}
\toprule
Symbol & Definition and role \\
\midrule
\multicolumn{2}{@{}l}{\emph{Part~I: Paired pass@$k$ inference (Sections~\ref{sec:infer}--\ref{sec:experiments})}} \\
$(P_i,Q_i),\,(c_i,d_i)$ & Prompt $i$'s latent success rates before/after RLVR and observed success counts out of $(n_i,n_i')$ rollouts \\
$\psi_i(k),\,\hat D(k)$ & Prompt-level unbiased pass@$k$ difference and empirical mean curve $m^{-1}\sum_{i=1}^m\psi_i(k)$ \\
$D(k)$ & Population pass@$k$ difference $\mathbb E[(1-P)^k-(1-Q)^k]$ \\
$\kstar$ & Empirical (or model-implied) point-estimate crossover budget where $\hat D(k)$ first turns negative \\
$\kappa,\,[\underline\kappa,\overline\kappa]$ & Population first-loss budget $\inf\{k\ge1:D(k)<0\}$ and its $(1-\alpha)$ simultaneous confidence interval (Proposition~\ref{prop:cert}) \\
\midrule
\multicolumn{2}{@{}l}{\emph{Part~II: Operator vs.\ kernel mechanism (Sections~\ref{sec:model}--\ref{sec:planned})}} \\
$H(q\mid p),\,T(p)$ & Conditional transition law of $Q\mid P=p$ (Markov kernel) and its conditional mean $T(p)=\mathbb E[Q\mid P=p]$ \\
$\hat\beta_{\mathrm{proj}}$ & Apparent logit slope estimated under the misspecified single-curve projection $Q=T(P)$ \\
$(\beta,\gamma),\,\phi,\,\pi_0$ & Conditional-mean logit slope and intercept, Beta-Binomial precision parameter, and excess-zero weight \\
$\pstar$ & Break-even base difficulty satisfying $T(\pstar)=\pstar$, i.e., $\logit\pstar=\gamma/(1-\beta)$ \\
$\widehat{\Cov}_{\mathrm{split}},\,\widehat{\mathrm{sd}}$ & Split-generation residual covariance estimator~\eqref{eq:split-stat} and conditional standard deviation $\sqrt{(\widehat{\Cov}_{\mathrm{split}})_+}\approx\sqrt{\mathbb E[\Var(Q\mid P)]}$ \\
$\Lambda_{\min}(n_0)$ & Distribution-free lower bound on the Lipschitz constant $\sup_{p}|T'(p)|$ of any deterministic link at depth $n_0$ (Theorem~\ref{thm:slope}) \\
$L,\,R$ & Locked-prompt set $\{i:c_i>0,d_i=0\}$ and newly-solved rescued set $\{i:c_i=0,d_i>0\}$, obeying $\hat D(n)=(|R|-|L|)/m$ \\
\bottomrule
\end{tabular}
\end{table}

\subsection{Proof of Theorem~\ref{thm:test} (Dominance Test)}

\paragraph{Notation and weak convergence.} Fix a maximum budget $K$. For
$k\le K$, write
$\psi_i(k)=\widehat{\passk}^{\mathrm{RL}}_i-\widehat{\passk}^{\mathrm{base}}_i$,
$D(k)=\mathbb E[\psi_i(k)]$, and $\hat D(k)=m^{-1}\sum_{i=1}^m\psi_i(k)$. Let
$\Sigma=\Var(\psi_i)$ be the $K\times K$ covariance matrix of
$\psi_i=(\psi_i(1),\dots,\psi_i(K))^\top$, with diagonal entries
$s(k)^2=\Sigma_{kk}$, and let
\[
\hat s(k)^2 = \frac{1}{m-1}\sum_{i=1}^m\big\{\psi_i(k)-\hat D(k)\big\}^2,
\qquad
\hat\sigma(k) = \frac{\hat s(k)}{\sqrt{m}}.
\]
Under the standing assumption $\min_{1\le k\le K}s(k)>0$, and because $K$ is
fixed and $\psi_i\in[-1,1]^K$ is bounded, the multivariate central limit
theorem and Cram\'er--Wold device yield
$Z_m:=\sqrt{m}(\hat D-D)\rightsquigarrow Z\sim N(0,\Sigma)$, while
$\max_{1\le k\le K}|\hat s(k)-s(k)|\to_p0$. Consequently, by the continuous
mapping theorem, the centered minimum statistic satisfies
\[
\mathcal T_m^0 := \min_{1\le k\le K}\frac{\hat D(k)-D(k)}{\hat\sigma(k)}
= \min_{1\le k\le K}\frac{Z_m(k)}{\hat s(k)}
\;\rightsquigarrow\;
\mathcal T^0_P := \min_{1\le k\le K}\frac{Z(k)}{s(k)}.
\]
The subscript $P$ emphasizes that the limiting distribution $\mathcal T^0_P$
depends on the across-$k$ correlation structure of $\psi_i$ under $P$, which
is not determined by the null hypothesis $H_0$.

\paragraph{Pathwise dominance of the all-binding boundary.} Under the null
hypothesis $H_0:D(k)\ge0$ for all $k\le K$, we have pathwise at every $k$:
\[
\frac{\hat D(k)}{\hat\sigma(k)}
= \frac{\hat D(k)-D(k)}{\hat\sigma(k)} + \frac{D(k)}{\hat\sigma(k)}
\;\ge\; \frac{\hat D(k)-D(k)}{\hat\sigma(k)},
\]
and therefore $\mathcal T_m\ge\mathcal T_m^0$. Thus, for any threshold $q$
(including the data-dependent bootstrap quantile $\hat q_\alpha$),
$\{\mathcal T_m<q\}\subseteq\{\mathcal T_m^0<q\}$. Equality holds when
$D\equiv0$, which fixes the marginal means to zero while leaving the
centered covariance $\Sigma$ arbitrary. To see why $\Sigma$ must be
estimated from the data, note that for $K=2$ with unit variances, perfectly
correlated components $(Z(1),Z(2))=(X,X)$ give $\mathcal T^0_P=X\sim N(0,1)$,
whereas antithetic components $(X,-X)$ give $\mathcal T^0_P=-|X|$. Sharing
one multiplier per prompt across all $k$ preserves the empirical covariance
of $\psi_i$.

\paragraph{Validity of the prompt-level multiplier bootstrap.} Let
$g_1,\dots,g_m$ be i.i.d.\ standard normal variables independent of the
data, and set
\[
Z_m^*(k) = \frac{1}{\sqrt{m}}\sum_{i=1}^m g_i\big\{\psi_i(k)-\hat D(k)\big\},
\qquad
\mathcal T_m^* = \min_{1\le k\le K}\frac{Z_m^*(k)}{\hat s(k)}.
\]
By the conditional multiplier central limit theorem
\citep[Lemma~2.9.5]{vaart1996weak}, $Z_m^*\rightsquigarrow_{\mathbb P}N(0,\Sigma)$
conditionally on the data in probability, and hence
$\mathcal T_m^*\rightsquigarrow_{\mathbb P}\mathcal T^0_P$. It remains to
verify that the $\alpha$-quantile $q_\alpha$ of $\mathcal T^0_P$ is unique
(i.e., the distribution function of $\mathcal T^0_P$ is continuous and
strictly increasing at $q_\alpha$) without imposing any rank condition on
$\Sigma$, which is essential because adjacent pass@$k$ differences are
nearly collinear. First, each marginal $Z(k)/s(k)\sim N(0,1)$ is atomless,
so for any $t\in\mathbb R$, a union bound gives
$\Pr(\mathcal T^0_P=t)\le\sum_{k=1}^K\Pr(Z(k)/s(k)=t)=0$; hence the CDF of
$\mathcal T^0_P$ is continuous on $\mathbb R$. Second, $\mathcal T^0_P$ is a
continuous function of the Gaussian vector $Z$, whose support is a linear
subspace of $\mathbb R^K$ and therefore connected; thus the support of
$\mathcal T^0_P$ is an interval, and since
$\Pr(\mathcal T^0_P\le t)\ge\Pr(Z(1)/s(1)\le t)>0$ for all $t\in\mathbb R$
and $\Pr(\mathcal T^0_P>t)\ge\Pr(Z(k)/s(k)>t\ \forall k\le K)>0$ for all
$t<0$ (as the origin lies in the support of $Z$), the support of
$\mathcal T^0_P$ contains $(-\infty,0]$. Consequently, the CDF of
$\mathcal T^0_P$ is strictly increasing wherever it lies in $(0,1)$, so
every quantile $q_\alpha$ ($\alpha\in(0,1)$) is unique. Therefore
$\hat q_\alpha\to_p q_\alpha$ and
$\Pr_P(\mathcal T_m^0<\hat q_\alpha)\to\alpha$, which combined with the
pathwise inclusion $\mathcal T_m\ge\mathcal T_m^0$ yields
\[
\limsup_{m\to\infty}\Pr_P\big(\mathcal T_m<\hat q_\alpha\big)
\le \lim_{m\to\infty}\Pr_P\big(\mathcal T_m^0<\hat q_\alpha\big) = \alpha
\]
for every $P\in H_0$, with equality when $D\equiv0$.

\paragraph{Consistency.} If $D(k_0)<0$ for some $k_0\le K$, then
$\mathcal T_m\le\sqrt{m}\,\hat D(k_0)/\hat s(k_0)=\sqrt{m}\,D(k_0)/s(k_0)+O_p(1)\to-\infty$,
whereas $\hat q_\alpha=O_p(1)$, so $\Pr_P(\mathcal T_m<\hat q_\alpha)\to1$.
\hfill$\square$

\subsection{Coverage and Conventions for Proposition~\ref{prop:cert}}
\label{app:prop-cert}

Applying the exact same conditional multiplier-bootstrap argument to the
two-sided studentized maximum statistic
$\max_{1\le k\le K}|\hat D(k)-D(k)|/\hat\sigma(k)$ shows that its limiting
law $\max_{1\le k\le K}|Z(k)|/s(k)$ has a continuous, strictly increasing
CDF on $(0,\infty)$, so the simultaneous confidence band
$[L(k),U(k)]=\hat D(k)\pm\hat c_{1-\alpha}\hat\sigma(k)$ satisfies
$\lim_{m\to\infty}\Pr_P(L(k)\le D(k)\le U(k)\ \forall k\le K)=1-\alpha$.
On this coverage event, if $L(j)>0$ and $U(\ell)<0$ for some $j<\ell$, then
$D(j)\ge L(j)>0$ and $D(\ell)\le U(\ell)<0$, certifying a positive-to-negative
sign change of $D$ between $j$ and $\ell$. Likewise, for every
$r<\underline\kappa$, $D(r)\ge L(r)>0$, so $\kappa=\inf\{k\ge1:D(k)<0\}\ge\underline\kappa$;
and if $\overline\kappa<\infty$, then $D(\overline\kappa)\le U(\overline\kappa)<0$,
so $\kappa\le\overline\kappa$. Thus $\kappa\in[\underline\kappa,\overline\kappa]$ holds
simultaneously on the coverage event with asymptotic probability at least
$1-\alpha$. \hfill$\square$

\subsection{Statement and Proof of Theorem~\ref{thm:ident} (Identifiability)}
\label{app:model-ident}

\begin{theorem}[Identifiability of the single-curve working model]\label{thm:ident}
Under the working model where $P_i\sim F$,
$c_i\mid P_i\sim\mathrm{Bin}(n_i,P_i)$, and
$d_i\mid(P_i=p)\sim\rho(p)\delta_0+(1-\rho(p))\mathrm{Bin}(n_i',T_{\beta,\gamma}(p))$
with $\rho(p)=\pi_0(1-p)^G$, $\pi_0\in[0,1)$, and known group size $G\ge1$,
consider the triangular-array regime $m\to\infty$ with
$\min_{i\le m}n_{i,m}\to\infty$. Assume $\operatorname{supp}(F)\cap(0,1)$
contains at least two distinct points. Then:
\begin{enumerate}
\item[(i)] $F$ is uniquely identified from the base counts $\{c_i\}$.
\item[(ii)] If $\min_{i\le m}n'_{i,m}\ge2$, then $(\beta,\gamma,\pi_0)$ is
uniquely identified from the paired counts $\{(c_i,d_i)\}$. The condition
$n_i'\ge2$ is sharp: when $n_i'\equiv1$, distinct triples
$(\beta,\gamma,\pi_0)$ can induce the exact same marginal distribution even
when $|\operatorname{supp}(F)\cap(0,1)|\ge3$.
\end{enumerate}
\end{theorem}

\paragraph{Proof of (i): Identification of $F$ from base counts.} The base counts
satisfy $c_i\sim\mathrm{Bin}(n_i,p_i)$ with $p_i\sim F$ independent of
$n_i$. For any fixed integer $k\ge1$, once $\min_i n_{i,m}\ge k$, the
unbiased U-statistic $U_k(c_i,n_i)=\binom{n_i-c_i}{k}/\binom{n_i}{k}$ has
expectation $\mu_k=\mathbb E_F[(1-P)^k]$, so the entire moment sequence
$\{\mu_k\}_{k\ge0}$ of $1-P$ on the compact interval $[0,1]$ is
consistently estimable as $m,\min_i n_{i,m}\to\infty$. Since the Hausdorff
moment problem on $[0,1]$ is determinate, $F$ is uniquely identified. (At a
fixed finite depth $n$, a binomial mixture identifies only the first $n$
moments of $F$, or equivalently the mixing law up to classes sharing those
$n$ moments; \citealp{teicher1963,kiefer1956}.)

\paragraph{(ii) Identification of $(\beta,\gamma,\pi_0)$ given $F$.} Under
the triangular array $\min_i n_{i,m}\to\infty$, $\hat p_i=c_i/n_i\to p_i$
almost surely, so for any interior support point $p^\circ\in(0,1)$ of $F$,
the conditional law of $d$ given $P=p^\circ$ is identified as the limit as
$\epsilon\downarrow0$ of the empirical law of $d_i$ among prompts with
$\hat p_i\in(p^\circ-\epsilon,p^\circ+\epsilon)$. Under the working model,
\[
d\mid(P=p) \;\sim\; \rho(p)\,\delta_0 + \big(1-\rho(p)\big)\,\mathrm{Bin}\!\big(n',T_{\beta,\gamma}(p)\big).
\]
When $n'\ge2$, the first two scaled factorial moments
$m_1(p)=\mathbb E[d\mid p]/n'=(1-\rho(p))T(p)$ and
$m_2(p)=\mathbb E[d(d-1)\mid p]/\{n'(n'-1)\}=(1-\rho(p))T(p)^2$ are both
identified, and since $T(p)\in(0,1)$ and $\rho(p)<1$ on $(0,1)$,
\[
T_{\beta,\gamma}(p) = \frac{m_2(p)}{m_1(p)}, \qquad
1 - \rho(p) = \frac{m_1(p)^2}{m_2(p)}.
\]
Evaluating $T_{\beta,\gamma}(p)$ at two distinct support points
$p_1\ne p_2$ in $(0,1)$ yields the non-singular linear system
$\logit T(p_j)=\beta\,\logit p_j+\gamma$ ($j=1,2$), which uniquely determines
$(\beta,\gamma)$, while $\pi_0=\rho(p_1)/(1-p_1)^G$ is uniquely determined
from either point.

\paragraph{Sharpness at $n'=1$ and behavior at $p=0$.} When $n'=1$, the
post-RL count $d_i\in\{0,1\}$ is Bernoulli with mean
$h(p)=[1-\pi_0(1-p)^G]T_{\beta,\gamma}(p)$, and global injectivity fails even
with three support points: for $G=4$ on support points
$p=(0.05015405, 0.27439098, 0.63422245)$, the distinct parameter triples
$(\beta,\gamma,\pi_0)=(0.27933, 2.69289, 0.82148)$ and
$(0.82954, 2.27721, 0.46072)$ produce identical success probabilities
$h(p)=(0.28714, 0.70928, 0.93125)$ to five decimal places. Finally, at any
finite base depth $n$, a point mass of $F$ at $p=0$ produces $(c_i,d_i)=(0,0)$
just as the excess-zero component does; as $\min_i n_{i,m}\to\infty$, every
prompt with $p_i>0$ eventually yields $c_i>0$ almost surely, separating
$p=0$ from $p>0$. At finite $n$, $\hat\pi_0$ is therefore interpreted as an
operational excess-zero index relative to the interior support grid.
\hfill$\square$

\subsection{Proof of Theorem~\ref{thm:crossover} (Tail Exponents and Crossover Scale~\eqref{eq:kstar})}
\label{app:model-kstar}

\paragraph{Base failure asymptotics.} Split
$G_0(k)=\int_0^\delta(1-p)^k f(p)\,dp+\int_\delta^1(1-p)^k f(p)\,dp$. On
$[\delta,1]$, $(1-p)^k\le(1-\delta)^k$ decays geometrically. On $[0,\delta]$,
since $f(p)\sim cp^{a-1}$ as $p\to0^+$, substituting $p=s/k$ gives
\[
\int_0^\delta(1-p)^k c p^{a-1}\,dp
= c k^{-a}\int_0^{k\delta}\Big(1-\frac{s}{k}\Big)^k s^{a-1}\,ds
\;\longrightarrow\; c\,\Gamma(a)\,k^{-a}
\qquad(k\to\infty)
\]
By dominated convergence, using $(1-s/k)^k\le e^{-s}$ on $[0,k]$.

\paragraph{Post-RL failure asymptotics.} Under regular variation
$T(p)\sim Cp^{\beta_{\mathrm{tail}}}$ as $p\to0^+$ (for the affine-logit
link, $\logit p=\log p+O(p)$ gives $T_{\beta,\gamma}(p)=e^\gamma p^\beta(1+O(p))$,
so $C=e^\gamma$ and $\beta_{\mathrm{tail}}=\beta$), for any $\epsilon\in(0,1)$
we may choose $\delta>0$ small enough that
$(1-\epsilon)Cp^{\beta_{\mathrm{tail}}}\le T(p)\le(1+\epsilon)Cp^{\beta_{\mathrm{tail}}}$
and $1-T(p)\ge(1+\epsilon)^{-1}$ on $(0,\delta]$. Since
$\inf_{p\in[\delta,1]}T(p)>0$, the integral of $(1-T(p))^k$ over $[\delta,1]$
is again geometrically negligible when $\pi_0=0$. On $(0,\delta]$, using
$e^{-kT/(1-T)}\le(1-T)^k\le e^{-kT}$ and substituting
$w=k(1\pm\epsilon)Cp^{\beta_{\mathrm{tail}}}$, we obtain
\begin{align*}
\int_0^\delta e^{-k(1\pm\epsilon)Cp^{\beta_{\mathrm{tail}}}}c p^{a-1}\,dp
&= \frac{c}{\beta_{\mathrm{tail}}}\big[(1\pm\epsilon)Ck\big]^{-a/\beta_{\mathrm{tail}}}
\int_0^{w(\delta)} e^{-w} w^{a/\beta_{\mathrm{tail}}-1}\,dw \\
&\sim \frac{c\,\Gamma(a/\beta_{\mathrm{tail}})}{\beta_{\mathrm{tail}}}
\big[(1\pm\epsilon)Ck\big]^{-a/\beta_{\mathrm{tail}}}.
\end{align*}
Letting $\epsilon\downarrow0$ establishes \eqref{eq:tails}. When
$\beta_{\mathrm{tail}}>1$, we have $a/\beta_{\mathrm{tail}}<a$, so
$G_1(k)/G_0(k)\to\infty$ and $D(k)=G_0(k)-G_1(k)<0$ for all sufficiently
large $k$; continuity of $k\mapsto D(k)$ on $(0,\infty)$ then guarantees a
positive-to-negative crossing whenever $D(k_0)>0$ at some finite $k_0$.
When $\pi_0>0$, $G_1(k)\downarrow\pi_0\mathbb E[(1-P)^G]>0$ while
$G_0(k)\sim c\Gamma(a)k^{-a}\to0$, yielding the atom-driven scale $\kstar_0$.

\paragraph{Proof of the $O(\pstar)$ remainder in Equation~\eqref{eq:kstar} of Theorem~\ref{thm:crossover}.}
Let $F=\mathrm{Beta}(a,b)$, $\pi_0=0$, and $T=T_{\beta,\gamma}$ with
$\beta>1$ and $\gamma=-(\beta-1)\logit\pstar\to\infty$ as $\pstar\to0$. Let
$\tilde k^\star$ be the root of the leading-order asymptote equation
$c\Gamma(a)k^{-a}=\frac{c\Gamma(a/\beta)}{\beta}(e^\gamma k)^{-a/\beta}$,
namely
\[
\log\tilde k^\star = \frac{\gamma}{\beta-1}
+ \frac{\beta}{a(\beta-1)}\log\frac{\beta\,\Gamma(a)}{\Gamma(a/\beta)}.
\]
With $v=p/(1-p)$ and $s=e^\gamma v^\beta=t/k$, the affine-logit link satisfies
$1-T(p)=(1+s)^{-1}$ exactly, and $f(p)\,dp=B(a,b)^{-1}v^{a-1}(1+v)^{-a-b}\,dv$,
so
\[
G_1(k) = \frac{(e^\gamma k)^{-a/\beta}}{\beta\,B(a,b)}
\int_0^\infty \Big(1+\frac{t}{k}\Big)^{-k} t^{a/\beta-1}
\big(1+e^{-\gamma/\beta}(t/k)^{1/\beta}\big)^{-a-b}\,dt.
\]
Using $0\le 1-(1+e^{-\gamma/\beta}(t/k)^{1/\beta})^{-a-b}\le(a+b)e^{-\gamma/\beta}(t/k)^{1/\beta}$
and $(1+t/k)^{-k}=e^{-t}(1+O(t^2/k))$, dominated convergence yields
$G_1(k)=\frac{c\Gamma(a/\beta)}{\beta}(e^\gamma k)^{-a/\beta}\{1+\rho_1(k,\gamma)\}$
with $c=B(a,b)^{-1}$ and
$|\rho_1(k,\gamma)|=O\big(k^{-1}+(e^\gamma k)^{-1/\beta}\big)$ uniformly in
$\gamma$. Similarly, Stirling's expansion of $G_0(k)=B(a,b+k)/B(a,b)$ gives
$G_0(k)=c\Gamma(a)k^{-a}\{1+O(k^{-1})\}$. Therefore
\[
\log G_0(k) - \log G_1(k)
= -a\Big(1-\frac{1}{\beta}\Big)\big(\log k - \log\tilde k^\star\big)
+ O\!\big(k^{-1} + (e^\gamma k)^{-1/\beta}\big).
\]
At $k\asymp\tilde k^\star\asymp e^{\gamma/(\beta-1)}\asymp(\pstar)^{-1}$,
both $k^{-1}$ and $(e^\gamma k)^{-1/\beta}\asymp e^{-\gamma/(\beta-1)}$ are
$O(\pstar)$. Since the crossing is unique by Theorem~\ref{thm:single}, the
true root $\kstar$ satisfies $\log\kstar=\log\tilde k^\star+O(\pstar)$.

\paragraph{Extension to a power-law support band for a kernel
(Theorem~\ref{thm:crossover}).} If $H(\cdot\mid p)$ is supported on
$[c_1p^\beta,c_2p^\beta]$ for $p\le p_0$ and $Q\ge c_1p_0^\beta$ almost
surely for $p>p_0$, monotonicity of $q\mapsto(1-q)^k$ implies
$(1-c_2p^\beta)^k\le\mathbb E[(1-Q)^k\mid P=p]\le(1-c_1p^\beta)^k$ on
$(0,p_0]$. Integrating against $F$ and applying the expansion above to the
two bounding curves $T_j(p)=c_jp^\beta$ gives
$\frac{c\Gamma(a/\beta)}{\beta}(c_2k)^{-a/\beta}(1+o(1))\le G_1(k)\le
\frac{c\Gamma(a/\beta)}{\beta}(c_1k)^{-a/\beta}(1+o(1))$, so
$G_1(k)\asymp k^{-a/\beta}$. \hfill$\square$

\subsection{Proof of Theorem~\ref{thm:single} (Single Crossing)}
\label{app:single-full}

Let $x=1-p$, $z=1-T(p)$, and $t^\star=1-\pstar$. For every $p\in[0,1]$ and
real $k>0$,
\[
x^k - z^k = k\int_0^1 t^{k-1}\Big[\mathbb 1\{z<t<x\} - \mathbb 1\{x<t<z\}\Big]\,dt.
\]
Consider first the case where $T$ crosses the diagonal from below ($T(p)\le p$
on $[0,\pstar]$ and $T(p)\ge p$ on $[\pstar,1]$, for which side-preservation
is automatic). When $p>\pstar$, we have $T(p)\ge p>\pstar$, so
$z\le x<t^\star$ and the positive indicator $(z,x)$ lies entirely inside
$[0,t^\star]$. When $p<\pstar$, we have $T(p)\le p<\pstar$, so
$z\ge x>t^\star$ and the negative indicator $(x,z)$ lies entirely inside
$[t^\star,1]$. Taking expectations over $P\sim F$ by Tonelli's theorem
defines two nonnegative finite measures $\mu_+$ and $\mu_-$ supported on
$[0,t^\star]$ and $[t^\star,1]$, respectively:
\[
\frac{D(k)}{k} = A_k - B_k, \qquad
A_k = \int_0^{t^\star} t^{k-1}\,d\mu_+(t), \qquad
B_k = \int_{t^\star}^1 t^{k-1}\,d\mu_-(t).
\]
Because $t\le t^\star$ on $\mathrm{supp}(\mu_+)$ and $t\ge t^\star$ on
$\mathrm{supp}(\mu_-)$, for every real increment $h>0$ we have
\[
A_{k+h} \le (t^\star)^h A_k, \qquad
B_{k+h} \ge (t^\star)^h B_k.
\]
If $\mu_+=0$, then $D(k)\le0$ for all $k>0$. Otherwise $A_k>0$ for all $k>0$
and the ratio $B_k/A_k$ is nondecreasing in $k$. Consequently, once
$D(k_0)\le0$ (i.e., $B_{k_0}/A_{k_0}\ge1$), we have $B_k/A_k\ge1$ and hence
$D(k)\le0$ for all $k\ge k_0$: $D(k)$ can change sign at most once, from
positive to negative. In the case where $T$ crosses from above and is
side-preserving, the supports of $\mu_+$ and $\mu_-$ are interchanged, so
$B_k/A_k$ is nonincreasing and $D(k)$ can change sign at most once, from
negative to positive. \hfill$\square$

\subsection{Proof of Proposition~\ref{prop:signs} (Marginal Sign-Change Bound)}
\label{app:marginal}

\paragraph{Integral representation and variation diminishing.} For any
joint law of $(P,Q)$, $D(k)=\mathbb E[(1-P)^k]-\mathbb E[(1-Q)^k]$ depends
only on the marginals $F_P,F_Q$. Substituting $Y=-\log(1-P)\in[0,\infty]$
with distribution function $U_P(y)=F_P(1-e^{-y})$, integration by parts
gives $\mathbb E[(1-P)^k]=\int_{[0,\infty)}e^{-ky}\,dU_P(y)=k\int_0^\infty e^{-ky}U_P(y)\,dy$
for $k>0$, and hence
\[
D(k) = k\int_0^\infty e^{-ky}\Delta(y)\,dy, \qquad
\Delta(y) = \big(F_P - F_Q\big)(1-e^{-y}),
\]
which becomes \eqref{eq:marginal} under $u=1-e^{-y}$. Since $y\mapsto 1-e^{-y}$
is strictly increasing, $\Delta(y)$ on $(0,\infty)$ has the exact same
number of sign changes as $F_P-F_Q$ on $(0,1)$. Suppose $\Delta$ is not
almost everywhere zero and changes sign at most $r$ times, at
$0<y_1<\dots<y_r<\infty$. We show by induction on $r$ that
$L(k)=\int_0^\infty e^{-ky}\Delta(y)\,dy$ has at most $r$ zeros on
$(0,\infty)$. For $r=0$, $\Delta$ has constant sign and $L(k)$ has no zeros.
For $r\ge1$, define $M(k)=e^{ky_1}L(k)$; differentiating under the integral
sign gives
\[
M'(k) = -\int_0^\infty e^{-k(y-y_1)}(y-y_1)\Delta(y)\,dy.
\]
Because multiplying $\Delta(y)$ by $(y-y_1)$ eliminates the sign change at
$y_1$, the integrand $(y-y_1)\Delta(y)$ changes sign at most $r-1$ times on
$(0,\infty)$. By the induction hypothesis, $M'(k)$ has at most $r-1$ zeros,
so by Rolle's theorem $M(k)$---and hence $D(k)=kL(k)$---has at most $r$
zeros and at most $r$ sign changes on $(0,\infty)$. \hfill$\square$

\paragraph{Abelian tail asymptotics from the marginals.} As $k\to\infty$,
$D(k)\to F_P(0)-F_Q(0)$. If $F_P(0)=F_Q(0)$ and the lower tails satisfy
$F_P(x)-F_P(0)\sim C_0 x^{a_0}$ and $F_Q(x)-F_Q(0)\sim C_1 x^{a_1}$ as
$x\downarrow0$, Karamata's Abelian theorem for Laplace--Stieltjes transforms
gives $G_0(k)-F_P(0)\sim C_0\Gamma(1+a_0)k^{-a_0}$ and
$G_1(k)-F_Q(0)\sim C_1\Gamma(1+a_1)k^{-a_1}$, so $D(k)<0$ for all large $k$
whenever $a_1<a_0$.

\subsection{Proof of Theorem~\ref{thm:slope} (Lipschitz Slope Bound)}
\label{app:slope}

\paragraph{Finite-depth indistinguishability without regularity bounds.} At
finite generation depths $(n,n')$, the joint count distribution of $(c,d)$
is a linear bijection of the mixed moments $\mathbb E[P^jQ^l]$ for
$0\le j\le n, 0\le l\le n'$, since
$\mathbb E[c^{\underline j}d^{\underline l}]=n^{\underline j}(n')^{\underline l}\mathbb E[P^jQ^l]$.
If $F_P$ is atomless, Carath\'eodory's theorem represents the first $n'$
conditional moments of $H(\cdot\mid p)$ as a convex combination of $n'+1$
point masses $q_r(p)$ with weights $w_r(p)$, and the
elimination-of-randomization theorem of \citet{dvoretzky1951elimination}
partitions $[0,1]$ into measurable sets $A_1,\dots,A_{n'+1}$ such that the
deterministic function $T(p)=q_r(p)$ on $A_r$ matches every mixed moment
$\mathbb E[P^jQ^l]$ and hence reproduces the exact count law of $(c,d)$.
Such a function switches values across a fine partition and has unbounded
slope, motivating Theorem~\ref{thm:slope}.

\paragraph{Proof of Theorem~\ref{thm:slope}.} Conditionally on $(P,Q)$,
$d\sim\mathrm{Bin}(n',Q)$ is independent of $c$, so for $n'\ge2$,
$\mathbb E[Q\mid c]=\mathbb E[d\mid c]/n'$ and
$\mathbb E[Q^2\mid c]=\mathbb E[d(d-1)\mid c]/\{n'(n'-1)\}$. Thus
$A^\star_n=\sum_{g=0}^n\Pr(c=g)\Var(Q\mid c=g)$ and
$\mathbb E[P(1-P)]=\mathbb E[c(n-c)]/\{n(n-1)\}$ are nonparametrically
identified from the joint count law. Because $c$ depends on $(P,Q)$ only
through $P$, $c$ and $Q$ are conditionally independent given $P$, and the
law of total variance yields
\[
A^\star_n = \mathbb E\big[\Var(Q\mid c)\big]
= \mathbb E\big[\Var(Q\mid P)\big] + \mathbb E\big[\Var\big(T(P)\mid c\big)\big],
\qquad T(p)=\mathbb E[Q\mid P=p].
\]
If $Q=T(P)$ for a $\Lambda$-Lipschitz function $T$, then $\Var(Q\mid P)=0$
and
\[
\Var\big(T(P)\mid c\big)
\le \mathbb E\Big[\big\{T(P)-T(\mathbb E[P\mid c])\big\}^2\,\Big|\,c\Big]
\le \Lambda^2\,\Var(P\mid c).
\]
Since the conditional expectation $\mathbb E[P\mid c]$ minimizes mean
squared error among all functions of $c$, we have
$\mathbb E[\Var(P\mid c)]\le\mathbb E[(P-c/n)^2]=\mathbb E[P(1-P)]/n$.
Combining these inequalities gives $A^\star_n\le\Lambda^2\mathbb E[P(1-P)]/n$,
and solving for $\Lambda$ yields \eqref{eq:slope-bound}. \hfill$\square$

\paragraph{Estimation across thinned base depths $n_0\le n$.} For any
$n_0\le n$, hypergeometric thinning $c_0\mid c\sim\mathrm{Hyp}(n,c,n_0)$
produces a marginal count $c_0\mid(P,Q)\sim\mathrm{Bin}(n_0,P)$ independent
of $d$. Within each stratum $c_0=g$ of size $m_g\ge2$, we estimate
$\Var(Q\mid c_0=g)$ unbiasedly by
$\bar y_g-\{m_g(m_g-1)\}^{-1}\sum_{i\ne j}x_i x_j$, where $x_i=d_i/n'$ and
$y_i=d_i(d_i-1)/\{n'(n'-1)\}$, setting singleton strata ($m_g=1$) to zero
so that $\hat A^\star_{n_0}=\sum_{g:m_g\ge2}(m_g/m)\widehat{\Var}(Q\mid c_0=g)$
is conservatively biased downward. One-sided 95\% lower bounds are obtained
from the 5th percentile of 2,000 prompt-level bootstrap resamples.

\section{Estimation, Calibration, and Profile-Likelihood Inference}\label{app:em}
\label{app:model}

\subsection{Grid EM Algorithm and Attenuation Calibration}

\paragraph{Grid EM for the working model.} We discretize the latent base
difficulty $p$ on an $L=200$ grid $\{p_\ell\}_{\ell=1}^L$ equispaced in
$\theta_\ell=\logit p_\ell\in[-9,9]$, with mixing weights $w_\ell$
initialized uniformly. In the E-step, for each prompt $i$ and grid node
$\ell$, we compute the posterior responsibilities of the active component
($Z_i=1$) and the excess-zero component ($Z_i=0$):
\[
A_{i\ell} \propto w_\ell\,\mathrm{Bin}(c_i;n_i,p_\ell)\,(1-\rho_\ell)\,\mathrm{Bin}\!\big(d_i;n_i',T_{\beta,\gamma}(p_\ell)\big),
\qquad
B_{i\ell} \propto w_\ell\,\mathrm{Bin}(c_i;n_i,p_\ell)\,\rho_\ell\,\mathbb 1\{d_i=0\},
\]
normalized so that $\sum_{\ell=1}^L(A_{i\ell}+B_{i\ell})=1$. In the M-step,
the mixing weights update via $w_\ell=m^{-1}\sum_{i=1}^m(A_{i\ell}+B_{i\ell})$;
$(\beta,\gamma)$ maximize the weighted logistic log-likelihood
$\sum_{\ell=1}^L\{s_\ell\eta_\ell-t_\ell\log(1+e^{\eta_\ell})\}$ with
$\eta_\ell=\beta\theta_\ell+\gamma$, $s_\ell=\sum_i A_{i\ell}d_i$, and
$t_\ell=\sum_i A_{i\ell}n_i'$ (a globally concave two-parameter program);
and $\pi_0$ solves the scalar score equation
$\sum_\ell \bar B_\ell/\pi_0=\sum_\ell \bar A_\ell u_\ell/(1-\pi_0 u_\ell)$
with $\bar A_\ell=\sum_i A_{i\ell}$ and $\bar B_\ell=\sum_i B_{i\ell}$.
Iterations terminate when the relative log-likelihood change falls below
$10^{-7}$ (typically under 100 steps), and 95\% confidence intervals use
2,000 prompt-level bootstrap refits.

\paragraph{Eliminating two-stage errors-in-variables attenuation.} At true
parameters $(\beta,\gamma)=(1.4,1.2)$ with $F=\mathrm{Beta}(0.6,2)$,
$m=300$, and shallow counts $(n,n')=(16,8)$ across 100 Monte Carlo
replicates, a two-stage plug-in GLM suffers severe attenuation bias
($\hat\beta=1.320\pm.078, \hat\gamma=1.093\pm.107$), whereas the Grid EM is
unbiased ($\hat\beta=1.410\pm.094, \hat\gamma=1.213\pm.121$). When true
$\pi_0=.15$ ($G=8$), the Grid EM recovers $\pi_0$ within $\pm.01$ across
$n'\in\{8,64,256\}$ and shrinks $\hat\pi_0$ to $\le.03$ when true $\pi_0=0$.
By contrast, a two-part estimator that fits the link on $d_i>0$ under a
zero-truncated binomial is severely upward-biased ($\hat\pi_0\in[.30,.45]$
when true $\pi_0=0$) because plug-in difficulty errors misclassify
continuous small-$p$ zeros as latent-atom mass.

\subsection{Simulation Calibration of the Inference Machinery}

\begin{figure}[t]
\centering
\includegraphics[width=\linewidth]{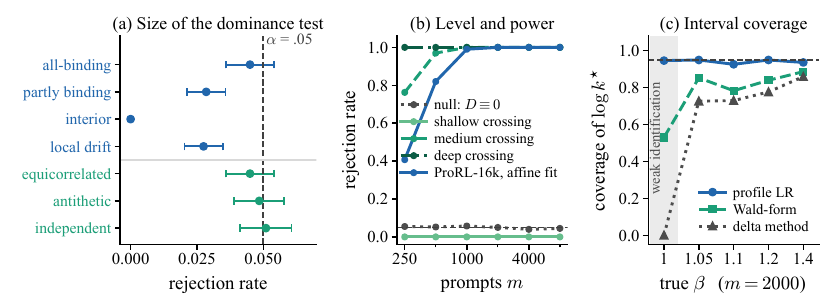}
\caption{\textbf{Finite-sample calibration of the dominance test and
profile-likelihood Fieller sets.} \emph{(a)} Empirical rejection rate of the
dominance test at nominal level $\alpha=.05$ ($2{,}000$ replicates, $m=400$,
$K=6$, $B=999$; error bars show $\pm1.96$ Monte Carlo SE). The top four rows
vary the null mean curve $D(k)\ge0$: the test has nominal size at the
all-binding boundary $D\equiv0$ ($.045$) and is conservative in the interior
of $H_0$. The bottom three rows fix $D\equiv0$ and vary the across-$k$
correlation (equicorrelated, half-antithetic, and independent); all three
achieve nominal size ($.045, .049, .051$) because the prompt-level
multiplier bootstrap adapts to the data's covariance. \emph{(b)} Rejection
rate versus number of prompts $m$ across null and alternative regimes.
\emph{(c)} Empirical coverage of three 95\% confidence-set constructions for
$\log\kstar$ under the parametric working model at $m=2000$: at the
no-sharpening boundary $\beta=1$ ($\kstar=\infty$), the delta method covers
$0\%$ and the Wald Fieller set covers $53\%$, whereas profile-likelihood
inversion maintains $.95$ coverage across all $\beta$.}
\label{fig:calib}
\end{figure}

Figure~\ref{fig:calib} reports the simulation checks verifying
Theorem~\ref{thm:test}. Across $m\in[250,8000]$, the multiplier dominance
test maintains nominal size $.050$--$.058$ at $D\equiv0$, is conservative
inside $H_0$, and adapts automatically to arbitrary across-$k$ covariance
matrices. Compared against a Bonferroni/Holm min-$p$ test over pointwise
studentized statistics at $m=1000, n=n'=256$ (400 replicates), Bonferroni is
conservative under realistic pass@$k$ correlation (size $.015$ versus $.052$)
and sacrifices half the power against the kernel-fitted ProRL-16k
alternative ($.110$ versus $.215$).

\subsection{Profile-Likelihood Fieller Sets under the Parametric Working Model}
\label{app:fieller}

Under the parametric working model $\eta=(a,b,\beta,\gamma,\pi_0)$ with
$F=\mathrm{Beta}(a,b)$, the model-implied crossover
$\kappa_{\mathrm{full}}(\eta)=\inf\{k\ge1:D_{\mathrm{model}}(k;\eta)<0\}$
diverges to $+\infty$ as $(\beta,\pi_0)\to(1,0)$ (see
Equation~\eqref{eq:kstar}), creating a classic denominator singularity
\citep{fieller1954} where delta-method intervals have $0\%$ coverage
(Figure~\ref{fig:calib}c). Inverting the profile likelihood-ratio statistic
\[
\mathcal C_{1-\alpha} = \Big\{\kappa\in[1,\infty] :
2\Big[\ell(\hat\eta) - \sup_{\eta:\kappa_{\mathrm{full}}(\eta)=\kappa}\ell(\eta)\Big]
\le \chi^2_{1,1-\alpha}\Big\}
\]
correctly handles both finite crossings and the unbounded boundary
$\kappa=\infty$ (which is included whenever the no-crossing boundary
$\{\beta\le1,\pi_0=0\}$ lies within the $\chi^2_{1,1-\alpha}$ threshold).
Table~\ref{tab:calib5} confirms that profile-likelihood inversion maintains
$92\%$--$100\%$ coverage across all $(\beta,\pi_0)$ regimes, including
atom-driven crossings at $\beta=1,\pi_0>0$.

\begin{table}[ht]
\centering
\small
\caption{\textbf{Empirical 95\% coverage and unboundedness rate of the
profile-likelihood Fieller set for $\kappa_{\mathrm{full}}$} (300 replicates
per cell, $m=1000, n=n'=64, G=8, \alpha=.05$). $^{\dagger}$At $\beta=1,
\pi_0=0$, the true crossing is $+\infty$ and coverage is the fraction of
sets containing $+\infty$; when $\pi_0>0$ at $\beta=1$, the true crossing is
finite and the interval closes with $0\%$ unbounded sets.}
\label{tab:calib5}
\begin{tabular}{@{}lccc@{}}
\toprule
& \multicolumn{3}{c}{Coverage of true $\kappa_{\mathrm{full}}$ / fraction unbounded} \\
\cmidrule(lr){2-4}
True $\beta$ & $\pi_0=0$ & $\pi_0=.10$ & $\pi_0=.20$ \\
\midrule
$1.00$ & $1.00\,/\,1.00^{\dagger}$ & $.93\,/\,.00$ & $.97\,/\,.00$ \\
$1.05$ & $.92\,/\,.21$             & $.96\,/\,.00$ & $.94\,/\,.00$ \\
$1.20$ & $.95\,/\,.00$             & $.94\,/\,.00$ & $.94\,/\,.00$ \\
$1.40$ & $.93\,/\,.00$             & $.94\,/\,.00$ & $.93\,/\,.00$ \\
\bottomrule
\end{tabular}
\end{table}

\subsection{Flexible Link Specifications (Cubic Spline and Monotone Bernstein)}

\paragraph{Cubic spline link.} For the affine-link likelihood-ratio test and
the parametric-bootstrap nulls of the split-generation test, we replace the
affine link $\beta x+\gamma$ ($x=\logit p$) with an unconstrained 6-parameter
cubic spline using the truncated-power basis
$\{1,x,(x-\xi_1)_+^3,\dots,(x-\xi_4)_+^3\}$, with knots $\xi_1,\dots,\xi_4$
placed at the 20th, 40th, 60th, and 80th percentiles of the empirical logit
difficulties $\logit\{(c_i+0.5)/(n_i+1)\}$, maximized in the M-step via
BFGS. Because Wilks's theorem need not hold in the semiparametric mixture,
the affine-vs-spline LRT is calibrated by parametric bootstrap under the
fitted affine null (observed $2\Delta\ell=404$ on DeepScaleR-16k and $5{,}642$
on ProRL-16k against a 95th null percentile of $\approx10$, giving
$p_{\mathrm{boot}}=.005$ at $B=200$).

\begin{figure}[!ht]
\centering
\includegraphics[width=\linewidth]{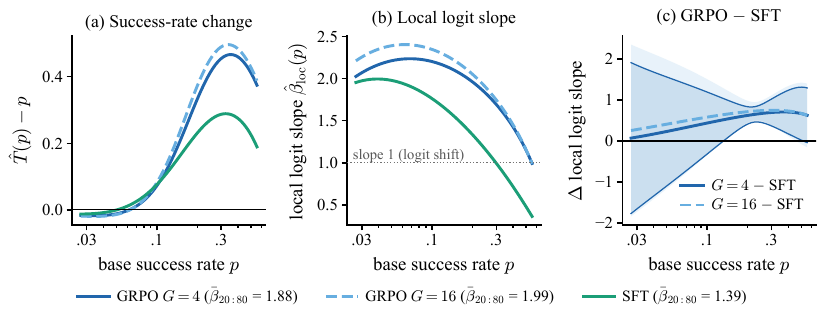}
\caption{\textbf{Controlled response-curve shape comparison: on-policy GRPO
versus verified-solution SFT from Qwen2.5-Math-1.5B} ($m=1000, n=n'=256$,
monotone quartic Bernstein link inside the 200-point Grid EM, shaded 95\%
bootstrap bands). \emph{(a)} Success-rate change $T(p)-p$ vs.\ base difficulty
$p$: GRPO ($G=4, 16$) peaks more sharply on medium prompts and falls below
zero on the hard tail, whereas SFT lifts success rates more uniformly.
\emph{(b)} Local logit slope $d(\logit T)/d(\logit p)$ vs.\ $x=\logit p$:
GRPO has central average slope $\bar\beta_{20:80}=1.88$ ($G=4$) and $1.99$
($G=16$), compared with $\bar\beta_{20:80}=1.39$ for SFT. \emph{(c)} Paired
logit-slope difference between GRPO and SFT: GRPO is significantly steeper
than SFT across the central difficulty band ($\Delta\bar\beta_{20:80}=+.50$
$[.28,.86]$ for $G=4$ and $+.60$ $[.35,1.02]$ for $G=16$, $p<.002$).}
\label{fig:shape}
\end{figure}

\paragraph{Monotone quartic Bernstein link.} For the controlled shape
comparison of Section~\ref{sec:planned} (Figure~\ref{fig:shape}), we fit a
monotone quartic Bernstein polynomial to $\eta(x)=\logit T(\sigma(x))$ on
$[x_{\min},x_{\max}]$ (the 5th and 95th percentiles of empirical logit
difficulty) inside the 200-point Grid EM, enforcing monotonicity via ordered
Bernstein coefficients without restricting local curvature. The central
average logit slope is evaluated over the 20th to 80th percentiles of the
posterior-mean base difficulties $\tilde p_i=\sum_\ell A_{i\ell}p_\ell$:
\[
\bar\beta_{20:80} = \frac{\eta(x_{.80})-\eta(x_{.20})}{x_{.80}-x_{.20}},
\qquad x_r = \operatorname{quantile}_r\{\logit\tilde p_i\},
\]
with simultaneous bands and confidence intervals computed from $B=500$
prompt-paired bootstrap refits. Table~\ref{tab:pairs} reports the single-curve
working-model parameters across 16 evaluations, and Figure~\ref{fig:shape}
displays the fitted flexible response curves for the controlled GRPO and SFT
study.

\begin{table}[!ht]
\centering
\footnotesize
\setlength{\tabcolsep}{3.6pt}
\caption{\textbf{Fitted single-curve working-model parameters across 16
model pairs and settings} ($m=1000$, except $m=1060$ on the 32k new
competition problems and $m=542$ on the code pairs; 95\% prompt-bootstrap
confidence intervals in brackets). Here $\hat\beta_{\mathrm{proj}}$ is the
apparent logit slope under the deterministic working model $Q=T(P)$ before
accounting for within-$p$ dispersion in Section~\ref{sec:kernel}.}
\label{tab:pairs}
\setlength{\tabcolsep}{0pt}
\begin{tabular*}{\textwidth}{@{\extracolsep{\fill}}lllcccc@{}}
\toprule
Model pair & Setting & Regime & $\hat\beta_{\mathrm{proj}}$ [95\% CI] & $\hat\gamma$ [95\% CI] & $\hat\pi_0$ [95\% CI] & $\pstar$ \\
\midrule
\multicolumn{7}{@{}l}{\emph{32k-token benchmark on 1,060 new competition problems ($n=n'=128$)}} \\
DeepScaleR-1.5B & 32k (new)      & Long-CoT RL      & 1.24 [1.18, 1.32] & 0.87 [0.76, 0.99] & .06 [.04, .09] & .026 \\
ProRL-1.5B      & 32k (new)      & Prolonged RL     & 1.81 [1.66, 1.98] & 2.52 [2.21, 2.86] & .09 [.06, .12] & .043 \\
\midrule
\multicolumn{7}{@{}l}{\emph{Public checkpoints on the 1,000-prompt mixture at 3k--16k tokens}} \\
DeepScaleR-1.5B & 16k ($n{=}256$) & Long-CoT RL      & 1.18 [1.14, 1.23] & 0.67 [0.58, 0.76] & .03 [.02, .05] & .024 \\
DeepScaleR-1.5B & 8k ($n{=}64$)   & Long-CoT RL      & 1.07 [1.03, 1.12] & 1.32 [1.22, 1.43] & .07 [.05, .10] & $<.001$ \\
ProRL-1.5B      & 16k ($n{=}256$) & Prolonged RL     & 1.94 [1.80, 2.10] & 2.66 [2.41, 2.94] & .16 [.13, .19] & .056 \\
ORZ-1.5B        & 8k ($n{=}64,32$)& Zero-RL          & 1.39 [1.28, 1.52] & 3.63 [3.36, 3.93] & .01 [.00, .03] & $<.001$ \\
LUFFY-1.5B      & 3k ($n{=}128$)  & Off-policy RLVR  & 0.96 [0.92, 1.01] & 0.49 [0.41, 0.57] & .02 [.01, .04] & -- \\
SimpleRL 1.5B   & 8k ($n{=}128$)  & Zero-RL (1.5B)   & 1.00 [0.96, 1.05] & 3.22 [3.08, 3.37] & .00 [.00, .01] & -- \\
SimpleRL 7B     & 8k ($n{=}128$)  & Zero-RL (7B)     & 2.00 [1.85, 2.16] & 2.64 [2.37, 2.94] & .04 [.02, .06] & .067 \\
SimpleRL 14B    & 8k ($n{=}128$)  & Zero-RL (14B)    & 2.19 [1.98, 2.41] & 2.98 [2.62, 3.38] & .03 [.01, .05] & .076 \\
SimpleRL 32B    & 8k ($n{=}128$)  & Zero-RL (32B)    & 2.30 [2.09, 2.56] & 3.55 [3.14, 4.02] & .04 [.02, .06] & .061 \\
SimpleRL Math-7B& 3k ($n{=}128$)  & Zero-RL (Math-7B)& 2.30 [2.11, 2.53] & 3.41 [3.07, 3.79] & .01 [.00, .03] & .068 \\
\midrule
\multicolumn{7}{@{}l}{\emph{Controlled training from Qwen2.5-Math-1.5B ($m=1000, n=128, n'=64$)}} \\
GRPO $G{=}4$    & 3k             & On-policy RLVR   & 1.67 [1.54, 1.81] & 2.28 [2.05, 2.53] & .01 [.00, .02] & .032 \\
GRPO $G{=}16$   & 3k             & On-policy RLVR   & 1.76 [1.62, 1.91] & 2.41 [2.17, 2.67] & .00 [.00, .02] & .040 \\
Matched SFT     & 3k             & Verified SFT     & 0.93 [0.87, 1.00] & 0.86 [0.74, 0.98] & .00 [.00, .01] & -- \\
\midrule
\multicolumn{7}{@{}l}{\emph{Code generation on HumanEval+ and MBPP+ ($m=542, n=n'=64$, 8k tokens)}} \\
ProRL-1.5B (code) & 8k           & Prolonged RL     & 1.56 [1.36, 1.80] & 1.44 [1.11, 1.81] & .09 [.05, .14] & .071 \\
\bottomrule
\end{tabular*}
\end{table}

\section{Experimental Protocols, Truncation Sensitivity, and Scale Analyses}\label{app:exp}

\subsection{Sampling, Verification, Compute, and Asset Licenses}

\paragraph{Decoding and verification.} All generations use vLLM
\citep{kwon2023vllm} at temperature $0.6$ and top-$p$ $0.95$ with each
checkpoint's official chat template and a standardized boxed-answer
instruction. Mathematical outputs are verified by symbolic equivalence using
Math-Verify \citep{mathverify2025} with a 5-second timeout per generation
(timeouts, parse failures, and wrong answers all score as failures under
identical verifier settings on both sides of every pair). Apart from 10 of
the 1,500 prompts in the 1,000-prompt mixture and 500-prompt 16k replication
sets, all problems have free-form numerical or algebraic answers (none of
the 1,060 new 32k competition problems or the 3,000 deep-$k$ problems has
multiple-choice options), so random-guessing floors \citep{beyond2025passk}
are negligible. Code evaluations on HumanEval+ (164 problems) and MBPP+
(378 problems; \citealp{liu2023evalplus}) are verified by test-suite
execution.

\paragraph{Controlled training hyperparameters.} All controlled runs start
from Qwen2.5-Math-1.5B \citep{yang2024qwen25math} on 8,000 problems sampled
from the DeepScaleR training corpus (excluding the 500 DeepScaleR-corpus
evaluation prompts by corpus index; text deduplication identifies 28
residual overlaps in the 1,000-prompt evaluation set, and dropping those 28
prompts changes $\hat\beta_{\mathrm{proj}}$ by at most $.02$, whereas the
3,000-problem deep-$k$ set has zero overlap). GRPO \citep{shao2024deepseekmath}
is trained in verl~0.8 \citep{sheng2025hybridflow} for 100 updates with a
binary rule-based reward on the final boxed answer: 128 prompts per update,
$G\in\{4,8,16\}$ rollouts per prompt at temperature $1.0$, maximum prompt
and response lengths 768 and 3,072 tokens, PPO mini-batch 64 prompts (one
epoch, clip ratio $0.2$, token-mean loss aggregation, group-standardized
advantages, no entropy bonus), low-variance KL penalty coefficient $10^{-3}$
($10^{-4}$ for the weak-KL arm), and AdamW ($\text{lr}=10^{-6}$, weight
decay $0.01$, gradient clip $1.0$). SFT trains on the 1,405 problems from
the same 8,000-problem pool that include reference solutions for 3 epochs
(132 updates) with completion-only cross-entropy loss, batch size 32, AdamW
($\text{lr}=10^{-5}$, cosine schedule with 3\% warm-up, 3,072-token limit,
bf16).

\paragraph{Compute and assets.} Sampling used $\approx350$ NVIDIA A30
GPU-hours for the initial 1.5B pairs, $\approx70$ H100 GPU-hours for the
SimpleRL-Zoo 7B--32B and non-Qwen pairs, $\approx36$ H100 GPU-hours for
Stage~1 of the 3,000-problem controlled study, and Intel Max~1100 GPUs for
the 32k long-chain evaluations and the Stage~2 $k=4096$ generations.
Controlled training used $\approx630$ A30 GPU-hours ($\approx1{,}040$
including development runs). Table~\ref{tab:assets} lists all models,
datasets, and software packages with their licenses.

\begin{table}[ht]
\caption{\textbf{Models, datasets, and software used in this study}, with
licenses as stated by their providers. $^{\dagger}$The Qwen2.5-14B
SimpleRL-Zoo model card states no license; the SimpleRL-Zoo repository is
MIT-licensed. $^{\S}$The Skywork-OR1-7B model card states no license; its
repository is Apache 2.0.}
\label{tab:assets}
\centering
\small
\begin{tabular}{@{}>{\raggedright\arraybackslash}p{.40\linewidth}>{\raggedright\arraybackslash}p{.35\linewidth}>{\raggedright\arraybackslash}p{.18\linewidth}@{}}
\toprule
Asset & Source & License \\
\midrule
\multicolumn{3}{@{}l}{\emph{Models}} \\
DeepSeek-R1-Distill-Qwen-1.5B, 7B & \citet{guo2025deepseekr1} & MIT \\
DeepScaleR-1.5B-Preview & \citet{luo2025deepscaler} & MIT \\
ProRL (Nemotron-Research-Reasoning-Qwen-1.5B) & \citet{liu2025prorl} & CC BY-NC 4.0 \\
AceReason-Nemotron-7B & \citet{chen2025acereason} & NVIDIA Open Model \\
Skywork-OR1-7B & \citet{he2025skywork} & Apache 2.0$^{\S}$ \\
LUFFY-Qwen-Math-1.5B-Zero & \citet{yan2025luffy} & MIT \\
Open-Reasoner-Zero-1.5B & \citet{hu2025orz} & MIT \\
SimpleRL-Zoo checkpoints & \citet{zeng2025simplerl} & Apache 2.0$^{\dagger}$ \\
Qwen2.5-1.5B, 7B, 14B, 32B & \citet{yang2024qwen25} & Apache 2.0 \\
Qwen2.5-Math-1.5B, Qwen2.5-Math-7B & \citet{yang2024qwen25math} & Apache 2.0 \\
Llama-3.1-8B & \citet{grattafiori2024llama3} & Llama 3.1 Community \\
DeepSeek-Math-7B-Base & \citet{shao2024deepseekmath} & DeepSeek Model \\
Mistral-7B-v0.1 & \citet{jiang2023mistral} & Apache 2.0 \\
\midrule
\multicolumn{3}{@{}l}{\emph{Datasets}} \\
MATH / MATH-500 & \citet{hendrycks2021math,lightman2024verify} & MIT \\
DeepScaleR-Preview-Dataset & \citet{luo2025deepscaler} & MIT \\
OlympiadBench & \citet{he2024olympiadbench} & MIT \\
Gaokao-2023-EN & \citet{liao2024mario} & MIT \\
AMC 2023, AIME 2024 & MAA, via \citet{zeng2025simplerl} & none stated \\
HumanEval+, MBPP+ & \citet{liu2023evalplus,chen2021humaneval,austin2021mbpp} & Apache 2.0 \\
\midrule
\multicolumn{3}{@{}l}{\emph{Software}} \\
vLLM & \citet{kwon2023vllm} & Apache 2.0 \\
verl & \citet{sheng2025hybridflow} & Apache 2.0 \\
Math-Verify & \citet{mathverify2025} & Apache 2.0 \\
\bottomrule
\end{tabular}
\end{table}

\FloatBarrier
\subsection{Multiplicity Structure and 32k Source-Specific Analysis}
\label{app:multiplicity}

\paragraph{Multiplicity adjustment across public pairs.} The initial 1.5B
public-pair discovery suite comprises five dominance tests (DeepScaleR-16k,
ProRL-16k, LUFFY-3k, ORZ-8k, and SimpleRL-Zoo 1.5B at 8k) evaluated at
family-wise Holm level $\alpha=.05$ (smallest-$p$ threshold $.01$). The four
SimpleRL-Zoo checkpoints of \citet{yue2025limit} (Qwen2.5-7B, 14B, 32B, and
Llama-3.1-8B) form a second family of four (smallest-$p$ Holm threshold
$.0125$). Consequently, neither DeepScaleR-16k ($p=.029$) nor SimpleRL-Zoo
7B ($p=.017$) is family-wise significant in the 3k--16k evaluations. By
contrast, on the 1,060 new competition problems at 32k, both DeepScaleR-1.5B
($p=.0027$) and ProRL-1.5B ($p=.0022$) clear the Holm threshold across the
four-model long-chain family.

\begin{figure}[!ht]
\centering
\includegraphics[width=\linewidth]{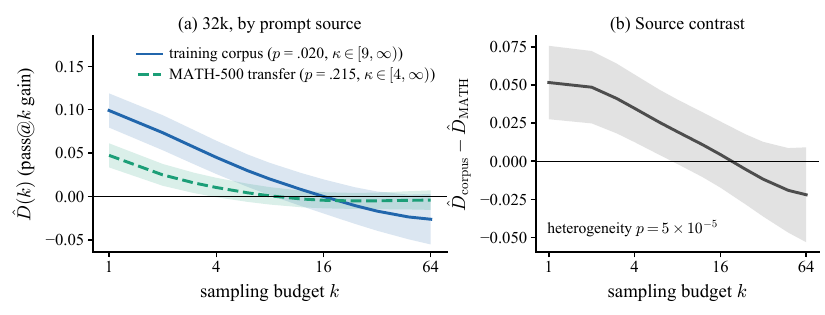}
\caption{\textbf{Source-specific analysis of the 1,000-prompt mixture at
32k tokens (DeepScaleR-1.5B).} \emph{(a)} Separate paired difference curves
$\hat D(k)$ on the 500 MATH-500 prompts and the 500 DeepScaleR-corpus
prompts, with unadjusted dominance $p$-values and first-loss intervals
$\kappa$. \emph{(b)} Between-source contrast
$\hat D_{\mathrm{corpus}}(k)-\hat D_{\mathrm{MATH}}(k)$. Shading uses a
single 95\% simultaneous critical value ($2.806, B=20{,}000$) jointly over
both curves, their contrast, and all $k\le64$. Both point estimates cross
zero, but neither source alone achieves a negative upper confidence bound
under the joint band ($\kappa_{\mathrm{corpus}}\in[9,\infty)$ and
$\kappa_{\mathrm{MATH}}\in[4,\infty)$); only the pooled 1,000-prompt mixture
does.}
\label{fig:dscale32k_sources}
\end{figure}

\paragraph{Source heterogeneity in the 1,000-prompt mixture.}
Figure~\ref{fig:dscale32k_sources} shows the source-stratified analysis of
DeepScaleR at 32k on the 1,000-prompt benchmark pool (500 MATH-500 and 500
DeepScaleR-corpus prompts). A global source-equality test strongly rejects
homogeneity ($p<10^{-4}$), driven by larger small-$k$ gains on the
in-training DeepScaleR-corpus slice, and neither 500-prompt slice alone has
sufficient power to certify the late-$k$ loss under a joint band. This
motivates our 1,060-problem deduplicated competition evaluation below.

\FloatBarrier
\subsection{32k Evaluation on 1,060 New Competition Problems and Truncation-Repair Sensitivity}
\label{app:prereg}
\label{app:longchain}

Table~\ref{tab:fresh32k} reports the complete curve and 95\% simultaneous
sup-$t$ confidence band ($B=4000$) for DeepScaleR-1.5B versus
R1-Distill-Qwen-1.5B on the 1,060 new competition problems at 32k tokens
($n=n'=128$). The band is strictly positive ($L(k)>0$) for all $1\le k\le10$
and strictly negative ($U(k)<0$) for all $61\le k\le128$, localizing the
first-loss budget to $\kappa\in[11,61]$ on the dense integer grid
$k=1,\dots,128$ (and $\kappa\in[9,64]$ on the 10-point sparse grid, with the
sup-$t$ critical value moving only from $2.546$ to $2.557$). Repeating the
multiplier bootstrap across 20 independent seeds supports the crossing in
20/20 runs. Stratified descriptively by benchmark source, OlympiadBench
alone ($m=633$) meets the crossing criterion with $\kappa\in[10,69]$, while
Gaokao-2023-EN ($m=358$) crosses in point estimate at $k=11$ with
$\kappa\in[7,\infty)$ ($p=.12$).

\begin{table}[!ht]
\centering
\small
\caption{\textbf{Paired difference curve and 95\% simultaneous confidence
band for DeepScaleR-1.5B at 32k tokens on the 1,060 new competition
problems} ($n=n'=128, B=4000$, dense grid $k=1,\dots,128$; selected values
of $k$ shown). ``$+$'' marks $L(k)>0$ and ``$-$'' marks $U(k)<0$.}
\label{tab:fresh32k}
\begin{tabular}{@{}rrrrc@{}}
\toprule
Budget $k$ & $\hat D(k)$ & $L(k)$ & $U(k)$ & Supported sign \\
\midrule
1   & $+.0760$ & $+.0655$ & $+.0864$ & $+$ \\
2   & $+.0524$ & $+.0422$ & $+.0625$ & $+$ \\
4   & $+.0308$ & $+.0214$ & $+.0402$ & $+$ \\
8   & $+.0145$ & $+.0052$ & $+.0237$ & $+$ \\
10  & $+.0104$ & $+.0010$ & $+.0198$ & $+$ \\
12  & $+.0077$ & $-.0018$ & $+.0172$ &     \\
16  & $+.0038$ & $-.0060$ & $+.0136$ &     \\
24  & $-.0011$ & $-.0115$ & $+.0093$ &     \\
32  & $-.0045$ & $-.0153$ & $+.0064$ &     \\
48  & $-.0094$ & $-.0211$ & $+.0023$ &     \\
61  & $-.0123$ & $-.0246$ & $-.0001$ & $-$ \\
64  & $-.0130$ & $-.0255$ & $-.0005$ & $-$ \\
96  & $-.0176$ & $-.0317$ & $-.0034$ & $-$ \\
128 & $-.0198$ & $-.0359$ & $-.0037$ & $-$ \\
\bottomrule
\end{tabular}
\end{table}

\FloatBarrier
\paragraph{Sensitivity to hypothetical truncation repair.} On the 1,060 new
problems at 32k, the base model truncates $7.8\%$ of generations and
DeepScaleR truncates $0.7\%$. To bound how much asymmetric base truncation
could affect the comparison, suppose each \emph{failed, truncated} base
generation (let $u_i$ denote their count on prompt $i$) were independently
repaired into a correct solution with probability $f\in[0,1]$, while leaving
RL outcomes unchanged. Under this stochastic repair $J_i\sim\mathrm{Bin}(u_i,f)$,
prompt $i$'s expected base pass@$k$ estimator is
\begin{align*}
\widetilde A_{i,k}(f) &= \sum_{j=0}^{u_i}\binom{u_i}{j}f^j(1-f)^{u_i-j}
\left(1 - \frac{\binom{n_i-c_i-j}{k}}{\binom{n_i}{k}}\right), \\
\hat D_f(k) &= \frac{1}{m}\sum_{i=1}^m\Big\{\widehat{\passk}^{\mathrm{RL}}_i - \widetilde A_{i,k}(f)\Big\}.
\end{align*}
Calibrating a single prompt-multiplier sup-$t$ critical value ($2.863,
B=4000$) jointly over all $k=1,\dots,128$ and all $f\in\{0,.01,\dots,1.00\}$
(and propagating exact interval bounds on two prompts with partially logged
truncation-correctness intersections), Table~\ref{tab:fresh32k-trunc} shows
that because repairing base failures only makes $\hat D_f(k)$ more negative
at large $k$, the crossover remains simultaneously certified for every repair
rate $0\le f\le.83$ (and up to $f\le.92$ when failed RL truncations are
repaired at the same rate $f$).

\begin{table}[!ht]
\centering
\small
\caption{\textbf{Truncation sensitivity and failed-truncation repair on the
1,060 new competition problems at 32k (DeepScaleR-1.5B).} Rows 1--3 report
separate simultaneous bands on the full set and low-truncation subsets.
Rows 4--6 report the joint simultaneous band calibrated across all
$k\le128$ and all base-truncation repair probabilities $f\in[0,1]$.}
\label{tab:fresh32k-trunc}
\begin{tabular}{@{}lrrrrrc@{}}
\toprule
Population / repair scenario & $m$ & $\hat D(1)$ & $L(1)$ & $\hat D(128)$ & $U(128)$ & $\kappa\in$ \\
\midrule
Full population (as reported)       & 1060 & $+.0760$ & $+.0655$ & $-.0198$ & $-.0037$ & $[11,61]$ \\
$\le5\%$ truncation on both sides   &  647 & $+.0404$ & $+.0320$ & $-.0093$ & $+.0003$ & $[3,25]$  \\
Zero truncation on both sides       &  342 & $+.0183$ & $+.0103$ & $-.0029$ & $+.0046$ & $[2,\infty)$ \\
\addlinespace[2pt]
$f=0$ (joint band across $f,k$)     & 1060 & $+.0760$ & $+.0642$ & $-.0198$ & $-.0018$ & $[11,73]$ \\
$f=.25$ failed base truncations     & 1060 & $+.0571$ & $+.0456$ & $-.1091$ & $-.0824$ & $[3,4]$   \\
$f=1.00$ failed base truncations    & 1060 & $+.0005$ & $-.0136$ & $-.1189$ & $-.0904$ & $[1,2]$   \\
\bottomrule
\end{tabular}
\end{table}

\FloatBarrier
\subsection{Checkpoint Scale, Protocol Sensitivity, and Answer-Space Concentration}

\paragraph{Scale dependence in the SimpleRL-Zoo family.} Evaluating the
SimpleRL-Zoo checkpoints across four scales of the Qwen2.5 base family
(1.5B, 7B, 14B, 32B) and on Qwen2.5-Math-7B under identical prompts and
decoding shows that the pure location shift at 1.5B
($\hat\beta_{\mathrm{proj}}=1.00$ $[0.96,1.05]$) jumps to $2.00$ $[1.85,2.16]$,
$2.19$ $[1.98,2.41]$, and $2.30$ $[2.09,2.56]$ at 7B, 14B, and 32B
(Table~\ref{tab:pairs}), and $2.30$ $[2.11,2.53]$ on Qwen2.5-Math-7B. On
 the 7B--32B general-purpose bases, truncation at 8k is at most $2\%$, and
restricting to prompts with $<10\%$ truncation on both sides ($m\ge937$)
changes $\hat\beta_{\mathrm{proj}}$ by at most $.05$. On Qwen2.5-1.5B,
however, the base model fails to terminate on $78.4\%$ of generations at 8k
and still truncates $76.9\%$ when re-sampled at 32k (with pass@1 unchanged
at $.043$), showing that the 1.5B base is censored by non-termination behavior
rather than the token cap.

\paragraph{Sampling depth and decoding temperature in the controlled runs.}
Extending the controlled evaluation on the 1,000 benchmark prompts from
$(n,n')=(128,64)$ to $n=n'=256$ raises the level of every affine slope
estimate ($\hat\beta_{\mathrm{proj}}=2.08, 2.14, 2.18, 2.12$ for GRPO
$G=4,8,16$, weak KL, and $1.37$ for SFT, compared with $1.67, 1.68, 1.76,
1.71$, and $0.93$ at $n'=64$) while leaving the GRPO-minus-SFT slope
contrasts invariant ($.74\to.71$ at $G=4$ and $.82\to.82$ at $G=16$). Raising
the evaluation decoding temperature of GRPO $G=4$ from $0.6$ to $1.0$ shifts
$(\hat\beta_{\mathrm{proj}},\hat\gamma)$ from $(1.67,2.28)$ to $(2.40,3.65)$,
demonstrating that fitted response curves describe the joint
model-plus-decoder law.

\paragraph{Answer-space concentration and modal wrong-answer amplification across GRPO and SFT.}
Evaluating the extracted \texttt{\textbackslash boxed\{...\}} answer strings
across all $n=256$ generations per prompt in the controlled study shows how
within-group relative updates produce locked prompts while rejected-sampled
SFT avoids them. On the Qwen2.5-Math-1.5B base model, each prompt produces
an average of $18.2$ distinct extracted answers with a modal answer share of
$51\%$. Across the four GRPO arms, the average number of distinct answers
collapses to $4.0$--$4.4$ per prompt and the modal answer share rises to
$83\%$--$85\%$; under SFT, by contrast, diversity remains twice as wide
($8.6$ distinct answers per prompt, modal share $69\%$). Inspecting the $7$
prompts where the base model and SFT succeed ($c_i\in[1,18]$, SFT
$d_i\in[2,85]$) while all four GRPO arms drop to $0/256$ confirms that GRPO
concentrates probability mass onto the base model's dominant plausible
distractor whenever no positive rollout enters the training group $G$: for
example, on a probability-bound problem asking for a Chebyshev inequality
guarantee, the base model places $1/256$ on the exact Chebyshev bound and a
plurality of generations on a normal-approximation answer; SFT (trained only
on verified positive trajectories) lifts the Chebyshev derivation to
$39/256$, whereas all four GRPO arms collapse to $0/256$ on the Chebyshev
bound and concentrate $>80\%$ of their generations on the normal-approximation
distractor.

\FloatBarrier
\subsection{Summary Re-analysis of Published Crossover Claims}\label{app:reanalysis}

Table~\ref{tab:reanalysis} summarizes how our paired statistical inference
compares against external claims in the literature across the nine public
3k--16k checkpoints.

\begin{table}[!ht]
\centering
\footnotesize
\setlength{\tabcolsep}{3pt}
\caption{\textbf{Comparison of published crossover positions with paired
statistical inference} on the 1,000-prompt evaluation at 3k--16k token
limits (32k new-problem outcomes noted for DeepScaleR and ProRL).}
\label{tab:reanalysis}
\begin{tabular}{@{}>{\raggedright\arraybackslash}p{2.1cm}>{\raggedright\arraybackslash}p{3.1cm}cc>{\raggedright\arraybackslash}p{3.5cm}@{}}
\toprule
Checkpoint & Literature position & \shortstack{Dominance\\$p$} & \shortstack{First-loss\\$\kappa\in$} & Statistical verdict \\
\midrule
DeepScaleR-1.5B & No inversion reported at 8k \citep{luo2025deepscaler} & $.99$ (8k) / $.029$ (16k) & $[14,\infty)$ (16k) & No crossing at 8k; unadjusted dominance rejection at 16k; \textbf{crossing certified at 32k} ($\kappa\in[11,61]$) \\
ProRL-1.5B & Prolonged RL expands boundary \citep{liu2025prorl} & $.20$ (16k) & $[24,\infty)$ (16k) & Not significant at 16k (power $.21$); \textbf{crossing certified at 32k on new problems} ($\kappa\in[12,63]$) \\
SimpleRL-Zoo 7B & Sharpening crossover \citep{yue2025limit} & $.017$ (8k) & $[13,114]$ & Per-pair crossing supported ($U(k)<0$ from $k=114$); not family-wise significant under Holm \\
SimpleRL-Zoo 14B & Sharpening crossover \citep{yue2025limit} & $.15$ (8k) & $[14,\infty)$ & Visual crossing at $k=34$; not statistically established \\
SimpleRL-Zoo 32B & Sharpening crossover \citep{yue2025limit} & $.050$ (8k) & $[16,\infty)$ & Visual crossing at $k=34$; not statistically established \\
SimpleRL-Zoo Llama-8B & Sharpening crossover \citep{yue2025limit} & $.41$ (8k) & $[30,\infty)$ & Visual crossing at $k=74$; not statistically established \\
SimpleRL-Zoo 1.5B & Same zero-RL recipe & $>.99$ (8k) & $[129,\infty)$ & RL ahead throughout $k\le128$; heavy base non-termination \\
ORZ-1.5B & Zero-RL family \citep{hu2025orz,yue2025limit} & $>.99$ (8k) & $[129,\infty)$ & No crossing in $k\le128$; base truncation-confounded \\
LUFFY-1.5B & Off-policy RLVR gains \citep{yan2025luffy} & $>.99$ (3k) & $[109,\infty)$ & No crossing in $k\le128$; off-policy mixed objective \\
\bottomrule
\end{tabular}
\end{table}

\FloatBarrier
\section{Extended Kernel, Heterogeneity, and \texorpdfstring{Deep-$k$}{Deep-k} Analyses}\label{app:het}
\label{app:het-split}\label{app:het-refit}\label{app:het-consequences}

\subsection{Full Split-Generation Residual Covariance, Beta-Binomial Refits, and Mean-Collapse Counterfactuals}

Table~\ref{tab:split} reports the split-generation residual covariance test
of Equations~\eqref{eq:split-stat}--\eqref{eq:split-decomp} across all 20
evaluated datasets, together with the in-sample log-likelihood gain
$\Delta\ell$ of the $80\times80$ 2D NPMLE kernel over the zero-inflated cubic
spline and the held-out generation log-likelihood gains $\Delta\log p(d_B\mid c,d_A)$.
Table~\ref{tab:phiprofile} reports the complete Beta-Binomial random-effects
refits across 10 pairs, and Table~\ref{tab:mean-main} reports the complete
mean-collapse counterfactuals across all 9 crossing pairs.

\begin{table}[!ht]
\centering
\footnotesize
\setlength{\tabcolsep}{3.0pt}
\caption{\textbf{Split-generation residual covariance and 2D NPMLE kernel
comparison across 20 analyses.} $\widehat{\Cov}_{\mathrm{split}}$ and
$\widehat{\mathrm{sd}}=\sqrt{(\widehat{\Cov}_{\mathrm{split}})_+}$ are
computed under the fitted zero-inflated cubic spline ($z=\text{est./SE}$).
``Placebo $z$'' splits base counts $(c_A,c_B)$ versus $(d_A,d_B)$ within the
same model. ``Cal.\ $p$'' is the one-sided parametric-bootstrap $p$-value
($B=400$) under the fitted zero-inflated spline null ($\le.0025$ means zero
null exceedances). ``In-sample $\Delta\ell$'' is the log-likelihood gain of
the $80\times80$ 2D NPMLE kernel over the spline; ``Held-out $\Delta$'' is
the predictive log-likelihood gain $\Delta\log p(d_B\mid c,d_A)$ on the
untouched half of post-RL generations. $^{\mathrm{h}}$170-prompt hard-tail
deep-sampled slices ($n'=2048$). $^{\mathrm{c}}$Code-generation pairs
($m=542, n=n'=64$, 8k tokens).}
\label{tab:split}
\setlength{\tabcolsep}{1.8pt}
\begin{tabular*}{\textwidth}{@{\extracolsep{\fill}}lccccccccc@{}}
\toprule
Model pair / setting & $n,n'$ & $\widehat{\Cov}_{\mathrm{split}}$ & $\widehat{\mathrm{sd}}$ &
$z$ & \shortstack{Plac.\\$z$} & \shortstack{Cal.\\$p$} & \shortstack{Split\\$\rho$} &
\shortstack{In-samp.\\$\Delta\ell$} &
\shortstack{Held-out $\Delta$\\kern.\ / BB} \\
\midrule
DeepScaleR-1.5B (32k, new) & 128, 128 & $+.0102$ & $.101$ & $10.0$ & $0.7$ & $\le.0025$ & $.82$ & $+652$  & $+395$ / $+419$ \\
ProRL-1.5B (32k, new)      & 128, 128 & $+.0249$ & $.158$ & $14.3$ & $1.6$ & $\le.0025$ & $.89$ & $+1026$ & $+710$ / $+772$ \\
AceReason-7B (32k, new)    & 128, 128 & $+.0106$ & $.103$ & $10.6$ & $-0.2$& $\le.0025$ & $.83$ & $+562$  & $+282$ / $+326$ \\
Skywork-OR1-7B (32k, new)  & 128, 128 & $+.0094$ & $.097$ &  $9.4$ & $-0.2$& $\le.0025$ & $.82$ & $+482$  & $+163$ / $+216$ \\
\addlinespace[2pt]
DeepScaleR-1.5B (16k)          & 256, 256 & $+.0124$ & $.111$ & $10.9$ & $-0.8$ & $\le.0025$ & $.87$ & $+1467$ & $+967$ / $+998$ \\
ProRL-1.5B (16k)               & 256, 256 & $+.0185$ & $.136$ & $12.6$ & $-0.5$ & $\le.0025$ & $.88$ & $+1043$ & $+515$ / $+562$ \\
DeepScaleR-1.5B (8k)           &  64, 64  & $+.0156$ & $.125$ & $11.5$ & $1.6$  & $\le.0025$ & $.79$ & $+582$  & $+275$ / $+299$ \\
SimpleRL-Zoo 1.5B (8k)         & 128, 128 & $+.0024$ & $.049$ &  $6.1$ & $0.7$  & $\le.0025$ & $.58$ & $+157$  & $+7$ / $+29$ \\
SimpleRL-Zoo 7B (8k)           & 128, 128 & $+.0221$ & $.149$ & $13.3$ & $1.6$  & $\le.0025$ & $.85$ & $+866$  & $+348$ / $+381$ \\
SimpleRL-Zoo 14B (8k)          & 128, 128 & $+.0268$ & $.164$ & $14.3$ & $0.6$  & $\le.0025$ & $.87$ & $+961$  & $+463$ / $+490$ \\
SimpleRL-Zoo 32B (8k)          & 128, 128 & $+.0258$ & $.161$ & $13.6$ & $1.6$  & $\le.0025$ & $.88$ & $+940$  & $+369$ / $+436$ \\
SimpleRL-Zoo Math-7B (3k)      & 128, 128 & $+.0309$ & $.176$ & $14.6$ & $1.1$  & $\le.0025$ & $.88$ & $+773$  & $+340$ / $+361$ \\
\addlinespace[2pt]
Controlled GRPO $G{=}4$        & 256, 256 & $+.0630$ & $.251$ & $20.6$ & $-0.2$ & $\le.0025$ & $.96$ & $+1706$ & $+1058$ / $+1054$ \\
Controlled GRPO $G{=}8$        & 256, 256 & $+.0642$ & $.253$ & $20.7$ & $1.0$  & $\le.0025$ & $.96$ & $+1683$ & $+930$ / $+947$ \\
Controlled GRPO $G{=}16$       & 256, 256 & $+.0636$ & $.252$ & $20.7$ & $0.6$  & $\le.0025$ & $.96$ & $+1383$ & $+741$ / $+743$ \\
Controlled GRPO weak KL        & 256, 256 & $+.0620$ & $.249$ & $20.6$ & $0.2$  & $\le.0025$ & $.96$ & $+1408$ & $+723$ / $+756$ \\
Controlled SFT                 & 256, 256 & $+.0322$ & $.179$ & $17.8$ & $-0.5$ & $\le.0025$ & $.92$ & $+1007$ & $+565$ / $+605$ \\
\addlinespace[2pt]
ProRL-1.5B (code, 8k)$^{\mathrm{c}}$      &  64, 64  & $+.0365$ & $.191$ & $10.0$ & $0.6$ & $\le.0025$ & $.91$ & $+482$ & $+272$ / $+285$ \\
DeepScaleR-1.5B (code, 8k)$^{\mathrm{c}}$ &  64, 64  & $+.0067$ & $.082$ &  $4.6$ & $0.9$ & $\le.0025$ & $.76$ & $+125$ & $+21$ / $+42$ \\
GRPO $G{=}4$ hard slice$^{\mathrm{h}}$   & 128, 2048 & $+.0226$ & $.150$ &  $5.4$ & $0.3$ & $\le.0025$ & $.99$ & $+347$ & $+55$ / $+49$ \\
\bottomrule
\end{tabular*}
\end{table}

\begin{table}[!ht]
\centering
\footnotesize
\setlength{\tabcolsep}{3.0pt}
\caption{\textbf{Conditional-mean logit slope $\hat\beta$ under the
Beta-Binomial random-effects model across 10 pairs} ($\phi$ profiled over
$[0.3,200]$), compared against the single-curve projection slope
$\hat\beta_{\mathrm{proj}}$ ($\phi\to\infty$; 95\% prompt-bootstrap CIs in
brackets where computed).}
\label{tab:phiprofile}
\begin{tabular*}{\textwidth}{@{\extracolsep{\fill}}lcccccc@{}}
\toprule
Model pair & \shortstack{Single-curve\\$\hat\beta_{\mathrm{proj}}$ ($\phi{=}\infty$)} & \shortstack{Profiled\\$\hat\phi$} & \shortstack{Beta-Binomial\\$\hat\beta$ ($\phi{=}\hat\phi$)} & \shortstack{Beta-Bin.\\$\hat\gamma$} & \shortstack{Resid.\\$\hat\pi_0$} & $\Delta\ell$ \\
\midrule
DeepScaleR-1.5B (32k, new)& $1.24$ $[1.18,1.32]$ & $8.0$ & $0.96$ $[0.89,1.03]$ & $0.82$ & $.05$ & $+702$ \\
ProRL-1.5B (32k, new)     & $1.81$ $[1.66,1.98]$ & $2.2$ & $0.84$ $[0.77,0.92]$ & $1.31$ & $.02$ & $+1467$ \\
AceReason-7B (32k, new)   & $1.16$               & $7.6$ & $0.91$               & $0.58$ & $.02$ & $+592$ \\
Skywork-OR1-7B (32k, new) & $1.21$               & $8.2$ & $0.95$               & $0.80$ & $.01$ & $+483$ \\
\addlinespace[2pt]
DeepScaleR-1.5B (16k)          & $1.18$ $[1.14,1.23]$ & $6.3$ & $0.92$ $[0.89,0.98]$ & $0.68$ & $.03$ & $+1459$ \\
ProRL-1.5B (16k)               & $1.94$ $[1.80,2.10]$ & $1.9$ & $0.76$ $[0.70,0.83]$ & $1.47$ & $.13$ & $+2748$ \\
SimpleRL-Zoo Math-7B (3k)      & $2.30$ $[2.11,2.53]$ & $1.9$ & $0.85$ $[0.79,0.91]$ & $1.75$ & $.00$ & $+2013$ \\
Controlled GRPO $G{=}4$        & $2.08$ $[1.93,2.24]$ & $1.4$ & $0.85$ $[0.80,0.91]$ & $1.21$ & $.00$ & $+4761$ \\
Controlled GRPO $G{=}16$       & $2.18$ $[2.02,2.35]$ & $1.4$ & $0.93$ $[0.88,1.00]$ & $1.35$ & $.00$ & $+4720$ \\
Controlled SFT                 & $1.37$ $[1.29,1.45]$ & $2.9$ & $0.83$ $[0.79,0.87]$ & $0.67$ & $.00$ & $+6529$ \\
\bottomrule
\end{tabular*}
\end{table}

\begin{table}[!ht]
\centering
\footnotesize
\setlength{\tabcolsep}{2.6pt}
\caption{\textbf{Mean-collapse counterfactual under the fitted 2D NPMLE
kernel across all 9 crossing pairs} ($k\le K$, where $K=128$ at 32k/8k and
$K=256$ at 16k/controlled). Collapsing $H(q\mid p_\ell)$ to a point mass at
its conditional mean $\bar q_\ell=\mathbb E_H[Q\mid P=p_\ell]$ eliminates all
seven 8k/16k/controlled crossings and delays or eliminates the 32k crossings.}
\label{tab:mean-main}
\begin{tabular*}{\textwidth}{@{\extracolsep{\fill}}lccccrr@{}}
\toprule
Model pair & \shortstack{Empirical\\$k^\star$} & \shortstack{Full kernel\\$k^\star$} &
\shortstack{Mean-coll.\\$k^\star$} & \shortstack{Hard-tail\\coll.\ $k^\star$} &
\shortstack{Full kernel\\$D(K)$} & \shortstack{Mean-coll.\\$D(K)$} \\
\midrule
DeepScaleR-1.5B (32k, new) &  22 &  23 & 51   & none & $-.0198$ & $-.0046$ \\
ProRL-1.5B (32k, new)      &  25 &  26 & none & none & $-.0207$ & $+.0061$ \\
DeepScaleR-1.5B (16k)           &  27 &  27 & none & none & $-.0107$ & $+.0029$ \\
ProRL-1.5B (16k)                &  56 &  58 & none & none & $-.0087$ & $+.0180$ \\
SimpleRL-Zoo 7B (8k)            &  28 &  28 & none & none & $-.0225$ & $+.0055$ \\
SimpleRL-Zoo 14B (8k)           &  34 &  34 & none & none & $-.0166$ & $+.0145$ \\
SimpleRL-Zoo 32B (8k)           &  34 &  34 & none & none & $-.0205$ & $+.0147$ \\
Controlled GRPO $G{=}4$         & 117 & 140 & none & none & $-.0018$ & $+.0258$ \\
Controlled GRPO $G{=}16$        & 150 & 168 & none & none & $-.0009$ & $+.0273$ \\
\bottomrule
\end{tabular*}
\end{table}

\subsection{2D NPMLE Boundary Checks and Held-Out Generation Prediction}

\paragraph{Boundary robustness of the 2D NPMLE kernel.} In the $80\times80$
2D NPMLE grid of Section~\ref{sec:kernel}, $q_j\in\{0\}\cup\{\sigma(v_j)\}$
includes an explicit atom at $q=0$ while $p_\ell=\sigma(u_\ell)$ uses
interior nodes down to $\sigma(-14)=8.3\times10^{-7}$. Four checks confirm
that the finite-$k$ kernel crossings $\kstar$ are insensitive to boundary
discretization: (1) the fitted $q=0$ mass is small ($.0096$ on
DeepScaleR-16k, $.0133$ on ProRL-16k, $.0009$ on GRPO $G=4$, and $.0018$ on
GRPO $G=16$, versus $.052$--$.073$ of prompts with $d_i=0$); (2) shifting all
$q=0$ mass to the smallest positive node $\sigma(-14)$ leaves every
$\kstar$ unchanged ($27, 58, 140, 168$) because $(1-\sigma(-14))^k\approx1$
for $k\le4000$; (3) adding a symmetric $p=0$ atom on the base side leaves
$\kstar$ virtually unchanged ($27, 58, 136, 184$); and (4) reassigning
$q=0$ mass to the first node above count resolution $1/n'=.0039$ ($q=.0043$)
moves DeepScaleR-16k's crossing only from $27$ to $30$, while moving
ProRL-16k's from $58$ to $288$ and GRPO $G=4$'s from $140$ to $163$---demonstrating
directly that sub-resolution mass placement governs extrapolations when the
late-$k$ loss is shallow.

\paragraph{Held-out generation likelihood and half-sample curve
reconstructions.} Table~\ref{tab:heldout} reports the posterior-predictive
log-likelihood gains $\Delta\log p(d_B\mid c,d_A)$ on untouched post-RL
generation halves $B$, and Table~\ref{tab:kernel-crossval} reports the
corresponding two-fold generation-split curve comparisons. Both the 2D
NPMLE kernel and the Beta-Binomial model outperform every deterministic
curve by hundreds to thousands of log-likelihood units wherever conditional
dispersion is present.

\begin{table}[!ht]
\centering
\footnotesize
\setlength{\tabcolsep}{3pt}
\caption{\textbf{Held-out generation log-likelihood comparison across 18
analyses.} All models are fit on $(c,d_A)$ (full base counts and post-RL
half $A$) and scored on the untouched post-RL half $B$ via
$\Delta\log p(d_B\mid c,d_A)$ relative to the zero-inflated cubic-spline
model (positive values favor the row model; $\pm$ prompt-level SE).
$\hat\phi$ is the profiled Beta-Binomial precision and BB $\hat\pi_0$ is the
residual zero atom. $^{\mathrm{h}}$170-prompt hard-tail deep-sampled slices
($n'=2048$).}
\label{tab:heldout}
\begin{tabular}{@{}lrrrrrr@{}}
\toprule
Model pair / setting & \shortstack{Det.\\spline} & \shortstack{ZI\\affine} & \shortstack{ZI Beta-Bin.\\$\pm$ SE} & \shortstack{2D NPMLE\\kernel $\pm$ SE} & $\hat\phi$ & \shortstack{BB\\$\hat\pi_0$} \\
\midrule
DeepScaleR-1.5B (32k, new)     & $+25$ & $-4$    & $+419\pm49$   & $+395\pm47$   &  8.0 & .054 \\
ProRL-1.5B (32k, new)          & $+473$& $-401$  & $+772\pm87$   & $+710\pm85$   &  2.2 & .019 \\
DeepScaleR-1.5B (16k)          & $-88$ & $-80$   & $+998\pm100$  & $+967\pm100$  &  6.3 & .025 \\
ProRL-1.5B (16k)               & $-6$  & $-1612$ & $+562\pm56$   & $+515\pm56$   &  1.9 & .127 \\
DeepScaleR-1.5B (8k)           & $+29$ & $-28$   & $+299\pm36$   & $+275\pm36$   &  5.4 & .063 \\
SimpleRL-Zoo 1.5B (8k)         & $+6$  & $0$     & $+29\pm8$     & $+7\pm8$      & 68.0 & .003 \\
SimpleRL-Zoo Math-7B (3k)      & $0$   & $-1238$ & $+361\pm34$   & $+340\pm35$   &  1.9 & .000 \\
Controlled GRPO $G{=}4$        & $0$   & $-3051$ & $+1054\pm69$  & $+1058\pm72$  &  1.4 & .000 \\
Controlled GRPO $G{=}8$        & $0$   & $-3296$ & $+947\pm63$   & $+930\pm65$   &  1.4 & .000 \\
Controlled GRPO $G{=}16$       & $0$   & $-3310$ & $+743\pm52$   & $+741\pm57$   &  1.4 & .000 \\
Controlled GRPO weak KL        & $+2$  & $-3447$ & $+756\pm59$   & $+723\pm63$   &  1.5 & .000 \\
Controlled SFT                 & $+149$& $-5501$ & $+605\pm51$   & $+565\pm54$   &  2.9 & .000 \\
Controlled GRPO $G{=}4$ ($T{=}1$) & $0$& $-527$  & $+52\pm16$    & $+91\pm16$    &  2.3 & .000 \\
ProRL-1.5B (code, 8k)          & $+95$ & $-149$  & $+285\pm46$   & $+272\pm45$   &  1.5 & .078 \\
DeepScaleR-1.5B (code, 8k)     & $+8$  & $+5$    & $+42\pm14$    & $+21\pm15$    & 10.3 & .011 \\
GRPO $G{=}4$ hard slice$^{\mathrm{h}}$  & $0$ & $-83$ & $+49\pm25$ & $+55\pm24$  &  2.8 & .000 \\
GRPO $G{=}16$ hard slice$^{\mathrm{h}}$ & $-1$& $-94$ & $+44\pm19$ & $+48\pm18$  &  2.9 & .000 \\
SFT hard slice$^{\mathrm{h}}$           & $0$ & $-38$ & $+48\pm17$ & $+28\pm15$  &  8.7 & .000 \\
\bottomrule
\end{tabular}
\end{table}

\begin{table}[ht]
\centering
\footnotesize
\setlength{\tabcolsep}{3.5pt}
\caption{\textbf{Two-fold generation-split comparison of model-implied
crossings within the half-sample window} ($k\le127$, except SimpleRL-Zoo
1.5B at $k\le63$). ``--'' indicates no crossing within the window.}
\label{tab:kernel-crossval}
\begin{tabular}{@{}lcccc@{}}
\toprule
Model pair & \shortstack{$A\to B$\\(pred.\ / obs.)} & \shortstack{$B\to A$\\(pred.\ / obs.)} &
\shortstack{Cross-fitted\\(pred.\ / obs.)} & \shortstack{Mean fold\\curve MAE} \\
\midrule
DeepScaleR-1.5B (16k)      & $26 / 32$ & $33 / 25$ & $29 / 28$ & $.0062$ \\
ProRL-1.5B (16k)           & $74 / 56$ & $61 / 68$ & $67 / 62$ & $.0013$ \\
Controlled GRPO $G{=}4$    & -- / --   & -- / $115$& -- / --   & $.0015$ \\
Controlled GRPO $G{=}16$   & -- / --   & -- / --   & -- / --   & $.0025$ \\
Controlled SFT             & -- / --   & -- / --   & -- / --   & $.0015$ \\
SimpleRL-Zoo 1.5B (8k)     & -- / --   & -- / --   & -- / --   & $.0063$ \\
\midrule
\multicolumn{5}{@{}l}{\emph{Spline-link Beta-Binomial under the identical protocol}} \\
DeepScaleR-1.5B (16k)      & $29 / 32$ & $38 / 25$ & $33 / 28$ & $.0062$ \\
ProRL-1.5B (16k)           & $70 / 56$ & $83 / 68$ & $75 / 62$ & $.0043$ \\
\bottomrule
\end{tabular}
\end{table}

\subsection{Non-Qwen Base Replications and Full Lipschitz Slope-Bound Table}
\label{app:het-nonqwen}

Table~\ref{tab:nonqwen} details the non-Qwen replications on Llama-3.1-8B
and DeepSeek-Math-7B (SimpleRL-Zoo recipe, $n=n'=128$ at 8k tokens). (A
third non-Qwen checkpoint, Mistral-7B-v0.1, is excluded because its RL
checkpoint fails to terminate on $1{,}008$ of $1{,}024$ generations at 8k,
with $82\%$ of completions containing no \texttt{\textbackslash boxed\{...\}} expression anywhere.)
Table~\ref{tab:slope} reports the distribution-free Lipschitz slope bounds
$\hat\Lambda_{\min}(n_0)$ of Theorem~\ref{thm:slope} across all 20 datasets
and base depths $n_0\in\{16,32,64,128\}$.

\begin{table}[ht]
\centering
\footnotesize
\setlength{\tabcolsep}{3.8pt}
\caption{\textbf{Replication of conditional dispersion on non-Qwen base
models} (SimpleRL-Zoo recipe, $m=1000, n=n'=128$, 8k-token limit). $z$ is
the studentized split-generation residual covariance; Cal.\ $p$ is the
parametric-bootstrap $p$-value ($B=400$); $\Delta\ell$ is the in-sample
log-likelihood gain of the 2D NPMLE kernel over the zero-inflated spline;
and Held-out $\Delta$ is the out-of-sample log-likelihood gain on half $B$.}
\label{tab:nonqwen}
\setlength{\tabcolsep}{0pt}
\begin{tabular*}{\textwidth}{@{\extracolsep{\fill}}lccccccc@{}}
\toprule
Model pair & \shortstack{pass@1\\base $\to$ RL} & \shortstack{Trunc.\\base / RL} &
Split $z$ & Plac.\ $z$ & Cal.\ $p$ & \shortstack{In-samp.\\$\Delta\ell$} &
\shortstack{Held-out $\Delta$\\kern.\ / BB} \\
\midrule
SimpleRL Llama-3.1-8B     & $.079\to.159$ & $.17 / .16$ & $12.2$ & $0.3$ & $\le.0025$ & $+637$ & $+66\pm18$ / $+89\pm18$ \\
SimpleRL DeepSeek-Math-7B & $.033\to.075$ & $.47 / .39$ &  $7.1$ & $1.4$ & $\le.0025$ & $+177$ & $+32\pm14$ / $+50\pm13$ \\
\bottomrule
\end{tabular*}
\end{table}

\begin{table}[ht]
\centering
\footnotesize
\setlength{\tabcolsep}{2.2pt}
\caption{\textbf{Distribution-free Lipschitz slope bounds across all 20
datasets} (Theorem~\ref{thm:slope}): point estimate $\hat\Lambda_{\min}(n_0)$
at thinned base depth $n_0$, with the one-sided 95\% prompt-bootstrap lower
bound in parentheses, compared against the maximum slope $\max_p|T'(p)|$ of
the pair's fitted affine-logit curve.}
\label{tab:slope}
\setlength{\tabcolsep}{0pt}
\begin{tabular*}{\textwidth}{@{\extracolsep{\fill}}lccccccc@{}}
\toprule
Model pair / dataset & $m$ & $n$ & $n_0{=}16$ & $n_0{=}32$ & $n_0{=}64$ & $n_0{=}128$ & \shortstack{Fitted\\$\max|T'|$} \\
\midrule
DeepScaleR-1.5B (32k, new) & 1060 & 128 & $1.78$ ($\ge 1.62$) & $2.16$ ($\ge 1.93$) & $2.92$ ($\ge 2.50$) & $4.02$ ($\ge 3.09$) & $1.88$ \\
ProRL-1.5B (32k, new)      & 1060 & 128 & $2.47$ ($\ge 2.25$) & $3.30$ ($\ge 2.93$) & $4.43$ ($\ge 3.79$) & $5.92$ ($\ge 4.65$) & $2.78$ \\
DeepScaleR-1.5B (16k)           & 1000 & 256 & $1.91$ ($\ge 1.73$) & $2.41$ ($\ge 2.13$) & $3.42$ ($\ge 2.88$) & $4.39$ ($\ge 3.35$) & $2.03$ \\
ProRL-1.5B (16k)                & 1000 & 256 & $2.55$ ($\ge 2.32$) & $3.27$ ($\ge 2.90$) & $4.75$ ($\ge 4.07$) & $6.36$ ($\ge 5.03$) & $3.62$ \\
DeepScaleR-1.5B (8k)            & 1000 &  64 & $1.90$ ($\ge 1.72$) & $2.56$ ($\ge 2.28$) & $3.36$ ($\ge 2.89$) & --                  & $2.75$ \\
SimpleRL-Zoo 1.5B (8k)          & 1000 & 128 & $1.76$ ($\ge 1.64$) & $2.08$ ($\ge 1.92$) & $2.59$ ($\ge 2.37$) & $3.00$ ($\ge 2.68$) & $3.14$ \\
SimpleRL-Zoo 7B (8k)            & 1000 & 128 & $2.35$ ($\ge 2.18$) & $3.18$ ($\ge 2.91$) & $4.33$ ($\ge 3.86$) & $5.97$ ($\ge 5.06$) & $2.80$ \\
SimpleRL-Zoo 14B (8k)           & 1000 & 128 & $2.57$ ($\ge 2.38$) & $3.33$ ($\ge 3.06$) & $4.75$ ($\ge 4.25$) & $6.55$ ($\ge 5.55$) & $3.33$ \\
SimpleRL-Zoo 32B (8k)           & 1000 & 128 & $2.30$ ($\ge 2.13$) & $3.08$ ($\ge 2.82$) & $4.19$ ($\ge 3.74$) & $5.70$ ($\ge 4.80$) & $3.86$ \\
SimpleRL-Zoo Math-7B (3k)       & 1000 & 128 & $2.82$ ($\ge 2.67$) & $3.66$ ($\ge 3.44$) & $4.97$ ($\ge 4.64$) & $6.95$ ($\ge 6.35$) & $5.69$ \\
Controlled GRPO $G{=}4$         & 1000 & 256 & $2.89$ ($\ge 2.73$) & $3.98$ ($\ge 3.73$) & $5.58$ ($\ge 5.16$) & $7.71$ ($\ge 6.89$) & $3.32$ \\
Controlled GRPO $G{=}8$         & 1000 & 256 & $2.91$ ($\ge 2.75$) & $4.00$ ($\ge 3.75$) & $5.60$ ($\ge 5.19$) & $7.76$ ($\ge 6.93$) & $3.46$ \\
Controlled GRPO $G{=}16$        & 1000 & 256 & $2.90$ ($\ge 2.74$) & $3.99$ ($\ge 3.74$) & $5.59$ ($\ge 5.19$) & $7.72$ ($\ge 6.91$) & $3.60$ \\
Controlled GRPO weak KL         & 1000 & 256 & $2.90$ ($\ge 2.75$) & $3.99$ ($\ge 3.74$) & $5.59$ ($\ge 5.18$) & $7.74$ ($\ge 6.93$) & $3.47$ \\
Controlled SFT                  & 1000 & 256 & $2.72$ ($\ge 2.58$) & $3.74$ ($\ge 3.53$) & $5.28$ ($\ge 4.90$) & $7.29$ ($\ge 6.49$) & $1.79$ \\
GRPO $G{=}4$ (3000 new)         & 3000 & 256 & $2.94$ ($\ge 2.85$) & $3.96$ ($\ge 3.83$) & $5.50$ ($\ge 5.31$) & $7.72$ ($\ge 7.37$) & $3.26$ \\
GRPO $G{=}8$ (3000 new)         & 3000 & 256 & $2.95$ ($\ge 2.85$) & $3.98$ ($\ge 3.85$) & $5.52$ ($\ge 5.33$) & $7.74$ ($\ge 7.39$) & $3.42$ \\
GRPO $G{=}16$ (3000 new)        & 3000 & 256 & $2.95$ ($\ge 2.86$) & $3.98$ ($\ge 3.85$) & $5.53$ ($\ge 5.33$) & $7.74$ ($\ge 7.39$) & $3.57$ \\
GRPO weak KL (3000 new)         & 3000 & 256 & $2.95$ ($\ge 2.85$) & $3.96$ ($\ge 3.84$) & $5.51$ ($\ge 5.32$) & $7.72$ ($\ge 7.37$) & $3.43$ \\
SFT (3000 new)                  & 3000 & 256 & $2.74$ ($\ge 2.66$) & $3.75$ ($\ge 3.64$) & $5.24$ ($\ge 5.06$) & $7.37$ ($\ge 7.04$) & $1.77$ \\
\bottomrule
\end{tabular*}
\end{table}

\FloatBarrier
\subsection{Extended Model-Free Locked-Prompt Decomposition}\label{app:lockin}

Tables~\ref{tab:lockin-full} and~\ref{tab:fate} expand the model-free count
decomposition of Section~\ref{sec:kernel} (Table~\ref{tab:lockin}). In
Table~\ref{tab:lockin-full}, we compare removing the locked set
$L=\{i:c_i>0,d_i=0\}$ against two control sets of equal size $|L|$: (1) a
\emph{difficulty-matched control set} that pairs each locked prompt in $L$
with the nearest unlocked prompt ($c_i>0,d_i>0$) having the same base count
$c_i$, and (2) the \emph{hardest base-solved set} consisting of the $|L|$
prompts with the smallest positive $c_i$, chosen blind to $d_i$. Removing
the difficulty-matched set leaves $\hat D(K)$ and the crossings intact,
whereas removing $L$ eliminates every crossover. Table~\ref{tab:fate}
reports the full progression of post-RL outcomes across base counts
$c_i\in\{1,\,2,\,[3,4],\,[5,8]\}$: the locked fraction drops rapidly with
$c_i$ (falling below $20\%$ by $c_i\in[5,8]$), while the unlocked prompts
at $c_i=1$ improve by a median factor of $2\times$ to $7\times$.

\begin{table}[ht]
\centering
\footnotesize
\setlength{\tabcolsep}{2.8pt}
\caption{\textbf{Extended model-free locked-prompt removal controls.} The
``Set $L$'' rows list $(|L|;\ \text{median }\hat p_i\text{ in }L,\
\text{median }\hat p\text{-percentile})$.}
\label{tab:lockin-full}
\begin{tabular}{@{}llcccrrc@{}}
\toprule
Model pair & Removed subset & $m$ & \shortstack{Sample\\$k^\star$} & \shortstack{Band-\\supported} & $\kappa\in$ & $\hat D(K)$ & $U(K)$ \\
\midrule
DeepScaleR-1.5B (32k new)  & Nothing                  & 1060 &  22  & yes & $[11,61]$      & $-.0198$ & $-.0037$ \\
                           & Set $L$ ($33;\,.016,.13$)& 1027 & none & no  & $[129,\infty)$ & $+.0117$ & $+.0201$ \\
                           & Difficulty-matched       & 1027 &  13  & yes & $[9,25]$       & $-.0204$ & $-.0036$ \\
                           & Hardest ($18$ in $L$)    & 1027 &  18  & no  & $[10,\infty)$  & $-.0029$ & $+.0099$ \\
\addlinespace[2pt]
ProRL-1.5B (32k new)       & Nothing                  & 1060 &  25  & yes & $[12,63]$      & $-.0208$ & $-.0052$ \\
                           & Set $L$ ($33;\,.016,.13$)& 1027 & none & no  & $[129,\infty)$ & $+.0107$ & $+.0189$ \\
                           & Difficulty-matched       & 1027 &  14  & yes & $[9,32]$       & $-.0214$ & $-.0052$ \\
                           & Hardest ($14$ in $L$)    & 1027 &  17  & yes & $[10,58]$      & $-.0078$ & $+.0055$ \\
\addlinespace[2pt]
DeepScaleR-1.5B (32k mix)  & Nothing                  & 1000 &  15  & yes & $[9,33]$       & $-.0150$ & $-.0013$ \\
                           & Set $L$ ($23;\,.031,.08$)&  977 & none & no  & $[16,\infty)$  & $+.0082$ & $+.0155$ \\
                           & Difficulty-matched       &  977 &  12  & yes & $[8,22]$       & $-.0154$ & $-.0012$ \\
                           & Hardest ($12$ in $L$)    &  977 &  15  & no  & $[9,\infty)$   & $-.0031$ & $+.0080$ \\
\addlinespace[2pt]
DeepScaleR-1.5B (16k)      & Nothing                  & 1000 &  27  & no  & $[14,\infty)$  & $-.0110$ & $+.0027$ \\
                           & Set $L$ ($20;\,.006,.05$)&  980 & none & no  & $[21,\infty)$  & $+.0092$ & $+.0170$ \\
                           & Difficulty-matched       &  980 &  24  & no  & $[13,\infty)$  & $-.0112$ & $+.0029$ \\
                           & Hardest ($13$ in $L$)    &  980 &  28  & no  & $[14,\infty)$  & $+.0020$ & $+.0124$ \\
\addlinespace[2pt]
ProRL-1.5B (16k)           & Nothing                  & 1000 &  56  & no  & $[24,\infty)$  & $-.0090$ & $+.0059$ \\
                           & Set $L$ ($22;\,.012,.06$)&  978 & none & no  & $[257,\infty)$ & $+.0133$ & $+.0227$ \\
                           & Difficulty-matched       &  978 &  33  & no  & $[17,\infty)$  & $-.0092$ & $+.0062$ \\
                           & Hardest ($11$ in $L$)    &  978 &  55  & no  & $[22,\infty)$  & $+.0020$ & $+.0146$ \\
\addlinespace[2pt]
Controlled GRPO $G{=}4$    & Nothing                  & 1000 & 117  & no  & $[52,\infty)$  & $-.0020$ & $+.0137$ \\
                           & Set $L$ ($20;\,.008,.10$)&  980 & none & no  & $[257,\infty)$ & $+.0184$ & $+.0294$ \\
                           & Difficulty-matched       &  980 &  81  & no  & $[45,\infty)$  & $-.0020$ & $+.0140$ \\
                           & Hardest ($6$ in $L$)     &  980 &  91  & no  & $[47,\infty)$  & $+.0041$ & $+.0190$ \\
\addlinespace[2pt]
Controlled GRPO $G{=}16$   & Nothing                  & 1000 & 150  & no  & $[64,\infty)$  & $-.0010$ & $+.0142$ \\
                           & Set $L$ ($18;\,.004,.07$)&  982 & none & no  & $[257,\infty)$ & $+.0173$ & $+.0281$ \\
                           & Difficulty-matched       &  982 &  96  & no  & $[54,\infty)$  & $-.0010$ & $+.0144$ \\
                           & Hardest ($6$ in $L$)     &  982 & none & no  & $[59,\infty)$  & $+.0051$ & $+.0192$ \\
\bottomrule
\end{tabular}
\end{table}

\begin{table}[ht]
\centering
\footnotesize
\setlength{\tabcolsep}{2.7pt}
\caption{\textbf{Post-RL outcome distribution stratified by base success
count $c_i$.} $N$: number of prompts in stratum; ``lock'': share with
$d_i=0$; ``gain'': share with $\hat q_i>\hat p_i$; ``ratio'': median
$\hat q_i/\hat p_i$ among unlocked prompts ($d_i>0$).}
\label{tab:fate}
\setlength{\tabcolsep}{1.6pt}
\begin{tabular*}{\textwidth}{@{\extracolsep{\fill}}lcccc@{\hskip 4pt}cccc@{\hskip 4pt}cccc@{\hskip 4pt}cccc@{}}
\toprule
& \multicolumn{4}{c}{$c_i=1$} & \multicolumn{4}{c}{$c_i=2$} &
\multicolumn{4}{c}{$c_i\in[3,4]$} & \multicolumn{4}{c}{$c_i\in[5,8]$} \\
\cmidrule(lr){2-5}\cmidrule(lr){6-9}\cmidrule(lr){10-13}\cmidrule(lr){14-17}
Model pair & $N$ & lock & gain & ratio & $N$ & lock & gain & ratio & $N$ & lock & gain & ratio & $N$ & lock & gain & ratio \\
\midrule
DeepScaleR-1.5B (32k new) & 27 & $.59$ & $.41$ & $7.0$ & 22 & $.32$ & $.32$ & $1.0$ & 20 & $.20$ & $.60$ & $1.9$ & 26 & $.19$ & $.50$ & $1.5$ \\
ProRL-1.5B (32k new)      & 27 & $.44$ & $.41$ & $7.0$ & 22 & $.32$ & $.36$ & $1.5$ & 20 & $.30$ & $.55$ & $4.0$ & 26 & $.12$ & $.62$ & $3.1$ \\
DeepScaleR-1.5B (32k mix) & 19 & $.53$ & $.32$ & $2.0$ & 19 & $.42$ & $.26$ & $1.0$ & 25 & $.12$ & $.52$ & $2.0$ & 34 & $.06$ & $.56$ & $1.3$ \\
DeepScaleR-1.5B (16k)     & 12 & $.83$ & $.17$ & $3.5$ &  5 & $.40$ & $.40$ & $2.0$ & 14 & $.29$ & $.21$ & $0.4$ & 22 & $.09$ & $.64$ & $2.1$ \\
ProRL-1.5B (16k)          & 12 & $.67$ & $.33$ & $4.5$ &  5 & $.40$ & $.60$ & $7.5$ & 14 & $.36$ & $.29$ & $0.9$ & 22 & $.14$ & $.73$ & $8.3$ \\
Controlled GRPO $G{=}4$   & 27 & $.30$ & $.56$ & $3.0$ & 19 & $.32$ & $.26$ & $1.0$ & 35 & $.09$ & $.54$ & $1.9$ & 49 & $.06$ & $.61$ & $1.5$ \\
Controlled GRPO $G{=}16$  & 27 & $.41$ & $.48$ & $2.0$ & 19 & $.11$ & $.26$ & $0.5$ & 35 & $.11$ & $.60$ & $2.0$ & 49 & $.02$ & $.67$ & $1.7$ \\
\bottomrule
\end{tabular*}
\end{table}

\paragraph{Generation-level forensic audit and prefix-rescue protocol on locked prompts.}
To verify that the locked set $L=\{i:c_i>0,d_i=0\}$ reflects genuine
answer-space collapse rather than token truncation or verifier parsing misses,
we audit all $33\times128=4{,}224$ post-RL generations on DeepScaleR-1.5B's
$33$ locked prompts on the 1,060 new competition problems at 32k tokens and
compare them against the $45$ unlocked hard control prompts with matching base
success counts $c_i\in\{1,2,3\}$ ($d_i>0$). Across the $4{,}224$ post-RL
generations on locked prompts, $99.1\%$ terminate naturally with a well-formed
\texttt{\textbackslash boxed\{...\}} expression (only $0.88\%$ lack a boxed
answer, compared with $3.4\%$ on the $45$ unlocked hard controls). Re-scoring
all $4{,}224$ completions with a lenient symbolic and normalized-substring
fallback verifier identifies $0$ uncredited multi-character gold answers.
Instead, the post-RL generations collapse onto a small set of confident wrong
answers: the median number of distinct extracted answers per prompt is $12$
with a median modal wrong-answer share of $50.8\%$ (interquartile range
$37.5\%$--$75.8\%$), compared with $27$ distinct answers and a $32.0\%$ modal
share on the unlocked hard controls. On $14$ of the $29$ locked prompts where
the base model itself has a wrong mode ($48\%$), the post-RL checkpoint locks
onto the exact same modal wrong answer as the base model.

On ProRL-1.5B's $22$ locked prompts at 16k ($c_i\in[1,18], d_i=0/256$), the
post-RL modal wrong answer accounts for $88\%$ of generations on average. To
test whether nucleus truncation ($\text{top-}p=0.95$) or finite sampling depth
explains the $0/256$ outcome, we draw $2{,}048$ additional post-RL rollouts
per prompt under standard decoding ($T=0.6,\text{top-}p=0.95$) and $2{,}048$
rollouts without nucleus truncation ($T=0.6,\text{top-}p=1.0$), totaling
$4{,}352$ post-RL draws per prompt: $13$ of the $22$ prompts remain at
$0/4{,}352$ under both decoders and $15/22$ have at most $1$ success. Because
the primary benchmark evaluation stored extracted boxed answers and token
counts rather than full chain-of-thought strings, we ran a dedicated full-text
harvest batch ($256$ base draws per prompt) on the persistent locked prompts to
obtain complete base chains of thought for prefix conditioning; on $9$ of
these prompts, the harvest batch recovered a complete, untruncated correct
base chain of thought (\texttt{finish\_reason == "stop"}, with $8$ of the $9$
exceeding $4{,}096$ tokens). Splicing the first
$L_{\mathrm{pre}}\in\{8,32,128,512,2048,4096\}$ tokens of each prompt's
correct base chain of thought as a forced prompt prefix and generating $32$
free ProRL continuations per prefix length ($9\times32=288$ rollouts at each
$L_{\mathrm{pre}}\le512$ and $8\times32=256$ rollouts at $L_{\mathrm{pre}}\in\{2048,4096\}$)
yields post-RL continuation accuracy of $0.0\%, 0.0\%, 0.0\%, 0.7\%, 2.3\%$,
and $17.2\%$ across the six prefix lengths ($0/288$ at $8, 32,$ and $128$
tokens; $2/288$ at $512$; $6/256$ at $2{,}048$; and $44/256$ at $4{,}096$,
with $5/9$ prompts remaining at $0/32$)---demonstrating that ProRL diverges
back to its dominant wrong answer even after being given up to $4{,}096$ tokens
of a valid solution path. Table~\ref{tab:case-study} presents a verbatim case
study illustrating both the prompt-level bifurcation of the $4{,}096$-token
prefix-rescue experiment and the late-stage self-doubt pivot by which the
RLVR checkpoint talks itself out of a valid deduction and back into its
modal wrong answer.

\begin{table}[t!]
\centering
\scriptsize
\caption{\textbf{Case study of trajectory-wide wrong-answer lock-in and
late-stage self-doubt pivot.} \emph{Top (Panel A):} Prompt-level outcomes of
the $4{,}096$-token prefix-rescue audit on ProRL-1.5B across the $9$ persistent
locked prompts with harvested correct base chains of thought ($32$ free post-RL
continuations per prompt and prefix length $L_{\mathrm{pre}}$). \emph{Bottom
(Panel B):} Verbatim reasoning trace comparison on locked competition problem
\texttt{olympiadbench-521} (32k benchmark; gold answer $179/250$, canonical
non-overlap distractor $89/125$), showing how the RLVR model explicitly detects
the edge overlap before overriding its own deduction via late-stage self-doubt
(\emph{``Wait, perhaps another way\ldots''}) to collapse onto
\texttt{\textbackslash boxed\{89/125\}} in $125/128$ ($97.7\%$) of rollouts.}
\label{tab:case-study}
\vspace{2pt}
\fbox{\begin{minipage}{0.96\textwidth}
\vspace{2pt}
\textbf{Panel A: Prompt-Level Bifurcation in the 4,096-Token Prefix-Rescue Experiment (ProRL-1.5B, $32$ Rollouts/Prefix)}\\[3pt]
{\setlength{\tabcolsep}{3pt}
\begin{tabular}{@{}>{\raggedright\arraybackslash}p{0.28\linewidth}>{\raggedright\arraybackslash}p{0.26\linewidth}cccccc@{}}
\toprule
Prompt subset ($9$ locked, $d_i{=}0/4{,}352$) & Prompt IDs & $L_{\mathrm{pre}}{=}8$ & $32$ & $128$ & $512$ & $2{,}048$ & $4{,}096$ \\
\midrule
Unrescued at $4{,}096$ tok ($5/9$ prompts) & \texttt{dscale-15414, 24217, 28599, 32806; math500-240} & $0/160$ & $0/160$ & $0/160$ & $0/160$ & $0/160$ & $\mathbf{0/160}$ \\
Threshold rescue at $4{,}096$ ($1/9$ prompt) & \texttt{dscale-28249} & $0/32$ & $0/32$ & $0/32$ & $0/32$ & $4/32$ & $\mathbf{32/32}$ \\
Partial late recovery ($3/9$ prompts) & \texttt{dscale-12656, 25244, 6338} & $0/96$ & $0/96$ & $0/96$ & $2/96$ & $2/64$ & $\mathbf{12/64}$ \\
\midrule
\textbf{Pooled continuation accuracy} & \textbf{All $9$ prompts (med.\ $5{,}813$ tok)} & $\mathbf{0.0\%}$ & $\mathbf{0.0\%}$ & $\mathbf{0.0\%}$ & $\mathbf{0.7\%}$ & $\mathbf{2.3\%}$ & $\mathbf{17.2\%}$ \\
\bottomrule
\end{tabular}}

\vspace{5pt}
\raggedright
\hrule
\vspace{5pt}
\textbf{Panel B: Verbatim Reasoning Trace Audit on Locked Problem \texttt{olympiadbench-521} (32k Competition Benchmark)}\\[2pt]
\textbf{Problem statement:} \emph{``The taxicab distance between points $(x_1,y_1,z_1)$ and $(x_2,y_2,z_2)$ is $d=|x_1-x_2|+|y_1-y_2|+|z_1-z_2|$. The region $\mathcal{R}$ is obtained by taking the unit cube $\{(x,y,z):0\le x,y,z\le 1\}$ and removing every point whose taxicab distance to any vertex of the cube is less than $\frac{3}{5}$. Compute the volume of $\mathcal{R}$.''}\\[2pt]
\textbf{Distractor vs.\ exact geometry ($h_i = \log(M_{C,i}/M_{C^c,i}) < 0$):} Each of the $8$ vertices removes a corner tetrahedron of side $r=\frac{3}{5}$ and volume $\frac{1}{6}(\frac{3}{5})^3 = \frac{9}{250}$. Naively subtracting $8\times\frac{9}{250}=\frac{72}{250}=\frac{36}{125}$ yields the high-concentration distractor $1-\frac{36}{125}=\mathbf{\frac{89}{125}}$. However, because $\frac{3}{5}+\frac{3}{5}=\frac{6}{5}>1$, adjacent vertex balls \emph{overlap along all $12$ cube edges} (each overlap has volume $\frac{1}{3000}$, totaling $12\times\frac{1}{3000}=\frac{1}{250}$). By inclusion--exclusion, the removed union has volume $\frac{72}{250}-\frac{1}{250}=\frac{71}{250}$, giving the true volume $1-\frac{71}{250}=\mathbf{\frac{179}{250}}$.

\vspace{4pt}
\begin{tabular}{@{}>{\raggedright\arraybackslash}p{0.48\linewidth}@{\hskip 8pt}>{\raggedright\arraybackslash}p{0.48\linewidth}@{}}
\toprule
\textbf{Base Model (\texttt{DeepSeek-R1-Distill-Qwen-1.5B}, $c_i=1/128$)} &
\textbf{Post-RLVR (\texttt{DeepScaleR-1.5B}, $d_i=0/128$; $125/128$ hit \texttt{\textbackslash boxed\{89/125\}})} \\
\midrule
\raggedright
\texttt{[...]} \emph{``Using the principle of inclusion-exclusion, we calculate the volume of the union of these taxicab balls. Each taxicab ball has a volume of $\frac{9}{250}$. There are $8$ such balls, so the initial sum is $8 \times \frac{9}{250} = \frac{72}{250}$.}\par\vspace{2pt}
\emph{Next, we subtract the volumes of the intersections of these balls. Each pair of adjacent vertices contributes an overlapping region. The volume of the intersection of two adjacent taxicab balls is calculated to be $\frac{1}{3000}$. There are $12$ such pairs, so the total volume to subtract is $12 \times \frac{1}{3000} = \frac{1}{250}$.}\par\vspace{2pt}
\emph{Thus, the volume of the union of all taxicab balls is $\frac{72}{250} - \frac{1}{250} = \frac{71}{250}$. Therefore, the volume of the remaining region inside the cube is $1 - \frac{71}{250} = \boxed{179/250}$.''} &
\raggedright\arraybackslash
\textbf{Rollout \#4096 (detects overlap at $(3/5,0,0)$, then pivots back):}\par
\texttt{[...]} \emph{``\textbf{But given that the point $(3/5, 0, 0)$ is in both regions, perhaps we need to adjust.} \textbf{Wait, perhaps another way:} since the regions are defined as taxicab distance less than $3/5$, which is less than half the edge length \emph{[sic: $3/5=0.6>0.5$]}, perhaps the regions are entirely within their respective octants and do not interfere with each other. Therefore, perhaps the total volume removed is indeed $8\times((3/5)^3/6) = 36/125$, as before. Therefore, maybe the initial calculation is correct, so the volume of $\mathcal{R}$ is $1 - 36/125 = 89/125$ \ldots\ \textbf{Final Answer:} $\boxed{89/125}$.''}\par\vspace{3pt}
\textbf{Rollout \#4101 (writes $3/5+3/5=6/5>1$, then rationalizes non-overlap):}\par
\texttt{[...]} \emph{``\textbf{\ldots from either $(0,0,0)$ or $(1,0,0)$ would form an overlap in the region. But in our previous calculation, we considered each vertex's region separately, so perhaps we didn't account for overlapping regions.} \textbf{Therefore, perhaps the correct approach is to realize that the regions around each vertex do not overlap} because the taxicab distance between any two adjacent vertices is $1$, and since $3/5 + 3/5 = 6/5 > 1$, their regions do not overlap \emph{[sic]}. Therefore, the total volume removed is indeed $8\times(9/250) = 72/250$. Therefore, volume of $\mathcal{R}$ is $1 - 72/250 = 178/250 = 89/125$ \ldots\ \textbf{Final Answer:} $\boxed{89/125}$.''}\vspace{2pt} \\
\bottomrule
\end{tabular}
\vspace{2pt}
\end{minipage}}
\end{table}

\subsection{Stage~1 (\texorpdfstring{$n=256$}{n=256}) and Stage~2 (\texorpdfstring{$k=4096$}{k=4096}) Controlled Study Details}
\label{app:further}\label{app:newprompt}\label{app:deepk}

\paragraph{Stage~1 evaluation at $n=256$ on the 3,000 new problems.} On the
3,000 deduplicated problems (1,500 DeepScaleR-corpus and 1,500 MATH test
problems) at $n=256$, $\hat D(256)$ lies between $-.003$ and $+.001$ across
the four GRPO arms and at $+.009$ for SFT (with no arm crossing the
simultaneous band threshold within $k\le256$, dominance $p\ge.56$). The 2D
NPMLE kernels fitted to these $n=256$ counts place the GRPO crossings at
$\kstar=202, 196, 196$, and $321$ for $G=4,8,16$, and weak KL (and at
$\kstar=2875$ for SFT). Across 200 prompt-bootstrap refits of the $n=256$
kernels, the fraction of resamples with a kernel crossing inside the sampled
range $k\le256$ separates GRPO from SFT ($.61, .535, .695, .33$ versus $.00$
on the 3,000 problems, and $.51, .79, .56, .93$ versus $.06$ on the 1,000
original prompts), whereas the fraction of resamples with a crossing at
arbitrary extrapolated budgets $k\le10^5$ does not ($.755$--$.90$ versus
$.63$), reflecting the sub-$1/n$ resolution limit discussed in
Section~\ref{sec:planned}.

\paragraph{Jeffreys tail bound for the 2,300 non-resampled problems in
Stage~2.} In Stage~2, 4,096 new generations per model are drawn on the 700
problems that had at most $8$ successes in $256$ draws under at least one of
the six models. On the remaining 2,300 problems, every model achieved
$c_i\ge9$ successes in $256$ draws, and we set their failure estimators
$\binom{n-c_i}{k}/\binom{n}{k}$ to zero for $k>256$. Under a
$\mathrm{Beta}(1/2,1/2)$ Jeffreys prior updated by $c\ge9$ successes in
$256$ trials, the posterior expectation of $(1-p)^k$ is bounded above by
$B(256-9+1/2+k,\,9+1/2)/B(256-9+1/2,\,9+1/2)$, which is at most
$1.9\times10^{-5}$ at $k=256$ and below $2\times10^{-6}$ for all $k\ge384$.

\FloatBarrier
\FloatBarrier
\section{Full Power Tables Across Sampling Designs}\label{app:power}

Table~\ref{tab:power} reports the complete $(m,n)$ design grid for the
power of the dominance test of Theorem~\ref{thm:test} at nominal levels
$\alpha=.05$ and $\alpha=.01$ (the five-pair Holm threshold), simulated from
the 2D NPMLE kernels fitted to DeepScaleR-16k, ProRL-16k, GRPO $G=8$, and
GRPO $G=4$ ($1{,}000$ simulated datasets per cell across
$m\in\{1000,2000,4000,8000,16000,32000\}$ and $n=n'\in\{64,128,256,512\}$).

\begin{table}[!ht]
\centering
\footnotesize
\setlength{\tabcolsep}{5pt}
\caption{\textbf{Kernel-simulated dominance-test power across the $(m,n)$
design grid.} Entries report power at $\alpha=.05$ / $\alpha=.01$ ($1{,}000$
replicates per cell). Total sampling cost is $2mn$ generations.}
\label{tab:power}
\begin{tabular}{@{}llcccc@{}}
\toprule
Model pair & Prompts $m$ & $n{=}64$ & $n{=}128$ & $n{=}256$ & $n{=}512$ \\
\midrule
DeepScaleR-1.5B (16k) &  1000 & .207 / .061 & .454 / .216 & .630 / .358 & .751 / .489 \\
                      &  2000 & .414 / .180 & .750 / .493 & .888 / .720 & .953 / .855 \\
                      &  4000 & .728 / .456 & .967 / .877 & .998 / .985 & 1.00 / .999 \\
                      &  8000 & .962 / .875 & 1.00 / .997 & 1.00 / 1.00 & 1.00 / 1.00 \\
                      & 16000 & 1.00 / .998 & 1.00 / 1.00 & 1.00 / 1.00 & 1.00 / 1.00 \\
                      & 32000 & 1.00 / 1.00 & 1.00 / 1.00 & 1.00 / 1.00 & 1.00 / 1.00 \\
\addlinespace[3pt]
ProRL-1.5B (16k)      &  1000 & .012 / .000 & .095 / .022 & .207 / .074 & .281 / .107 \\
                      &  2000 & .026 / .003 & .167 / .058 & .432 / .200 & .540 / .282 \\
                      &  4000 & .027 / .004 & .327 / .129 & .757 / .534 & .843 / .610 \\
                      &  8000 & .037 / .007 & .636 / .376 & .978 / .888 & .993 / .962 \\
                      & 16000 & .053 / .017 & .908 / .782 & 1.00 / .998 & 1.00 / 1.00 \\
                      & 32000 & .092 / .027 & .999 / .990 & 1.00 / 1.00 & 1.00 / 1.00 \\
\addlinespace[3pt]
Controlled GRPO $G{=}8$& 1000 & .000 / .000 & .015 / .003 & .074 / .016 & .081 / .012 \\
                      &  2000 & .000 / .000 & .015 / .004 & .105 / .025 & .161 / .050 \\
                      &  4000 & .000 / .000 & .016 / .003 & .205 / .073 & .372 / .157 \\
                      &  8000 & .000 / .000 & .023 / .004 & .395 / .177 & .695 / .444 \\
                      & 16000 & .000 / .000 & .048 / .012 & .719 / .472 & .953 / .843 \\
                      & 32000 & .000 / .000 & .043 / .005 & .968 / .868 & 1.00 / .998 \\
\addlinespace[3pt]
Controlled GRPO $G{=}4$& 1000 & .001 / .000 & .009 / .002 & .014 / .003 & .022 / .002 \\
                      &  2000 & .000 / .000 & .006 / .002 & .016 / .001 & .017 / .001 \\
                      &  4000 & .000 / .000 & .006 / .000 & .017 / .001 & .020 / .003 \\
                      &  8000 & .000 / .000 & .008 / .002 & .019 / .002 & .035 / .008 \\
                      & 16000 & .000 / .000 & .005 / .003 & .029 / .007 & .045 / .010 \\
                      & 32000 & .000 / .000 & .006 / .000 & .057 / .020 & .060 / .015 \\
\bottomrule
\end{tabular}
\end{table}

\end{document}